\documentclass[journal]{IEEEtran}
\usepackage{graphicx} 
\usepackage{cite}
\usepackage[colorlinks=true, linkcolor=blue, citecolor=green, urlcolor=blue]{hyperref}
\usepackage{bm}
\usepackage{amsmath}
\usepackage{makecell}
\def\BibTeX{{\rm B\kern-.05em{\sc i\kern-.025em b}\kern-.08em
    T\kern-.1667em\lower.7ex\hbox{E}\kern-.125emX}}
\usepackage[table,dvipsnames]{xcolor}
\usepackage{bbding}
\usepackage{booktabs} 
\usepackage{multirow}
\usepackage{amssymb} 
\usepackage{pifont}  

\colorlet{colorFst}{Green!25}       
\colorlet{colorSnd}{SpringGreen!45} 
\colorlet{colorTrd}{Yellow!30}      
\newcommand{\fs}{\cellcolor{colorFst}\bf}   
\newcommand{\nd}{\cellcolor{colorSnd}}      

\newcommand{\greencheck}{{\color{green}\checkmark}}
\newcommand{\redx}{{\color{red}\ding{55}}}

\begin{document}

\title{RP-SLAM: Real-time Photorealistic SLAM with Efficient 3D Gaussian Splatting}
\author{Lizhi Bai, Chunqi Tian*, Jun Yang, Siyu Zhang, Masanori Suganuma, Takayuki Okatani
  \thanks{Lizhi Bai, Chunqi Tian, Jun Yang and Siyu Zhang are with Department of Computer Science and Technology, Tongji University, Shanghai, 201804, China (e-mail: \{bailizhi, tianchunqi, junyang,  zsyzsy\}@tongji.edu.cn).}
  \thanks{Lizhi Bai, Masanori Suganuma and Takayuki Okatani are with Graduate School of Information Sciences, Tohoku University, Sendai, 980-0845, Japan (e-mail: \{suganuma, okatani\}@vision.is.tohoku.ac.jp).}
  \thanks{* Corresponding author}}

\maketitle

\begin{abstract}
3D Gaussian Splatting has emerged as a promising technique for high-quality 3D rendering, leading to increasing interest in integrating 3DGS into realism SLAM systems. However, existing methods face challenges such as Gaussian primitives redundancy, forgetting problem during continuous optimization, and difficulty in initializing primitives in monocular case due to lack of depth information. In order to achieve efficient and photorealistic mapping, we propose RP-SLAM, a 3D Gaussian splatting-based vision SLAM method for monocular and RGB-D cameras. RP-SLAM decouples camera poses estimation from Gaussian primitives optimization and consists of three key components. Firstly, we propose an efficient incremental mapping approach to achieve a compact and accurate representation of the scene through adaptive sampling and Gaussian primitives filtering. Secondly, a dynamic window optimization method is proposed to mitigate the forgetting problem and improve map consistency. Finally, for the monocular case, a monocular keyframe initialization method based on sparse point cloud is proposed to improve the initialization accuracy of Gaussian primitives, which provides a geometric basis for subsequent optimization. The results of numerous experiments demonstrate that RP-SLAM achieves state-of-the-art map rendering accuracy while ensuring real-time performance and model compactness.
\end{abstract}


\section{Introduction}
 Visual Simultaneous Localization and Mapping (vSLAM) has long served as a foundational technology in robotics and computer vision, with decades of research advancing its applications across diverse fields\cite{campos2021orb,teed2021droid,engel2017direct,engel2014lsd}. The advent of sophisticated, real-world applications such as autonomous driving, augmented and virtual reality, and embodied intelligence has given rise to a multitude of novel demands that extend beyond the conventional real-time tracking and mapping. These advancements require SLAM systems not only to provide precise spatial localization but also to deliver photorealistic scene reconstruction, where achieving high-fidelity visual representation is crucial.

Traditional vSLAM systems, exemplified by ORB-SLAM3\cite{campos2021orb}, LSD-SLAM\cite{engel2014lsd} and DSO\cite{engel2017direct}, rely on sparse feature extraction methods and are primarily concerned with providing accurate but visually simplistic representations of the surrounding environment. Consequently, these approaches are not sufficient to meet the demands of photorealistic scene reconstruction. 

Recent studies have investigated the potential of implicit scene representations, in particular Neural Radiance Field (NeRF)\cite{mildenhall2021nerf}, as a solution for dense and realistic 3D reconstruction. A number of approaches have been developed which integrate NeRF into SLAM system, with the objective of optimizing camera poses and map representation through neural rendering. To illustrate, iMap\cite{sucar2021imap} employs a multi-layer perceptron (MLP) to represent the scene and optimize the camera poses. NICE-SLAM\cite{zhu2022nice} utilizes a hierarchical grid to represent the 3D scene and optimize the implicit features stored in the grid nodes. Additionally, there are many techniques\cite{rosinol2023nerf,bai2024neb,wang2023co} that achieve enhanced performance through the utilization of multi-resolution hash grids\cite{muller2022instant}. Despite the considerable potential of NeRF in the generation of realistic scenes, it is confronted with a number of challenges when integrated into SLAM systems, including the high computational cost, prolonged training time, and vulnerability to catastrophic forgetting.

A recently developed explicit scene representation technique based on 3D Gaussian Sputtering (3DGS)\cite{kerbl3Dgaussians} has demonstrated the potential to provide a compelling solution for high-quality 3D rendering. In comparison to NeRF, 3DGS has been shown to achieve comparable rendering quality while exhibiting significantly superior rendering speed. 

In light of the enhancements in rendering efficiency and optimization of 3DGS, there has been a amount of research conducted into the integration of 3DGS into dense and photorealistic SLAM systems. Notably, SplaTAM\cite{keetha2024splatam} and MonoGS\cite{matsuki2024gaussian} are the inaugural approaches to utilisze 3DGS for coupling estimation of camera poses with optimization of scene representation. 
These coupled approaches require tedious iterations to optimize the camera poses based on the scene representation, making them challenging to execute in real time.

Concurrent works\cite{peng2024rtg,huang2024photo,ha2025rgbd} decouple the estimation of camera poses from the optimization of Gaussian primitives. They employ conventional techniques\cite{newcombe2011kinectfusion,campos2021orb,segal2009generalized} to estimate the camera poses, circumventing the time-consuming iterations inherent to the coupled approaches, thereby enhancing real-time performance. For incremental mapping, these methods typically rely on dense pixel sampling when initializing new Gaussian primitives, leading to redundant primitives and significant storage overhead. In order to optimize the scene representation, the prevailing method involves the selection of a specific number of keyframes from the keyframe set, which are then merged with the newly added keyframe into a fixed keyframe window. Only keyframes located within this fixed window are used for map optimization. However, this approach may potentially lead to erroneous local minima and forgetting problems during successive iterations.

The absence of depth information in the monocular case makes it challenging to accurately add new primitives. MonoGS\cite{matsuki2024gaussian} employs random depths with no geometric basis for initialization, and its complete dependence on the mapping process to optimize the initial primitives makes it difficult to obtain an accurate representation of the scene.
Photo-SLAM\cite{huang2024photo} and CaRtGS\cite{feng2024cartgs} employ only the spatial gradient-based densification method in the original 3DGS\cite{kerbl3Dgaussians} to add new primitives. However, the limited number of iterations is insufficient to optimize the new primitives to accurate positions, resulting in artifacts due to the inconsistency between the geometric and photometric properties of the Gaussian primitives.

In order to address these challenges, we propose a real-time photorealistic SLAM with efficient 3D Gaussian splatting for RGB-D and monocular cameras, namely RP-SLAM. RP-SLAM is a decoupled system that utilizes feature-based SLAM for camera tracking and 3DGS for optimizing the photorealistic scene representation. 

Our method comprises three main components. Firstly, We propose an efficient incremental mapping method that incorporates image sampling and Gaussian primitives filtering. In contrast to previous dense image sampling, we propose adaptive sampling using image gradients to focus the sampling and computational resources on texture-rich regions and reduce the generation of redundant samples. In conjunction with Gaussian primitives filtering, redundant primitives are further eliminated to achieve an efficient and compact scene representation. Secondly, to achieve a consistent scene representation, we propose dynamic keyframe window optimization. Compared to the fixed keyframe window, this method dynamically adjusts the keyframes to be optimized at each iteration based on the covisibility between keyframes, alleviating the forgetting problem during the continuous optimization process. Finally, in order to add new Gaussian primitives in the monocular case where geometric information is absent, we propose a monocular keyframe initialization method based on sparse point cloud. In comparison to alternative methods with uncertainty in creating new primitives, this method can create new primitives relatively accurately using the initial geometric information and provide a geometric basis for subsequent optimization.

To summarize, our contributions are as follows:
\begin{itemize}
    \item \textbf{Efficient Incremental Mapping}: By using adaptive sampling guided by local image gradients and Gaussian primitives filtering, our method reduces redundant Gaussian primitives while maintaining high rendering quality.
\end{itemize}

\begin{itemize}
    \item \textbf{Robust Mapping with Dynamic Keyframe Window}: Dynamic keyframe window optimization mitigates the impact of forgetting, enhancing mapping consistency.
\end{itemize}

\begin{itemize}
    \item \textbf{Improved Monocular Keyframe Initialization}: For monocular cameras, our keyframe initialization approach enables relatively accurate placement of Gaussian primitives, improving rendering quality and reducing redundancy.
\end{itemize}

\begin{itemize}
    \item We validate our approach on standard RGB-D and monocular datasets (TUM\cite{sturm2012benchmark}, Replicia\cite{straub2019replica} and ScanNet++\cite{yeshwanth2023scannet++}), demonstrating its superior efficiency, rendering quality, and robustness compared to existing methods.
\end{itemize}

\section{Related Work}

\subsection{Classical Visual SLAM}
Classical SLAM methods typically employ factor graphs to model the intricate optimization challenges between variables and measurements. In order to achieve real-time operation, these methods incrementally update the estimated pose, thereby avoiding computationally expensive processes. To illustrate, the ORB-SLAM series\cite{mur2015orb,mur2017orb,campos2021orb} are based on the extraction and tracking of lightweight geometric features across sequential frames, with bundle adjustment conducted locally rather than globally. Direct SLAM approaches, such as LSD-SLAM\cite{engel2014lsd} and DSO\cite{engel2017direct}, operate directly with raw image intensities, circumventing the necessity for geometric feature extraction. They are capable of maintaining a sparse or semi-dense map represented by point cloud, even in the context of resource-constrained systems. DTAM\cite{newcombe2011dtam}, an early dense SLAM system, employs photometric consistency across pixels to track the camera, leveraging multi-view stereo constraints to update the dense scene model and represent it as a cost volume. KinectFusion\cite{newcombe2011kinectfusion} employs RGB-D cameras to facilitate real-time camera pose estimation and scene geometry updates through iterative closest point (ICP) and TSDF-Fusion\cite{curless1996volumetric}. Subsequent studies have introduced a range of data structures to enhance the scalability of SLAM systems, including VoxelHash\cite{niessner2013real,kahler2015very,chen2013scalable} and Octrees\cite{zeng2013octree,vespa2018efficient}.

Although classical SLAM techniques offer real-time performance and efficiency, they often fall short of the level of detail and visual fidelity required to accurately represent complex environments. Furthermore, they are inadequate for achieving realistic scene reconstruction in advanced applications.

\begin{figure*}
    \centering
    \includegraphics[width=1\linewidth]{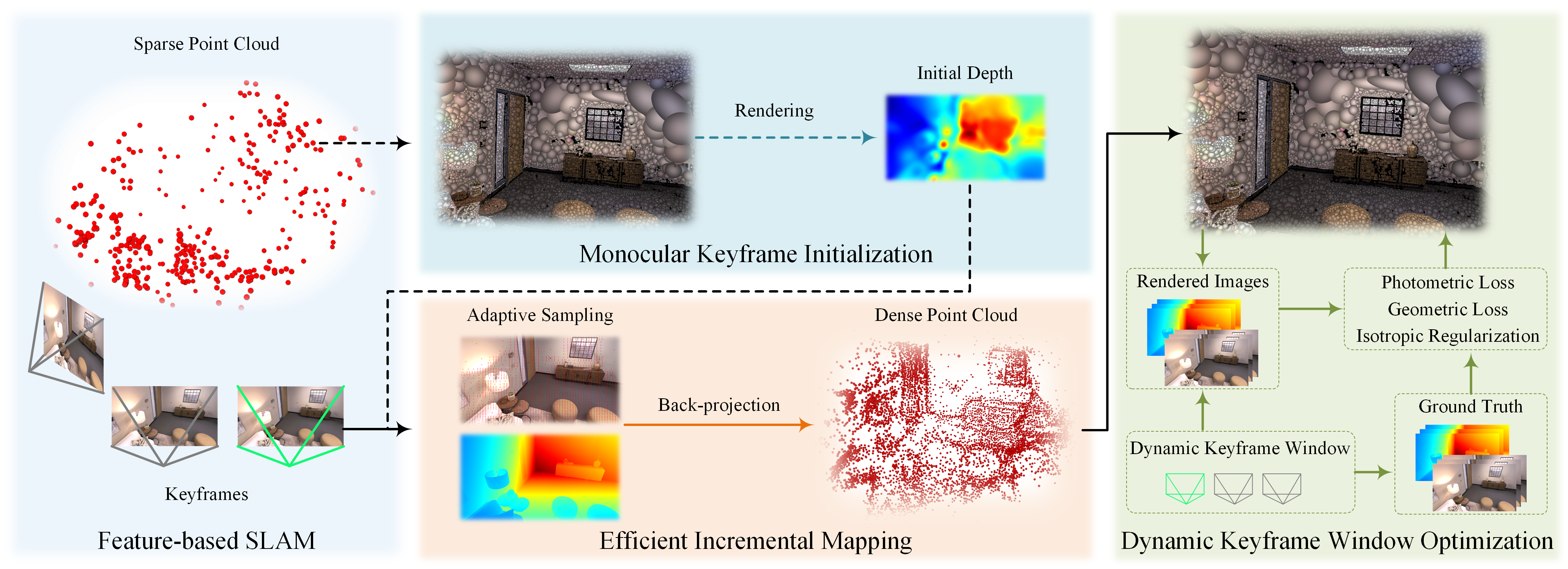}
    \caption{Overview of our RP-SLAM. Keyframes and sparse point cloud are provided the feature-based SLAM, where the point cloud are used for monocular keyframe initialization. Afterwards, the dense point cloud is obtained by efficient incremental mapping, which in turn initializes the Gaussian primitives. Finally, the scene representation is optimized by dynamic keyframe window, where the geometric loss is used only for the RGB-D case. The dashed arrows are for the monocular case only}
    \label{overview}
\end{figure*}

\subsection{Neural Implicit Visual SLAM}
iNeRF\cite{yen2021inerf} represents a pioneering attempt to demonstrate that camera pose regression is feasible within a trained NeRF model for a specific scene. However, rather than being a complete SLAM system, iNeRF addresses a localization task based on a existing model. Barf\cite{lin2021barf} further presents a method to simultaneously fit NeRF while refining camera poses, even with an initially rough estimate, by establishing a theoretical bridge between classical image alignment and joint alignment and reconstruction within neural radiance fields. iMAP\cite{sucar2021imap}, on the other hand, introduces the application of NeRF to reconstruct accurate 3D geometry from RGB-D images without requiring initial pose information. By using a single MLP to represent the global scene, iMAP jointly optimizes both the scene map and camera poses. However, this single MLP architecture struggles to capture fine geometric details and to scale effectively in larger environments. NICE-SLAM\cite{zhu2022nice} addresses this scalability limitation by partitioning the world coordinate frame into uniform grids, thereby enhancing both inference speed and accuracy. NeRF-SLAM\cite{rosinol2023nerf} integrates Droid-SLAM\cite{teed2021droid} with Instant-NGP\cite{muller2022instant}, utilizing Droid-SLAM for camera pose estimation, dense depth mapping, and uncertainty measurement, and leveraging this data to optimize Instant-NGP's scene representation. GO-SLAM\cite{zhang2023go} further improves global scene consistency by introducing loop closure and global bundle adjustment, while Co-SLAM\cite{wang2023co} achieves high-quality scene reconstruction by combining coordinate encoding with sparse grid representations.

Despite the significant potential of NeRF in the generation of realistic scenes, a number of challenges emerge when integrating this approach into SLAM systems. These include the high computational cost, the prolonged training time required, and the vulnerability to catastrophic forgetting.

\subsection{3D Gaussian Splatting Visual SLAM}
3DGS\cite{kerbl3Dgaussians} has rapidly gained interest within the field of SLAM research due to its rapid rendering capabilities and explicit scene representation. MonoGS\cite{matsuki2024gaussian} and SplaTAM\cite{keetha2024splatam} represent foundational efforts in coupled GS-SLAM algorithms, pioneering an approach that jointly optimizes Gaussian primitives and camera poses through gradient backpropagation. The concept of sub-maps is introduced by Gaussian-SLAM\cite{yugay2023gaussian} as a means of mitigating the issue of catastrophic forgetting. LoopSplat\cite{zhu2024loopsplat} builds upon the foundations of Gaussian-SLAM\cite{yugay2023gaussian} through the utilization of a Gaussian splatting-based loop closure strategy, thereby enhancing the accuracy of poses through the implementation of improved registration techniques. However, a significant obstacle with these methods is that the time-consuming process of estimating the camera pose for each frame using 3DGS places a considerable computational burden, which limits its ability to achieve real-time performance. In order to address this challenge, decoupled 3DGS-based SLAM approaches have been introduced. Splat-SLAM\cite{sandstrom2024splat} and IG-SLAM\cite{sarikamis2024ig} employ pre-trained dense bundle adjustment\cite{teed2021droid} for the purpose of camera pose tracking, while simultaneously utilising proxy depth maps for optimizing map representation. RTG-SLAM\cite{peng2024rtg} integrates frame-to-model ICP\cite{newcombe2011kinectfusion} for pose tracking and employ the opaque Gaussian primitives for depth rendering. GS-ICP-SLAM\cite{ha2025rgbd} attains remarkable speeds by capitalising on the shared covariance  between G-ICP\cite{segal2009generalized} and 3DGS. Photo-SLAM\cite{huang2024photo} and CaRtGS\cite{feng2024cartgs} deploy ORB-SLAM3\cite{campos2021orb} for tracking and incorporates a coarse-to-fine approach in map optimization, thereby enhancing system robustness and overall performance. For incremental mapping, these methods often rely on dense pixel sampling to initialize new Gaussian primitives, resulting in redundancy and high storage costs. To optimize scene representation, a common approach selects a fixed number of keyframes, merging them with the latest keyframe into a fixed keyframe window for map optimization. However, this strategy risks local minima and forgetting issues over iterations.

In the monocular case, the lack of depth information complicates the accurate addition of new primitives. MonoGS\cite{matsuki2024gaussian} initializes primitives using random depths without a geometric foundation, relying entirely on the mapping process for optimization, which hinders precise scene representation. Photo-SLAM\cite{huang2024photo} and CaRtGS\cite{feng2024cartgs} adopt the spatial gradient-based densification method from the original 3DGS\cite{kerbl3Dgaussians} to add primitives. However, the limited iterations fail to optimize these primitives, leading to artifacts caused by mismatches between the geometric and photometric properties of the Gaussian primitives.

\begin{figure*}
  \centering
  \scriptsize
  \setlength{\tabcolsep}{0.3pt}
  \newcommand{\sz}{0.33}  %
  \begin{tabular}{ccc}
    \makecell{\includegraphics[width=\sz\linewidth]{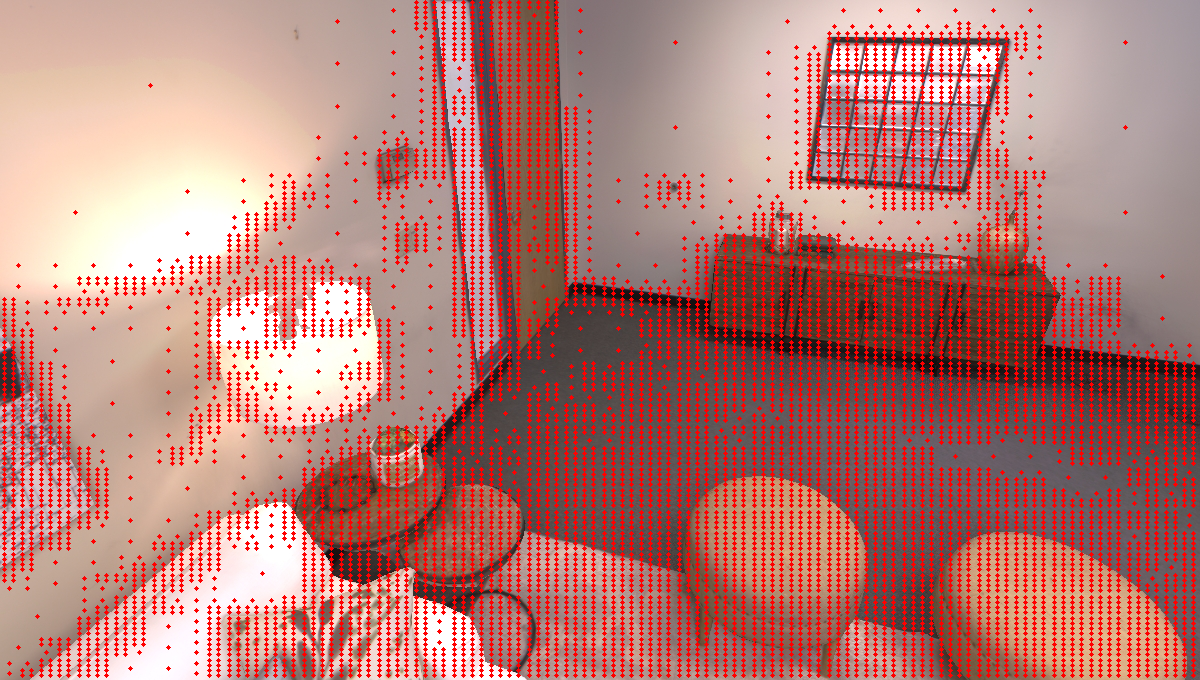}}   &
    \makecell{\includegraphics[width=\sz\linewidth]{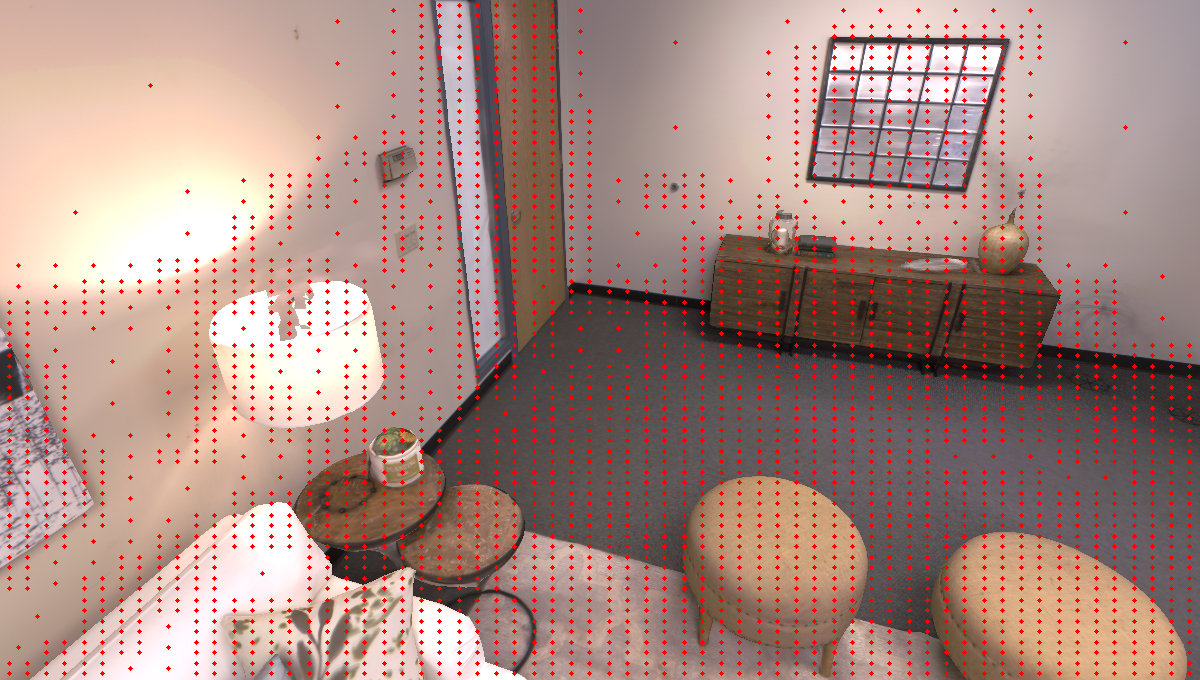}}   &
    \makecell{\includegraphics[width=\sz\linewidth]{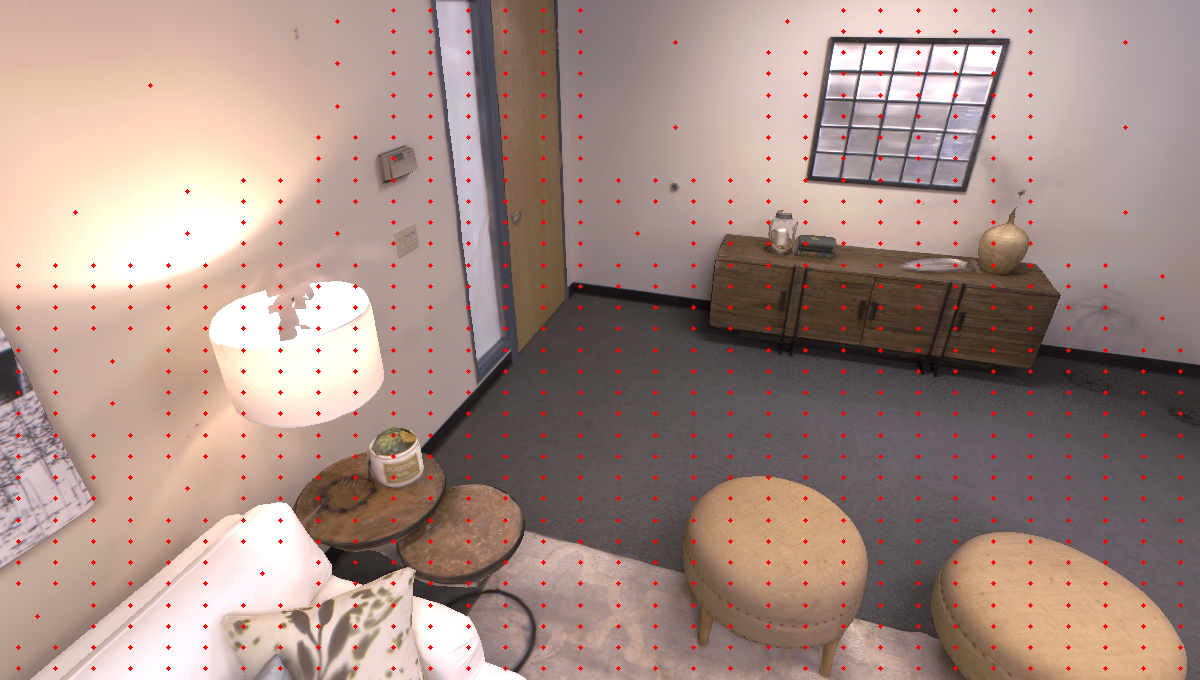}}    \\
    \vspace{1mm}
    \makecell{(a) $c=4$. No. of sampling points: 9898.}                                         &
    \makecell{(b) $c=8$. No. of sampling points: 2872.}                                         &
    \makecell{(c) $c=16$. No. of sampling points: 826.}                                                \\
  \end{tabular}
  \vspace{-1mm}
  \caption{Quadtree-based adaptive image sampling guided by image local gradients at different minimum cell sizes: 4, 8, 16. The method is capable of adaptively focusing sampling on regions that are rich in texture. A smaller minimum cell size allows for a more detailed sampling, but this is accompanied by an increased need for processing of the resulting data.}
  \label{sampling}
  \vspace{-5pt}
\end{figure*}

\section{Methods}
The overview of RP-SLAM is illustrated in Fig.~\ref{overview}. Feature-based SLAM is primarily utilized for camera poses estimation, the generation of keyframes and their covisibility, as well as the provision of a sparse initial point cloud for monocular mode. Then, the received keyframes is used for efficient incremental mapping to generate Gaussian primitives. The Gaussian primitives are optimized using the dynamic keyframe window based on the covisibility. In the monocular mode, sparse point cloud must be employed for monocular keyframe initialization.

\subsection{3D Gaussian Splatting}
In our method, 3DGS is employed for the purpose of mapping the scene, which comprises a set of Gaussian primitives $\mathcal{G}$ and renders photorealistic images by tile-based rasterisation. Each Gaussian primitive $G_i$ is characterised by both optical and geometric attributes. The former include color $c_i$ and opacity $\alpha_i$, whereas the latter are defined in the world coordinate, comprising a mean $\bm{\mu}_i^w$ indicative of its position and a covariance $\bm{\Sigma}_i^w$ indicative of its ellipsoidal shape. And the covariance $\bm{\Sigma}_i^w$ is decomposed into a scale vector $\bm{s}_i$ and a quaternion $\bm{q}_i$. Given $n$ Gaussians, the color of a pixel can be obtained by rendering:
\begin{equation}
    C=\sum_{i=1}^n c_i\alpha_i\prod_{j=1}^{i-1}(1-\alpha_j).
\end{equation}
Following\cite{matsuki2024gaussian}, per-pixel depth can also be rasterised by alpha-blending:
\begin{equation}
    D=\sum_{i=1}^n z_i\alpha_i\prod_{j=1}^{i-1}(1-\alpha_j),
    \label{eq_depth_ren}
\end{equation}
where $z_i$ is the distance to the mean $\mu_i^w$ of Gaussian $G_i$ along the camera ray.

In contrast to marching along the camera rays, the rasterization process iterates over the Gaussians. As such, free space is disregarded during the rendering process. The contributions of $\alpha$ are decayed during rasterization by the Gaussian functions based on the 2D Gaussians $\mathcal{N}(\bm{\mu},\bm{\Sigma})$ formed by splatting the 3D Gaussians $\mathcal{N}(\bm{\mu}_w,\bm{\Sigma}_w)$:
\begin{equation}
    \begin{split}
        &\bm{\mu}=\pi(\bm{T}_{cw}\bm{\mu}_w),\\
        &\bm{\Sigma}=\bm{J}\bm{R}\bm{\Sigma}_w\bm{R}^\top\bm{J}^\top,
    \end{split}
\end{equation}
where $\pi$ is the projection operator, and $\bm{T}_{cw}$ represents the camera pose in the world. $\bm{J}$ refers to the Jacobian of the projective transformation, and $\bm{R}$ is the rotation of $\bm{T}_{cw}$. This formulation makes the 3D Gaussian differentiable, and the blending operation provides the Gaussian with a gradient flow.

To optimize the scene representation, photometric loss $\mathcal{L}_{pho}$ and geometric loss $\mathcal{L}_{geo}$ are considered, where geometric loss is only used for RGB-D case. Following MonoGS\cite{matsuki2024gaussian}, isotropic regularization $\mathcal{L}_{iso}$ is introduced with the aim of reducing the generation of artifacts by penalize the scaling parameters $\bm{s}$. As a result, the final monocular loss $\mathcal{L}_{mono}$ and RGB-D loss $\mathcal{L}_{RGB-D}$ are as follows:
\begin{equation}
\begin{split}
    &\mathcal{L}_{pho}=||I(\mathcal{G},T_{cw})-\bar{I}||_1,\\
    &\mathcal{L}_{geo}=||D(\mathcal{G},T_{cw})-\bar{D}||_1,\\
    &\mathcal{L}_{iso}=\sum_{i=1}^{|\mathcal{G}|}||\bm{s}_i-\tilde{\bm{s}_i}\cdot\bm{1} ||_1,\\
    &\mathcal{L}_{mono}=\mathcal{L}_{pho}+\lambda_{iso}\mathcal{L}_{iso}.\\
    &\mathcal{L}_{RGB-D}=\lambda_{pho} \mathcal{L}_{pho}+(1-\lambda_{pho})\mathcal{L}_{goe}+\lambda_{iso}\mathcal{L}_{iso}.
\end{split}
    \label{loss}
\end{equation}
where $I(\mathcal{G},T_{cw})$ and $D(\mathcal{G},T_{cw})$ are the rendered rgb image and depth map from the given Gaussian primitives $\mathcal{G}$ and camera pose $T_{cw}$, and $\bar{I}$ and $\bar{D}$ are the ground truth rgb image and depth map. $\tilde{\bm{s}_i}$ is the mean of $\bm{s}_i$. $\lambda_{pho}$ and $\lambda_{iso}$ are the hyperparameters.

\begin{figure*}
  \centering
  \scriptsize
  \setlength{\tabcolsep}{0.3pt}
  \newcommand{\sz}{0.33}  %
  \begin{tabular}{cccc}
    \makecell{\includegraphics[width=\sz\linewidth]{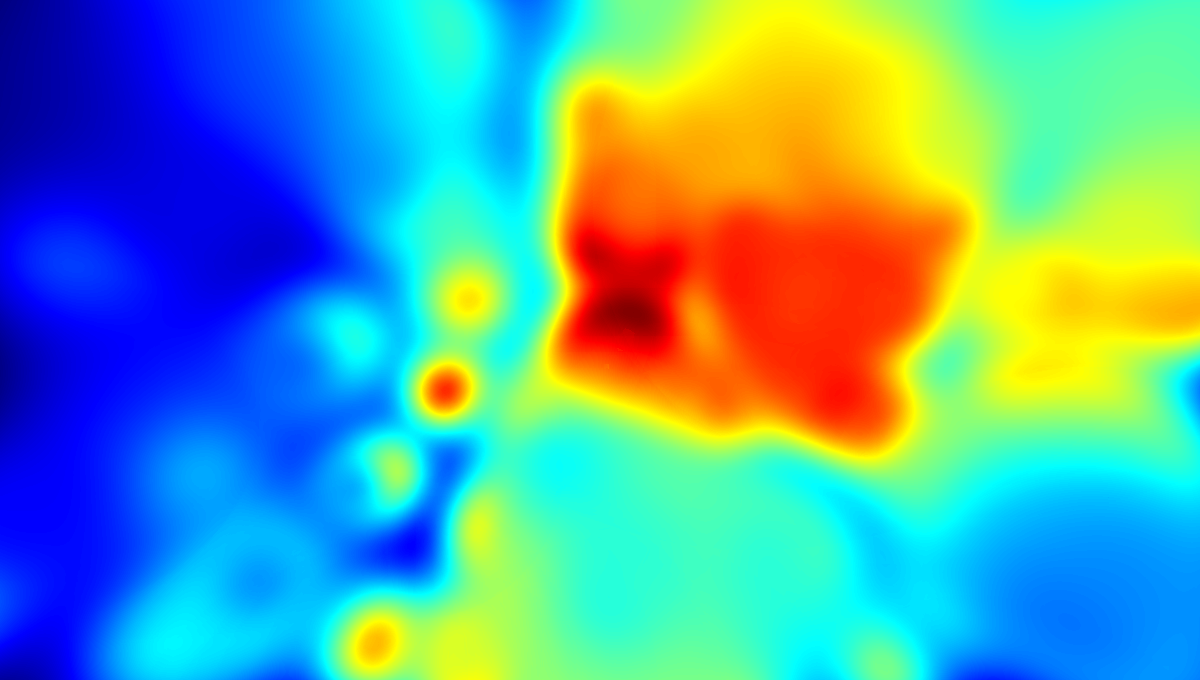}}   &
    \makecell{\includegraphics[width=\sz\linewidth]{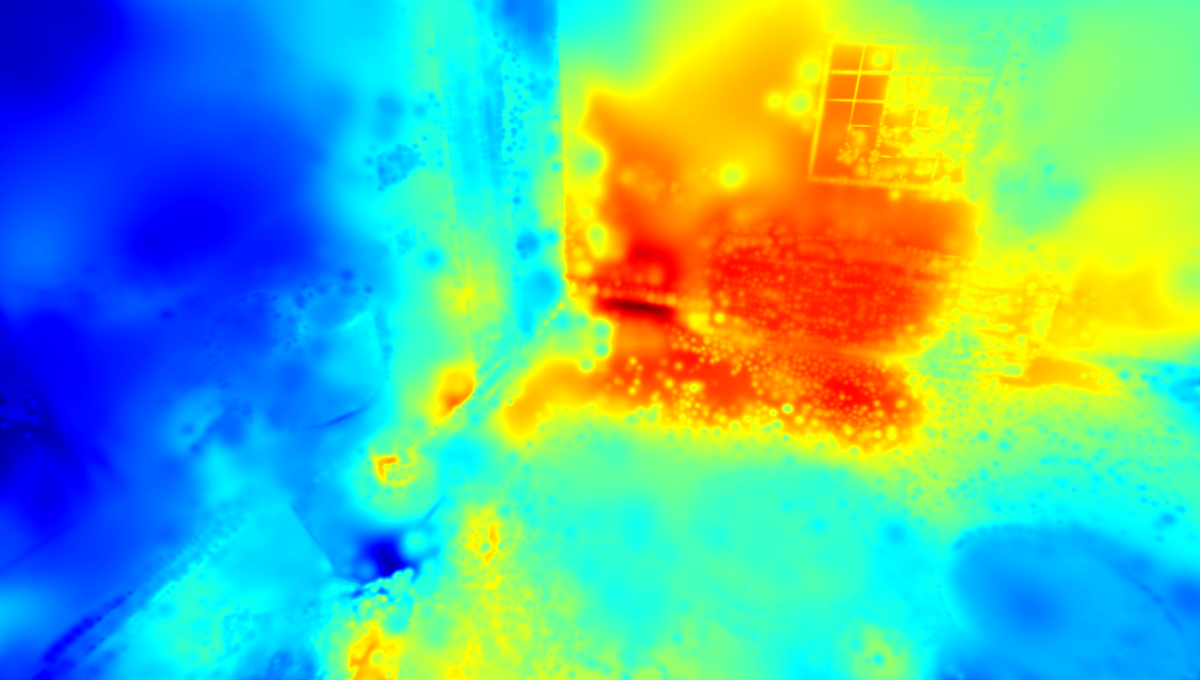}}   &
    \makecell{\includegraphics[width=\sz\linewidth]{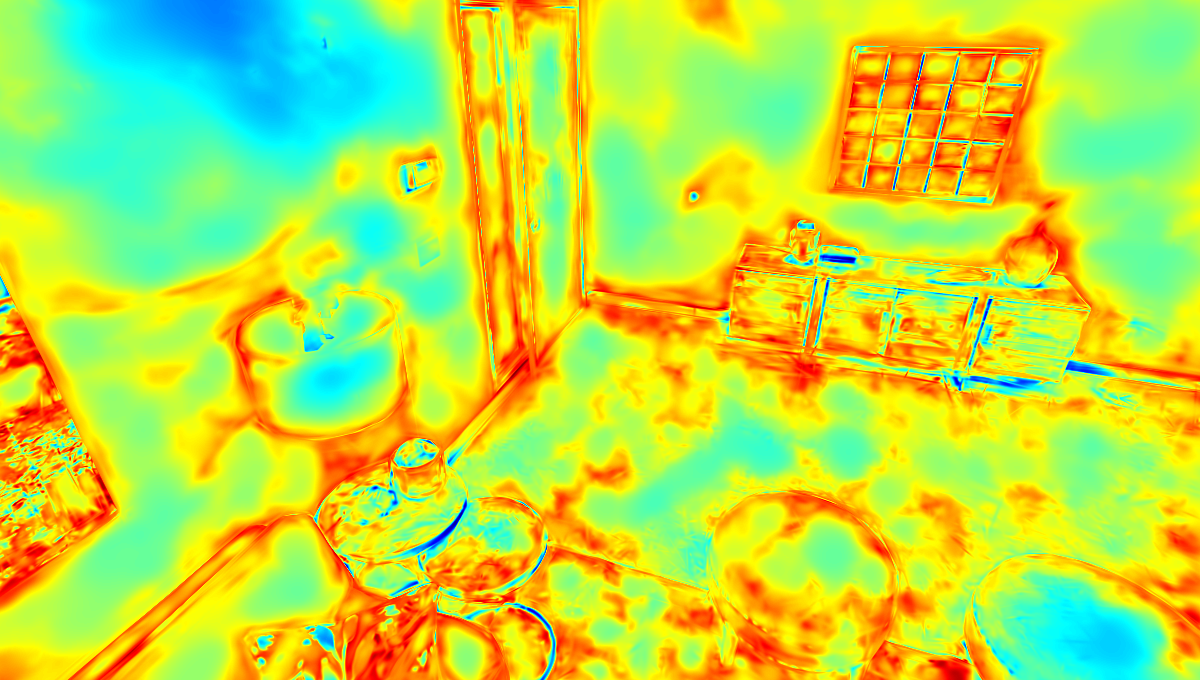}}        \\
    \vspace{1mm}
    \makecell{(a)}                                         &
    \makecell{(b)}                                         &
    \makecell{(c)}                                                \\
  \end{tabular}
  \vspace{-1mm}
  \caption{Rendered depths in the monocular case. (a) Initial depth obtained by our RP-SLAM from a sparse point cloud, which describes the initial geometry at the viewpoint. (b) Depth obtained by initial iterations using the depth of (a) based on the dense point cloud obtained in Sec.~\ref{mapping}. (c) Depth obtained according to MonoGS's\cite{matsuki2024gaussian} monocular initial method. Following preliminary iterations, the Gaussian primitives obtained by our method in the monocular case have been found to describe the scene structure with reasonable accuracy. This is in comparison to the result obtained by MonoGS\cite{matsuki2024gaussian}, which is less satisfactory in this regard.}
  \label{depth}
  \vspace{-5pt}
\end{figure*}

\subsection{Efficient Incremental Mapping}
\label{mapping}
The efficient scene representation in 3DGS-based SLAM is reliant upon a meticulous balance between the number of map parameters (number of Gaussian primitives) and map fidelity. The majority of 3DGS-based methods frequently encounter difficulties in maintaining this balance, as uniformly dense sampling may result in superfluous computation and storage of simple regions, whereas sparse sampling may prove inadequate for capturing crucial details in complex textures. In order to address this challenge, we propose the efficient incremental mapping which consists of quadtree-based adaptive image sampling and KNN-based Gaussian primitive filtering.  

For each keyframe from feature-based SLAM, the quadtree-based adaptive image sampling is implemented through a recursive subdivision of this keyframe into smaller quadtree cells. Subdivision of the image continues until the adaptive minimum cell size $c_{th}=\eta c$ or the variance in the gradient of each cell falls below an adaptive threshold $\tau_{th}=\eta \tau$, with each cell evaluated in turn. Here, $\eta=\frac{\sqrt{H\times W}}{512}$, $c$ and $\tau$ are the predefined thresholds, and $H$ and $W$ are the image sizes. Formally, the gradient variance $V$ of a cell $C$ is defined as:
\begin{equation}
    V(C)=\frac{1}{\left\lvert C \right\rvert}\sum_{p\in C}(\nabla I(p)-\mu_C)^2,
\end{equation}
where $\nabla I(p)$ is the gradient magnitude at pixel $p$, and $\mu_C$ is the mean gradient magnitude within cell $C$. Subdivision occurs if $V(C)>\tau$. Fig.~\ref{sampling} depicts the sampling of the image at different minimum cell sizes. The quadtree-based adaptive image sampling dynamically adjusts the sampling density based on the gradient information of the input image. This method guarantees that image sampling is concentrated on regions of high structural complexity, thereby ensuring an efficient sampling process.

Once the sampling pixels are selected, they are back-projected into 3D space using the camera pose and intrinsic parameters, resulting in a set of candidate Gaussian primitives. For a sampled pixel $p$ in the image, the corresponding 3D position $P$ in the world coordinate frame is computed as:
\begin{equation}
    P=T_{wc}\pi^{-1}(p,d),
\end{equation}
where $\pi^{-1}$ represents the inverse projection using the corresponding depth $d$, and $T_{wc}$ is the  transformation from camera to world coordinate frame. To circumvent redundancy and guarantee an efficacious map representation, KNN-based filtering step is implemented for the newly generated Gaussian primitives. For each candidate primitive, the method assesses its spatial proximity to existing primitives in the map. Specifically, 
For a newly generated Gaussian primitive $G_{new}$, a KNN\cite{ravi2020pytorch3d} search is performed to find its closest neighbors $\mathcal{G}=\{G_1,G_2,...,G_k\}$ and $k=3$ in our setting. The new Gaussian $G_{new}$ is considered redundant if the distance $dis(G_{new},G_i)$ to every neighbor $G_i$ satisfies:
\begin{equation}
    dis(G_{new},G_i)<\lambda r_i \ \  \forall G_i \in \mathcal{G},
\end{equation}
where $\lambda$ is a predefined scaling factor that adjusts the effective influence of the neighbor's radius, $r_i$, which is the minimum value in the scale vector $s_i$ corresponding to the Gaussian primitive $G_i$. If this condition is met, $G_{new}$ is deemed redundant and discarded. Otherwise, $G_{new}$ is added to the map as a new Gaussian primitive. The KNN-based Gaussian primitive filtering eliminates redundant primitives based on local Gaussian primitive relations, thereby ensuring an efficient scene representation and, consequently, the efficient utilization of computational and storage resources.

By combining quadtree-based adaptive sampling and KNN-based filtering, the proposed method achieves efficient incremental mapping. Quadtree-based adaptive sampling ensures that computational resources are concentrated on regions with significant structural details, while KNN-based filtering eliminates redundant Gaussians, maintaining a compact and non-redundant map representation.

\subsection{Dynamic Keyframe Window}
\label{window}
The addition of a new keyframe typically signifies the necessity for optimization in previously unexplored areas. The prevailing methods entail the selection of a specific number of keyframes from the keyframe set, which are then incorporated into a fixed window of keyframes with the newly added keyframe. Only those keyframes situated within this fixed window are employed for map optimization. Nevertheless, during the iterative process, this approach may give rise to severe forgetting and overfitting issues, ultimately leading to a deterioration in the quality of the final map representation.

To address this issue, we propose the optimization of a dynamic keyframe window at each iteration. Specifically, for each new keyframe $K_{new}$ generated by feature-based SLAM, the system categorizes all keyframes $\mathcal{K}$ into two distinct sets: co-visible keyframes $\mathcal{K}_{1}$ and other keyframes $\mathcal{K}_{2}$. This categorization is based on the degree of visibility of other keyframes in relation to the current keyframe $K_{new}$. Co-visible keyframes $\mathcal{K}_{1}$ are defined as keyframes that observe the same or near-neighbouring regions as the current keyframe. In contrast, other keyframes $\mathcal{K}_{2}$ represent those that do not observe the same region as the current keyframe. At each iteration, a random selection of $k_1$ and $k_2$ keyframes is made from the two sets, respectively. These keyframes are then combined with the current keyframe, thus forming a dynamic keyframe window $\mathcal{W}$:
\begin{equation}
    \begin{split}
        &\mathcal{W}=\{K_{new}\}\cup\mathcal{S}_1\cup\mathcal{S}_2,\\
        &\mathcal{S}_1\subseteq\mathcal{K}_1, \ \ |\mathcal{S}_1|=k_1,\\
        &\mathcal{S}_2\subseteq\mathcal{K}_2, \ \ |\mathcal{S}_2|=k_2.\\
    \end{split}
\end{equation}
Subsequently, the dynamic keyframe window is employed for the purpose of optimizing the map representation, specifically the parameters of the Gaussian primitives.

This approach ensures that the optimization process is focused on both local consistency, with consideration given to keyframes that are closely related to $K_{new}$, and global consistency, with the inclusion of non-co-visible keyframes that are critical for maintaining long-term map accuracy. By dynamically adapting the keyframe set, this method prevents overfitting to the newly added keyframe and mitigates forgetting of previously observed regions, thereby ensuring a more balanced and robust map optimization process.

\begin{table}[t]
  \centering
  \caption{Quantitative results for Replica\cite{straub2019replica} dataset in the monocular case. MS denotes model size (Mb)}
  \footnotesize
  \setlength{\tabcolsep}{0.15em}
  \begin{tabular}{c c *{9}{c}}
    \toprule
    Methods  &Metrics&\texttt{o0}&\texttt{o1} & \texttt{o2} & \texttt{o3} & \texttt{o4} & \texttt{r0} & \texttt{r1} & \texttt{r2} & Avg.             \\
    \midrule
    \multirow{6}{*}{\makecell{MonoGS \\ \cite{matsuki2024gaussian}}}
            &PSNR$\uparrow$      &30.58 &32.49	&32.65 &26.34 &24.69 &23.76	&24.99 &23.57 &27.38         \\
            &SSIM$\uparrow $     &0.892 &0.904	&0.836 &0.851 &0.871 &0.755	&0.785 &0.833 &0.841         \\
            &LPIPS$\downarrow$   &0.249 &0.191	&0.302 &0.22  &0.315 &0.322 &0.352 &0.353 &0.288            \\
            &ATE$\downarrow$     &25.4	&24.8	&48.9  &11.8  &61.6	 &13.6	&55.4  &26.1  &33.45            \\
            &FPS$\uparrow$       &1.4	&1.6	&1.3   &1.3   &1.3   &1.3   &1.3   &1.3   &1.5\\
            &MS$\downarrow$      &\fs{6.5}	&\fs{4.6}	&\fs{7.2}  &\fs{8.4}  &\fs{5.9}	 &\fs{8.5}	&\fs{7.3}  &\fs{5.2}  &\fs{6.7}\\
    \midrule
    \multirow{6}{*}{\makecell{Photo-\\SLAM\\ \cite{huang2024photo}}}
            &PSNR$\uparrow$      &\nd{36.98} &\nd{37.59} &31.79      &31.62	     &34.16      &29.77      &31.3	     &33.18	     &33.30\\
            &SSIM$\uparrow $     &\nd{0.955} &\nd{0.950} &0.929      &0.920      &0.941      &0.871      &0.911	     &0.934      &0.927\\
            &LPIPS$\downarrow$   &\fs{0.061} &0.062	     &0.091      &0.086	     &0.072      &0.106      &0.083      &0.067	     &0.079\\
            &ATE$\downarrow$     &\nd{0.32}  &0.45	     &2.53       &\nd{0.39}  &0.61       &0.43       &0.68       &\nd{0.27}	 &0.71\\
            &FPS$\uparrow$       &\nd{35.9}  &\nd{35.4}  &\fs{34.1}  &\fs{34.8}  &\nd{36.1}  &\nd{34.1}  &\nd{36.4}  &\fs{34.9}  &\nd{35.2}\\
            &MS$\downarrow$      &19.1	     &18.4       &25.4       &18.4       &19.8       &26.8       &32.1       &21.1	     &23.14\\
    \midrule
    \multirow{6}{*}{\makecell{CaRtGS \\ \cite{feng2024cartgs}}}
            &PSNR$\uparrow$      &35.49	     &36.22	     &\nd{33.54} &\fs{32.82} &\nd{35.31} &\nd{31.92} &\nd{32.44} &\nd{34.53} &\nd{34.03}\\
            &SSIM$\uparrow $     &0.946	     &0.947	     &\nd{0.938} &\nd{0.936} &\fs{0.944} &\nd{0.913} &\nd{0.915} &\fs{0.953} &\nd{0.934}\\
            &LPIPS$\downarrow$   &0.078      &\fs{0.052} &\nd{0.076} &\nd{0.077} &\fs{0.061} &\nd{0.074} &\nd{0.071} &\fs{0.051} &\nd{0.068}\\
            &ATE$\downarrow$     &\fs{0.21}  &\fs{0.23}  &\fs{1.31}  &\fs{0.13}  &\fs{0.18}	 &\fs{0.17}  &\fs{0.42}	 &\fs{0.19}	 &\fs{0.36} \\
            &FPS$\uparrow$       &\fs{36.5}  &\fs{36.4}  &\nd{33.6}  &\nd{34.7}  &\fs{36.9}  &\fs{34.5}  &\fs{37.5}  &\nd{34.6}  &\fs{35.6}\\
            &MS$\downarrow$      &\nd{9.9}	 &11.4  &16.1  &14.1  &13.4	 &23.6  &18.4	 &15.5	 &15.3\\
    \midrule
    \multirow{6}{*}{\makecell{\textbf{RP-SLAM} \\ (ours)}}
            &PSNR$\uparrow$      &\fs{37.74} &\fs{40.35} &\fs{34.29} &\nd{32.78} &\fs{35.92} &\fs{31.95} &\fs{33.61} &\fs{35.85} &\fs{35.31}\\
            &SSIM$\uparrow $     &\fs{0.958} &\fs{0.961} &\fs{0.941} &\fs{0.937} &\nd{0.942} &\fs{0.921} &\fs{0.925} &\nd{0.935} &\fs{0.940}\\
            &LPIPS$\downarrow$   &\nd{0.062} &\nd{0.058} &\fs{0.071} &\fs{0.075} &\nd{0.065} &\fs{0.071} &\fs{0.067} &\nd{0.062} &\fs{0.067}\\
            &ATE$\downarrow$     &0.35	     &\nd{0.31}  &\nd{2.24}  &0.44       &\nd{0.57}	 &\nd{0.35}  &\nd{0.46}	 &0.38	     &\nd{0.64}\\
            &FPS$\uparrow$       &18.0	     &18.3       &16.5	     &17.2       &17.8	     &16.3	     &18.2	     &17.0       &17.3\\
            &MS$\downarrow$      &11.8	     &\nd{8.1}	 &\nd{12.3}  &\nd{10.9}	 &\nd{11.8}	 &\nd{12.7}	 &\nd{11.9}	 &\nd{11.1}	 &\nd{11.3}\\
    \bottomrule
  \end{tabular}
  \label{mono_replica}
\end{table}

\begin{table}[t]
  \centering
  \caption{Quantitative results for Replica\cite{straub2019replica} dataset in the RGB-D case. MS denotes model size (Mb).}
  \footnotesize
  \setlength{\tabcolsep}{0.15em}
  \begin{tabular}{*{11}{c}}
    \toprule
    Methods  &Metrics&\texttt{o0}&\texttt{o1} & \texttt{o2} & \texttt{o3} & \texttt{o4} & \texttt{r0} & \texttt{r1} & \texttt{r2} & Avg.             \\
    \midrule
    \multirow{6}{*}{\makecell{MonoGS \\ \cite{matsuki2024gaussian}}}
    &PSNR$\uparrow$      &\nd{40.08} &\fs{41.22} &35.57      &\nd{34.34} &33.74 &\nd{32.78} &\nd{35.87} &36.61     &\nd{36.78}         \\
    &SSIM$\uparrow $     &0.971      &0.974      &0.956      &\nd{0.952} &0.936 &0.934	    &0.953      &0.972     &0.956         \\
    &LPIPS$\downarrow$   &\nd{0.058} &0.049      &\nd{0.065} &0.063      &0.099 &0.074      &0.081      &0.071     &0.070            \\
    &ATE$\downarrow$     &0.43	     &0.51       &\nd{0.23}  &\fs{0.21}  &2.36	&0.44	    &0.26       &0.35      &0.60            \\
    &FPS$\uparrow$       &1.2	     &1.3        &1.1        &1.0        &1.1   &1.1        &1.1        &1.0       &1.1\\
    &MS$\downarrow$      &16.2	     &15.2       &26.6       &24.7       &23.1	&\nd{25.2}	&\nd{18.5}  &\nd{22.1} &21.5\\
    \midrule
    \multirow{6}{*}{\makecell{Photo-\\SLAM\\ \cite{huang2024photo}}}
    &PSNR$\uparrow$      &38.47	     &39.08        &33.03	   &33.78	   &\nd{36.02} &30.71	   &33.51	   &35.12	   &34.97  \\
    &SSIM$\uparrow $     &0.964	     &0.961	       &0.938      &0.938	   &0.952	   &0.899	   &0.934	   &0.951      &0.942  \\
    &LPIPS$\downarrow$   &0.05	     &\nd{0.047}   &0.077	   &0.066	   &\nd{0.054} &0.075	   &\nd{0.057} &\nd{0.043} &0.059   \\
    &ATE$\downarrow$     &0.48	     &0.45         &1.32	   &0.79	   &0.57	   &0.49	   &0.45	   &0.29	   &0.61   \\
    &FPS$\uparrow$       &\fs{32.3}  &\nd{32.2}    &\nd{29.2}  &{\fs28.8}  &\fs{28.2}  &\fs{28.2}  &\fs{29.6}  &\fs{26.3}  &\fs{29.3}\\
    &MS$\downarrow$      &19.9	     &20.1         &31.4	   &24.3	   &30.9	   &51.8	   &39.7	   &38.2	   &32.0\\
    \midrule
    \multirow{6}{*}{\makecell{SplaTAM\\ \cite{keetha2024splatam}}}
    &PSNR$\uparrow$      &37.97	    &38.95     &32.92	   &29.81	 &31.96 &32.56	     &33.61	    &35.11	&34.11         \\
    &SSIM$\uparrow $     &0.981  	&0.981     &0.966      &0.951	 &0.948 &\fs{0.955}  &0.961	    &0.980	&0.965        \\
    &LPIPS$\downarrow$   &0.089 	&0.097	   &0.098	   &0.117	 &0.155 &\nd{0.071}  &0.097	    &0.075	&0.100         \\
    &ATE$\downarrow$     &\fs{0.41} &\nd{0.25} &0.32	   &0.35	 &0.57	&\nd{0.29}	 &0.51	    &0.33	&\nd{0.38}         \\
    &FPS$\uparrow$       &0.2   	&0.2       &0.3 	   &\nd{0.3} &0.2	&0.2	     &0.2	    &0.2	&0.2\\
    &MS$\downarrow$      &323.6 	&294.2     &237	       &273.3    &276.7 &265.8       &348.7	    &298.4	&289.7\\
    \midrule
    \multirow{6}{*}{\makecell{RTG-\\SLAM\\ \cite{peng2024rtg}}}
    &PSNR$\uparrow$      &39.09      &39.22	     &\nd{33.45} &33.33	&35.53  	&30.79	    &34.52	    &35.65	    &35.07         \\
    &SSIM$\uparrow $     &\fs{0.987} &\fs{0.986} &\fs{0.981} &0.951	&\fs{0.984}	&0.945   	&\fs{0.977}	&\nd{0.981} &\fs{0.974}        \\
    &LPIPS$\downarrow$   &0.098      &0.135	     &0.075	     &0.298	&0.116	    &0.154	    &0.131	    &0.137	    &0.143         \\
    &ATE$\downarrow$     &\fs{0.15}	 &\fs{0.14}  &\fs{0.22}	 &0.26	&\fs{0.25}	&\fs{0.21}	&\fs{0.19}	&0.12	    &\fs{0.19}         \\
    &FPS$\uparrow$       &8.3        &8.4        &8.1	     &7.9	&8.1	    &7.9	    &7.8	    &7.8	    &8.0\\
    &MS$\downarrow$      &51.2	     &56.5       &53.6	     &63.3	&62.4	    &61.9	    &78.3	    &66.8	    &61.8\\
    \midrule
    \multirow{6}{*}{\makecell{CaRtGS \\ \cite{feng2024cartgs}}}
    &PSNR$\uparrow$      &35.52	     &37.85	      &33.58      &34.04	  &35.34	  &30.02	  &33.58	  &\nd{36.65} &34.57  \\
    &SSIM$\uparrow $     &0.953	     &0.955	      &0.946      &0.943	  &0.956	  &0.847	  &0.936	  &0.964	  &0.938  \\
    &LPIPS$\downarrow$   &0.061	     &\fs{0.044}  &0.066	  &\nd{0.061} &\fs{0.051} &0.072	  &\fs{0.058} &\fs{0.043} &\nd{0.057}   \\
    &ATE$\downarrow$     &0.49  	 &0.39	      &1.03       &0.42	      &\nd{0.47}  &0.32	      &0.36	      &\fs{0.19}  &0.46          \\
    &FPS$\uparrow$       &\nd{31.5}  &\fs{32.5}   &\fs{29.8}  &\nd{28.6}  &\nd{27.1}  &\nd{27.8}  &\nd{29.5}  &\nd{26.1}  &\nd{29.1}\\
    &MS$\downarrow$      &\nd{10.3}	 &\nd{14.2}	  &\nd{20.1}  &\nd{16.6}  &\nd{20.2}  &29.9	      &25.3       &27.9	      &\nd{20.6}\\
    \midrule
    \multirow{6}{*}{\makecell{\textbf{RP-SLAM} \\ (ours)}}
    &PSNR$\uparrow$      &\fs{41.26} &\nd{41.01} &\fs{36.27} &\fs{34.98} &\fs{36.05} &\fs{33.74} &\fs{36.71} &\fs{36.96} &\fs{37.12}  \\
    &SSIM$\uparrow $     &\nd{0.982} &\nd{0.985} &\nd{0.975} &\fs{0.972} &\nd{0.958} &\nd{0.951} &\nd{0.963} &\fs{0.982} &\nd{0.971}         \\
    &LPIPS$\downarrow$   &\fs{0.055} &0.057      &\fs{0.064} &\fs{0.048} &\fs{0.051} &\fs{0.064} &0.062 	 &0.049	     &\fs{0.056}    \\
    &ATE$\downarrow$     &0.43  	 &0.38   	 &0.53	     &0.36       &0.56	     &0.43	     &\nd{0.25}	 &\nd{0.23}	 &0.40          \\
    &FPS$\uparrow$       &19.1       &19.5	     &17.8	     &17.6	     &18.5	     &17.6	     &18.7	     &17.9	     &18.3\\
    &MS$\downarrow$      &\fs{10.1}	 &\fs{7.3}	 &\fs{10.9}	 &\fs{9.5}	 &\fs{10.7}	 &\fs{11.7}	 &\fs{10.9}	 &\fs{9.2}	 &\fs{10.0}\\
    \bottomrule
  \end{tabular}
  \label{rgbd_replica}
\end{table}

\subsection{Monocular Keyframe Initialization}
In the context of monocular case, the initialization of a dense scene representation is a challenging yet indispensable step for the generation of high-quality mapping. Feature-based SLAM is capable of providing a sparse point cloud representation of the scene. However, this representation is inadequate for tasks that necessitate high-density and realistic reconstruction, given the limitations of sparse point cloud. To address this issue, we propose a monocular keyframe initialization method that employs the sparse point cloud to generate an initial set of Gaussian primitives, and subsequently achieves a dense and accurate scene representation through a refinement process.

For each newly added keyframe, the sparse point cloud generated from the feature-based SLAM corresponding to the new keyframe are first extracted. These point cloud provide a simple description of the geometric information corresponding to the new keyframe. They are subsequently used to initialize some of the new Gaussian primitives, which are combined with the existing Gaussian primitives to provide a preliminary representation of the scene. The initialised Gaussian primitives are then optimised using the new keyframe in accordance with Eq.~\ref{loss}, and an initial depth map is generated for the new keyframe in terms of Eq.~\ref{eq_depth_ren}. The initial depth map, shown in Fig.~\ref{depth} (a), provides a foundational estimation of the new scene's structure, enabling efficient image sampling and Gaussians generation as described in Sec.~\ref{mapping}, which facilitates the creation of a dense scene representation corresponding to the new keyframe in monocular mode. Once the dense Gaussian primitives for the new keyframe have been generated by the described method, the scene representation is optimized through the implementation of the dynamic keyframe window, as detailed in Sec.~\ref{window}. The rendered depths before and after initialization are illustrated in Fig.~\ref{depth}. MonoGS\cite{matsuki2024gaussian} employs a random value initialization process for Gaussian primitives, which results in depth map that is challenging to describe the scene geometry after initial optimization as shown in Fig.~\ref{depth} (c). For MonoGS, the inherent uncertainty in the geometry further necessitates additional steps to eliminate outliers and to assign more primitives to new viewpoints, with the objective of enhancing the rendering quality. 

This monocular keyframe initialization approach effectively bridges the gap between sparse and dense representations in monocular case. By leveraging the sparse point cloud in ORB-SLAM3 and combining it with adaptive sampling and redundancy removal methods, our approach ensures that the monocular initialization process is both efficient and capable of capturing fine scene details.

\section{Experiments}
To validate the effectiveness of the proposed RP-SLAM, we conduct extensive experiments designed to evaluate its performance in both monocular and RGB-D settings across a range of both real and synthetic datasets. In addition, we perform ablation studies to justify our design choices.

\begin{table}[t]
  \centering
  \caption{Quantitative results for TUM\cite{sturm2012benchmark} dataset. MS denotes model size (Mb).}
  \footnotesize
  \setlength{\tabcolsep}{0.35em}
  \begin{tabular}{*{7}{c}cc}
    \toprule
    Mode &Methods  &Metrics&\texttt{fr1}&\texttt{fr2} & \texttt{fr3} & Avg.  &ATE$\downarrow$  &MS$\downarrow$         \\
    \midrule
    \multirow{12}{*}{Mono} 
        &\multirow{3}{*}{\makecell{MonoGS \\ \cite{matsuki2024gaussian}}}
            &PSNR$\uparrow$      &16.71	&15.59 &19.19      &17.16  \\ 
            &&SSIM$\uparrow $    &0.634	&0.665 &\nd{0.727} &0.675 &4.16 &\fs{2.3}    \\
            &&LPIPS$\downarrow$  &0.411	&0.338 &0.351      &0.367      \\
        
        \cmidrule{2-9}
        &\multirow{3}{*}{\makecell{Photo-\\SLAM\\ \cite{huang2024photo}}}
            &PSNR$\uparrow$      &\nd{20.97}	&21.07      &19.59 &20.54      \\
            &&SSIM$\uparrow $    &\nd{0.743}	&\nd{0.726} &0.692 &\nd{0.720} &1.22 &18.2	  \\
            &&LPIPS$\downarrow$  &\fs{0.228}	&\nd{0.166} &0.239 &\nd{0.211}	   \\
       
        \cmidrule{2-9}
        &\multirow{3}{*}{\makecell{CaRtGS \\ \cite{feng2024cartgs}}}
            &PSNR$\uparrow$      &20.61	&\nd{21.53} &\nd{20.08} &\nd{20.74}  	  \\
            &&SSIM$\uparrow $    &0.726	&0.718      &0.717      &\nd{0.720}	&\fs{1.13} &12.9 	 \\
            &&LPIPS$\downarrow$  &0.248	&\fs{0.159} &\nd{0.236} &0.214   \\
       
        \cmidrule{2-9}
        &\multirow{3}{*}{\makecell{\textbf{RP-SLAM} \\ (ours)}}
            &PSNR$\uparrow$      &\fs{21.87} &\fs{22.48} &\fs{21.32} &\fs{21.89}     \\
            &&SSIM$\uparrow $    &\fs{0.751} &\fs{0.738} &\fs{0.775} &\fs{0.755} &\nd{1.21} &\nd{6.3}  \\
            &&LPIPS$\downarrow$  &\nd{0.235} &0.162      &\fs{0.231} &\fs{0.209}	     \\
    \midrule
    \midrule
    \multirow{18}{*}{RGB-D} 
        &\multirow{3}{*}{\makecell{MonoGS \\ \cite{matsuki2024gaussian}}}
            &PSNR$\uparrow$      &18.67	&15.94 &19.23 &17.95  \\ 
            &&SSIM$\uparrow $    &0.708	&0.798 &0.742 &0.749 &1.45 &\fs{2.7}     \\
            &&LPIPS$\downarrow$  &0.317	&0.311 &0.325 &0.318    \\
        
        \cmidrule{2-9}
        &\multirow{3}{*}{\makecell{Photo-\\SLAM\\ \cite{huang2024photo}}}
            &PSNR$\uparrow$      &20.87	&22.09 &22.74      &21.90    \\
            &&SSIM$\uparrow $    &0.743	&0.765 &\fs{0.780} &0.763 &1.07 &17.1 \\
            &&LPIPS$\downarrow$  &0.235	&0.169 &\nd{0.154} &\fs{0.186}  \\
       
        \cmidrule{2-9}
        &\multirow{3}{*}{\makecell{SplaTAM\\ \cite{keetha2024splatam}}}
            &PSNR$\uparrow$      &\nd{21.88}	&\fs{23.35} &20.61 &21.94   \\
            &&SSIM$\uparrow $    &\fs{0.831}	&\nd{0.852} &0.744 &\fs{0.809} &3.34 &140.4    \\
            &&LPIPS$\downarrow$  &\nd{0.238}	&\fs{0.154} &0.211 &0.201	    \\
        
        \cmidrule{2-9}
        &\multirow{3}{*}{\makecell{RTG-\\SLAM\\ \cite{peng2024rtg}}}
            &PSNR$\uparrow$      &19.38	&17.53 &18.86 &18.59   \\
            &&SSIM$\uparrow $    &0.716	&0.686 &0.761 &0.721 &1.05 &76.5  \\
            &&LPIPS$\downarrow$  &0.465	&0.464 &0.438 &0.456         \\
       
        \cmidrule{2-9}
        &\multirow{3}{*}{\makecell{CaRtGS \\ \cite{feng2024cartgs}}}
            &PSNR$\uparrow$      &20.59	&22.75      &\nd{22.99} &\nd{22.11}  	  \\
            &&SSIM$\uparrow $    &0.729	&0.748      &0.773      &0.750      &\nd{0.99} &13.9	 \\
            &&LPIPS$\downarrow$  &0.253	&\nd{0.158} &\fs{0.151} &\nd{0.187}   \\
       
        \cmidrule{2-9}
        &\multirow{3}{*}{\makecell{\textbf{RP-SLAM} \\ (ours)}}
            &PSNR$\uparrow$      &\fs{22.89}	&\nd{23.32} &\fs{23.07} &\fs{23.09}     \\
            &&SSIM$\uparrow$    &\nd{0.783}	&\fs{0.85} &\nd{0.776} &\nd{0.805} &\fs{0.98} &\nd{3.8}\\
            &&LPIPS$\downarrow$  &\fs{0.228}	&0.161      &0.213      &0.200    \\
    \bottomrule
  \end{tabular}
  \label{tum}
\end{table}

\begin{table}
  \centering
  \caption{Quantitative results for ScanNet++\cite{yeshwanth2023scannet++} dataset in the RGB-D case. MS denotes model size (Mb).}
  \footnotesize
  \setlength{\tabcolsep}{0.5em}
  \begin{tabular}{cccccccccccc}
\toprule
Views &Methods &Metrics &\texttt{S1} &\texttt{S2} &\texttt{S3}  &\texttt{S4} &Avg. \\
\midrule
\multirow{12}{*}{\rotatebox{90}{Training View}} 
    &\multirow{4}{*}{\makecell{SplaTAM\\ \cite{keetha2024splatam}}} 
            &PSNR$\uparrow$     &\nd{27.82}	&\nd{23.41}	&\nd{26.69} &\nd{27.39}	&\nd{25.97}	\\
           & &SSIM$\uparrow$    &\fs{0.946} &\fs{0.885} &0.891      &\fs{0.941} &\nd{0.907}	\\
           & &LPIPS$\downarrow$ &\nd{0.119} &\nd{0.263} &\nd{0.214} &\nd{0.121} &\nd{0.199} \\
           & &MS$\downarrow$    &194.3      &210.7      &313.4      &\nd{150.1} &242.8\\
    \cmidrule{2-8}
    &\multirow{4}{*}{\makecell{RTG-SLAM\\ \cite{peng2024rtg}}} 
            &PSNR$\uparrow$     &20.56	    &21.23	    &24.88		&23.24	   &22.48	\\
           & &SSIM$\uparrow$    &0.872	     &0.981	     &\nd{0.906} &0.901	    &0.893\\
           & &LPIPS$\downarrow$ &0.244	     &0.228	     &0.232		 &0.238	    &0.236\\
           & &MS$\downarrow$    &\nd{150.8} &\nd{149.8} &\nd{168.1} &158.4     &\nd{156.8}\\
    \cmidrule{2-8}
    &\multirow{4}{*}{\makecell{\textbf{RP-SLAM}\\ (ours)}} 
            &PSNR$\uparrow$     &\fs{29.73}	&\fs{27.59}	&\fs{28.96}	&\fs{29.77}	&\fs{29.01}	 \\
           & &SSIM$\uparrow$    &\nd{0.941} &\nd{0.863} &\fs{0.945} &\nd{0.931} &\fs{0.920}	 \\
           & &LPIPS$\downarrow$ &\fs{0.101} &\fs{0.197} &\fs{0.122} &\fs{0.112} &\fs{0.133}	 \\
           & &MS$\downarrow$    &\fs{25.3}  &\fs{32.1}  &\fs{22.8}  &\fs{26.8}  &\fs{26.8}\\
\midrule
\midrule
\multirow{12}{*}{\rotatebox{90}{Novel View}} 
    &\multirow{3}{*}{\makecell{SplaTAM\\ \cite{keetha2024splatam}}} 
             &PSNR$\uparrow$    &\nd{23.31}	&\nd{22.08}	&\nd{24.22}   &\nd{23.07} &\nd{23.17}	\\
           & &SSIM$\uparrow$    &\nd{0.839}  &\fs{0.849} &\fs{0.882}   &\nd{0.889}	&\nd{0.865}	\\
           & &LPIPS$\downarrow$ &\nd{0.288}  &\nd{0.276} &\nd{0.257} 	&\fs{0.196}	&\nd{0.266} \\
    \cmidrule{2-8}
    &\multirow{3}{*}{\makecell{RTG-SLAM\\ \cite{peng2024rtg}}} 
             &PSNR$\uparrow$    &19.32 &19.14	  &19.98	  &20.03	  &19.62	\\
           & &SSIM$\uparrow$    &0.803	&0.801	    &0.812		&0.798	    &0.802\\
           & &LPIPS$\downarrow$ &0.288	&0.297	    &0.268		&0.259	    &0.284\\
    \cmidrule{2-8}
    &\multirow{3}{*}{\makecell{\textbf{RP-SLAM}\\ (ours)}} 
             &PSNR$\uparrow$    &\fs{25.41}	&\fs{25.62}	&\fs{25.83}		&\fs{26.15}	&\fs{25.75}	 \\
           & &SSIM$\uparrow$    &\fs{0.879}  &\nd{0.835} &\nd{0.878}	  &\fs{0.894} &\fs{0.872}	 \\
           & &LPIPS$\downarrow$ &\fs{0.208}  &\fs{0.241} &\fs{0.214}	  &\nd{0.212} &\fs{0.219}	 \\
\bottomrule
  \end{tabular}
  \label{scannet++}
\end{table}

\subsection{Experimental Setup}
\subsubsection{Datesets}
For our quantitative analysis, we evaluate our method on three datesets: TUM-RGBD dataset\cite{sturm2012benchmark} (3 sequences), Replica dataset\cite{straub2019replica} (8 sequences) and ScanNet++ dataset\cite{yeshwanth2023scannet++} (4 sequences). The Replica\cite{straub2019replica} benchmark is the simplest because it contains synthetic scenes, highly accurate and complete (synthetic) depth maps, and small displacements between consecutive camera poses. In contrast, the TUM-RGBD\cite{sturm2012benchmark} benchmark is more difficult because it uses older, lower-quality cameras with poor-quality RGB and depth images. The depth images are quite sparse with a lot of missing information, while the color images have a very high degree of motion blur. ScanNet++\cite{yeshwanth2023scannet++} displays remarkable color and depth image quality in comparison to other benchmarks. Additionally, a supplementary trajectory is provided for each scene, allowing for the assessment of the rendering capabilities in a new viewpoint. We evaluate four scenes (8b5caf3398, 39f36da05b, b20a261fdf, f34d532901) as used in \cite{peng2024rtg}.

\subsubsection{Implementation Details}
All experimental evaluations are executed on a desktop PC with an Intel Core i9-14900KF CPU and NVIDIA RTX 4090 GPU. In our implementation, we use ORB-SLAM3\cite{campos2021orb} for camera tracking due to its robustness in multiple environments. The 3DGS module from MonoGS\cite{matsuki2024gaussian} is utilized, and Gaussian primitives parameters initialization method and the original hyperparameters are retained. Additionally, the Gaussians densification and pruning methods present in MonoGS are also maintained. The spherical harmonics degree is set to 0 as in MonoGS. The iterations of optimization for each keyframe is set to 100 and the initial iterations is set to 50 in monocular case. We follow MonoGS by setting $\lambda_{pho}$ and $\lambda_{iso}$ to 0.9 and 10, respectively. For other hyperparameters, we set $\tau=15$, $\lambda=1.0$, $k_1=5$, $k_2=3$. In addition, we set $c$ to 8 on Replica\cite{straub2019replica} and TUM\cite{sturm2012benchmark} datasets and to 4 on ScanNet++\cite{yeshwanth2023scannet++} dataset based on its high-resolution details.

\subsubsection{Metrics}
In regard to the precision of camera tracking, we present the Root Mean Square Error (RMSE) of the Absolute Trajectory Error (ATE) of the keyframes in centimeter. In order to evaluate the quality of map rendering, three commonly used metrics are considered: PSNR, SSIM, and LPIPS. Furthermore, the computational and storage efficiency are evaluated by measuring the system frame rate and the model size after mapping. The average of three runs is reported for all evaluations. Best results are highlighted as \colorbox{colorFst}{\bf first} and \colorbox{colorSnd}{second}. 

\subsubsection{Baseline Methods}
We compare our method with several state-of-the-art Gaussian SLAM approaches such as MonoGS\cite{matsuki2024gaussian}, SplaTAM\cite{keetha2024splatam}, Photo-SLAM\cite{huang2024photo}, RTG-SLAM\cite{peng2024rtg} and CaRtGS\cite{feng2024cartgs}. Among these methods, MonoGS, Photo-SLAM and CaRtGS can be applied to monocular and RGB-D cases, whereas others are exclusively utilized for RGB-D data. We reproduce the results using the official code and run all experiments on the same desktop computer. Since only SplaTAM\cite{keetha2024splatam} and RTG-SLAM\cite{peng2024rtg} support ScanNet++\cite{yeshwanth2023scannet++} dataset, we compare with these two methods on ScanNet++ in the RGB-D case.

\begin{table}
  \centering
  \caption{Ablation study for Monocular Keyframe Initialization (MKI), Efficient Incremental Mapping (EIM) and Dynamic Keyframe Window (DKW) on Replica Office0\cite{straub2019replica} in monocular case. MS denotes model size (Mb).}
  \footnotesize
  \setlength{\tabcolsep}{1.8em}
  \begin{tabular}{ccccc}
    \toprule
    MKI         &EIM         &DKW         &PSNR$\uparrow$ & MS$\downarrow$ \\
    \midrule
    \redx       &\redx       &\redx       &34.87          &19.3\\
    \greencheck &\redx       &\redx       &36.67          &16.8\\
    \redx       &\greencheck &\redx       &34.23          &15.6\\
    \redx       &\redx       &\greencheck &35.28          &18.9\\
    \greencheck &\greencheck &\greencheck &\textbf{37.74} &\textbf{11.8}\\
    \bottomrule
  \end{tabular}
  \label{ab_modules}
\end{table}

\begin{table}
  \centering
  \caption{Ablation study for minimum cell size on Replica Office0\cite{straub2019replica} in RGB-D case. MS denotes model size (Mb).}
  \footnotesize
  \setlength{\tabcolsep}{2.7em}
  \begin{tabular}{ccc}
    \toprule
    minimum cell size         &PSNR$\uparrow$ & MS$\downarrow$\\
    \midrule
    4       &42.08          &29.4\\
    8       &41.26          &10.1\\
    16      &37.12          &5.8\\
    \bottomrule
  \end{tabular}
  \label{ab_cell_size}
\end{table}

\subsection{Camera Tracking Accuracy}
The evaluation of camera tracking accuracy across Tab.~\ref{mono_replica}, \ref{rgbd_replica} and \ref{tum} demonstrates the robust performance of RP-SLAM in both monocular and RGB-D modes. As RP-SLAM builds upon ORB-SLAM3\cite{campos2021orb} for camera tracking, it inherently inherits the well-established precision of ORB-SLAM3's feature-based pose estimation. This foundation ensures that RP-SLAM provides accurate camera pose estimation across diverse datasets, as evidenced by the consistently low Absolute Trajectory Error (ATE) values reported.

For monocular mode, RP-SLAM achieves an ATE of 0.63 on the Replica dataset (Tab.~\ref{mono_replica}) and 1.21 on the TUM dataset (Tab.~\ref{tum}). These results are comparable to or slightly higher than state-of-the-art methods such as Photo-SLAM\cite{huang2024photo} and CaRtGS\cite{feng2024cartgs} but significantly outperform the coupled approach like MonoGS\cite{matsuki2024gaussian}. The accuracy of camera tracking in the coupled approach is contingent upon the availability of a high-quality representation of the scene. In the monocular case, MonoGS is unable to acquire accurate scene geometry in a timely manner, resulting in a low-quality representation of the scene. This, in turn, leads to poor camera tracking accuracy. In the RGB-D mode, coupled methods, such as MonoGS\cite{matsuki2024gaussian} and SplaTAM\cite{keetha2024splatam}, are capable of achieving relatively accurate scene representations through the utilization of the depth map. Consequently, they are able to attain camera tracking accuracies on the Replica\cite{straub2019replica} dataset that are comparable to those of decoupled methods, as evidenced in Tab.~\ref{rgbd_replica}. In challenging real-world scenarios, however, the decoupled approaches do prove to be more effective, as evidenced by the results presented in Tab.~\ref{tum}, which demonstrate that our approach achieves the highest ATE. 

By decoupling camera tracking from Gaussians optimization, RP-SLAM achieves a key benefit: real-time performance. As illustrated in Tab.~\ref{rgbd_replica}, the system frame rate of RP-SLAM is approximately 20 times higher than that of MonoGS\cite{matsuki2024gaussian} and nearly 100 times higher than that of SplaTAM\cite{keetha2024splatam}. The decoupled design ensures that the computational load associated with rendering and optimizing Gaussian primitives does not interfere with the responsiveness of the camera tracking pipeline. This approach allows RP-SLAM to maintain competitive tracking accuracy while also achieving real-time system frame rates.

\begin{figure}
  \centering
  \scriptsize
  \setlength{\tabcolsep}{0.3pt}
  \newcommand{\sz}{0.49}  %
  \begin{tabular}{cccc}
    \makecell{\includegraphics[width=\sz\linewidth]{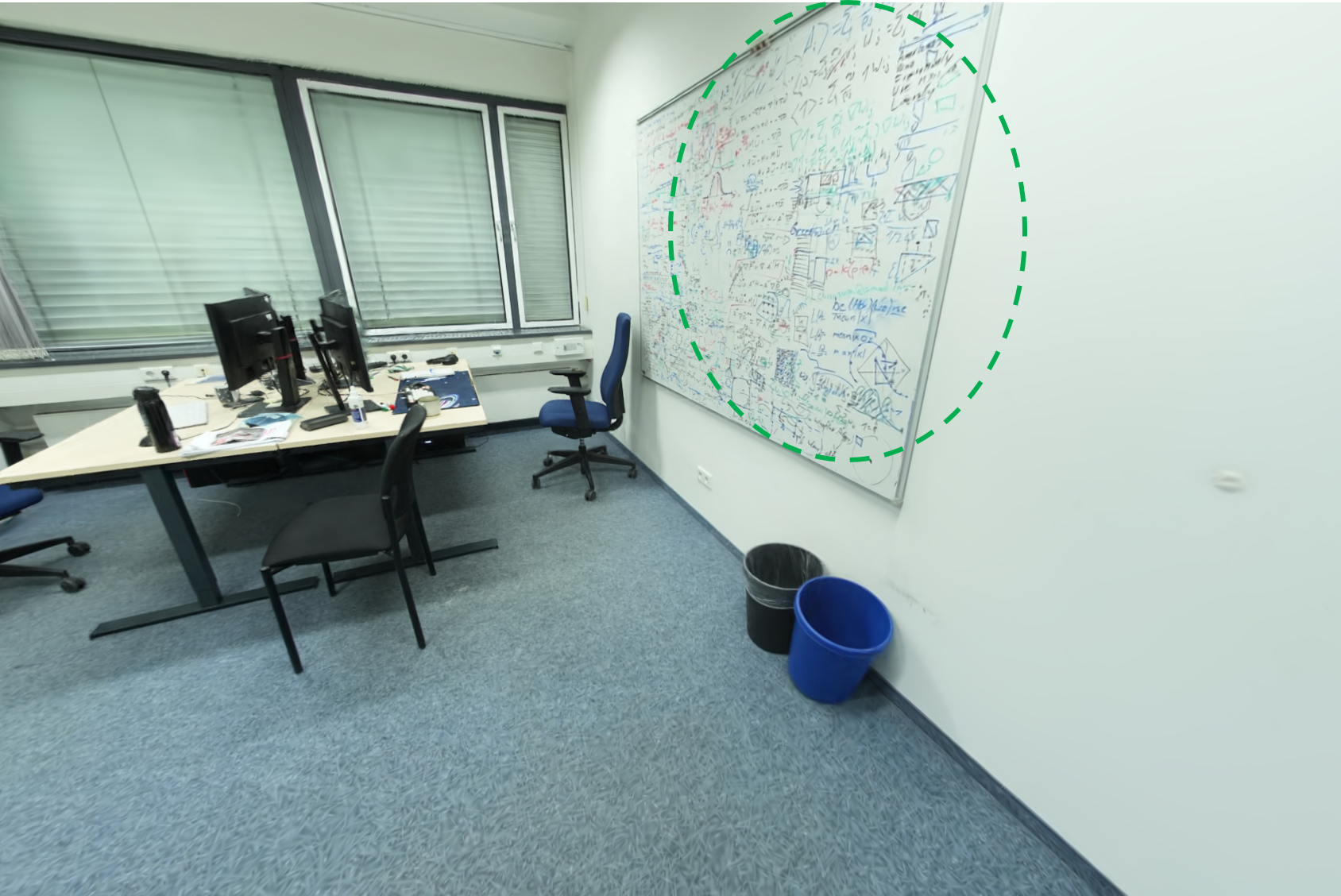}}   &
    \makecell{\includegraphics[width=\sz\linewidth]{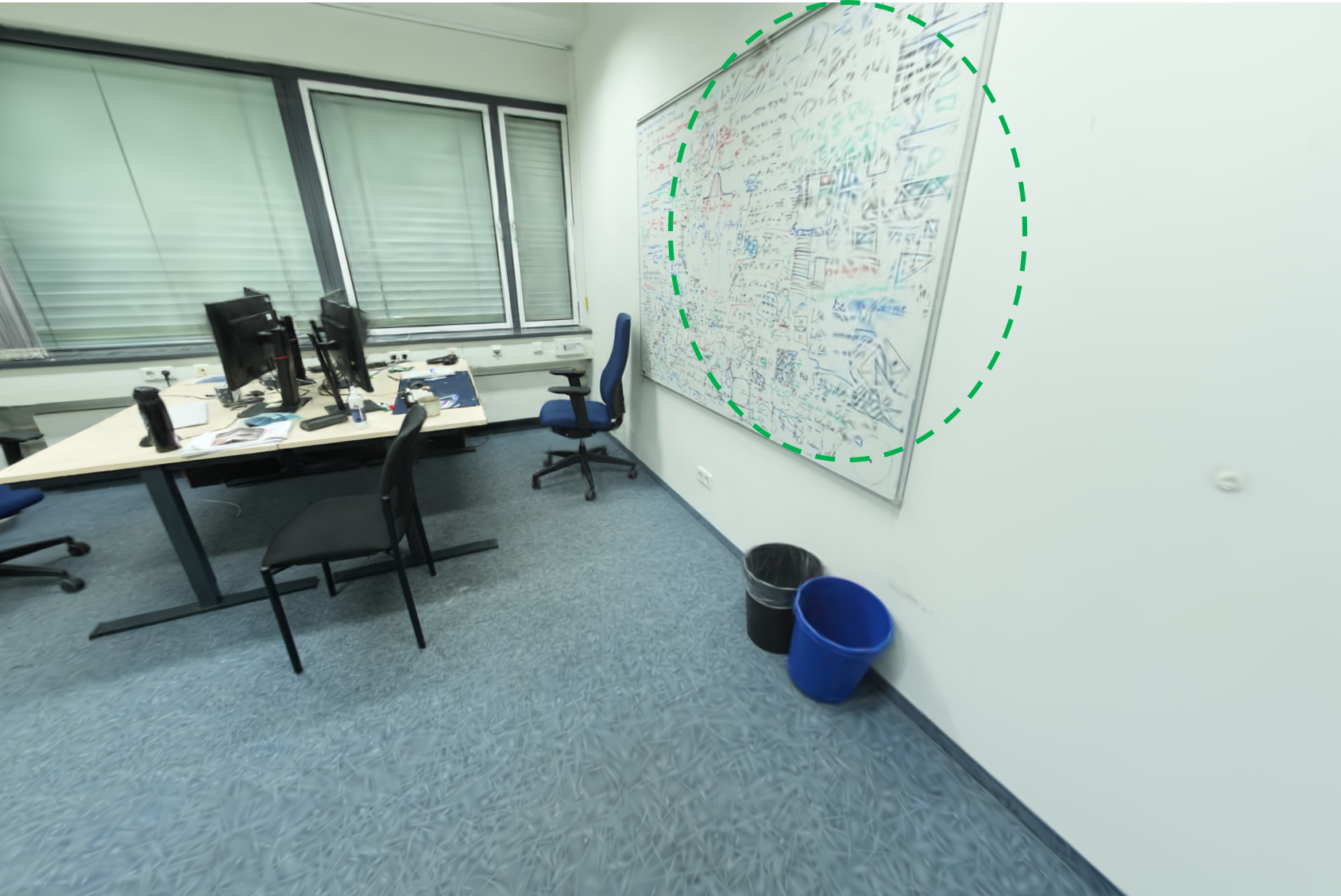}}        \\
    \vspace{1mm}
    \makecell{$c=4$}                                         &
    \makecell{$c=8$}                                                \\
  \end{tabular}
  \caption{Effect of different minimum cell sizes on rendering high-resolution image in ScanNet++\cite{yeshwanth2023scannet++} dataset. When $c=4$, the handwriting on the whiteboard is observed to be more discernible. Zoom in for a clearer view.}
  \label{cell_size_render}
\end{figure}

\begin{figure*}
  \centering
  \scriptsize
  \setlength{\tabcolsep}{0.5pt}
  \newcommand{\sz}{0.19}  %
  \begin{tabular}{lcccccc}
\vspace{-0.5mm}
    \makecell{\rotatebox{90}{GT}}                               &
    \makecell{\includegraphics[width=\sz\linewidth]{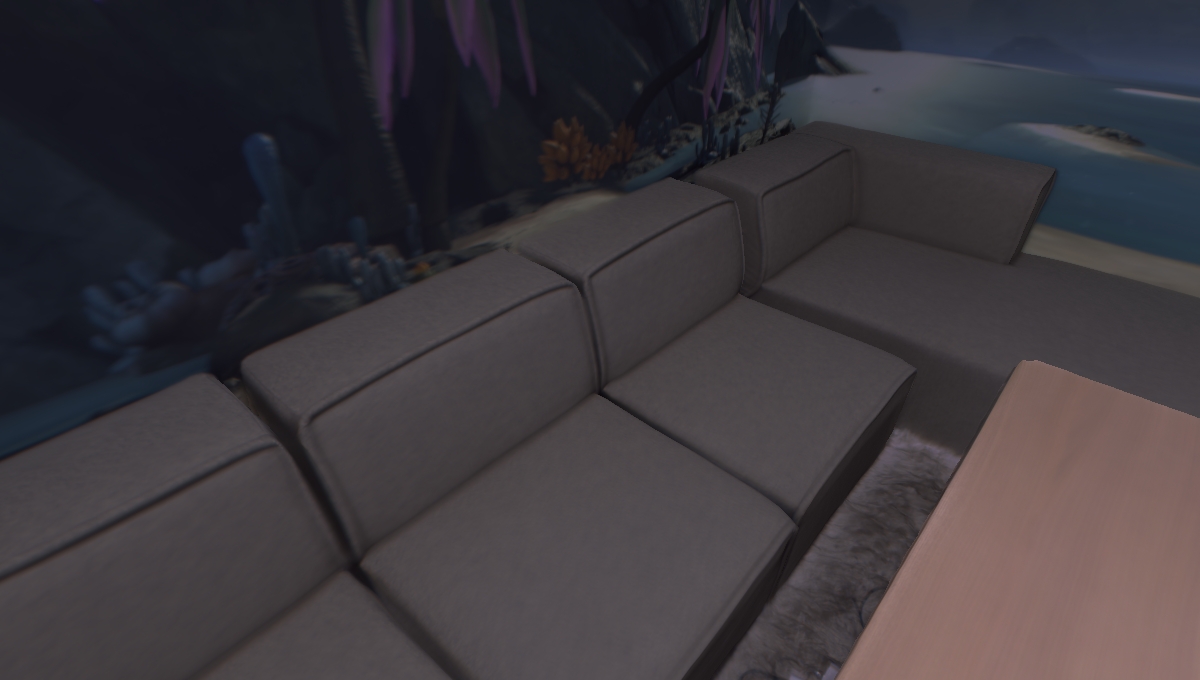}}        &
    \makecell{\includegraphics[width=\sz\linewidth]{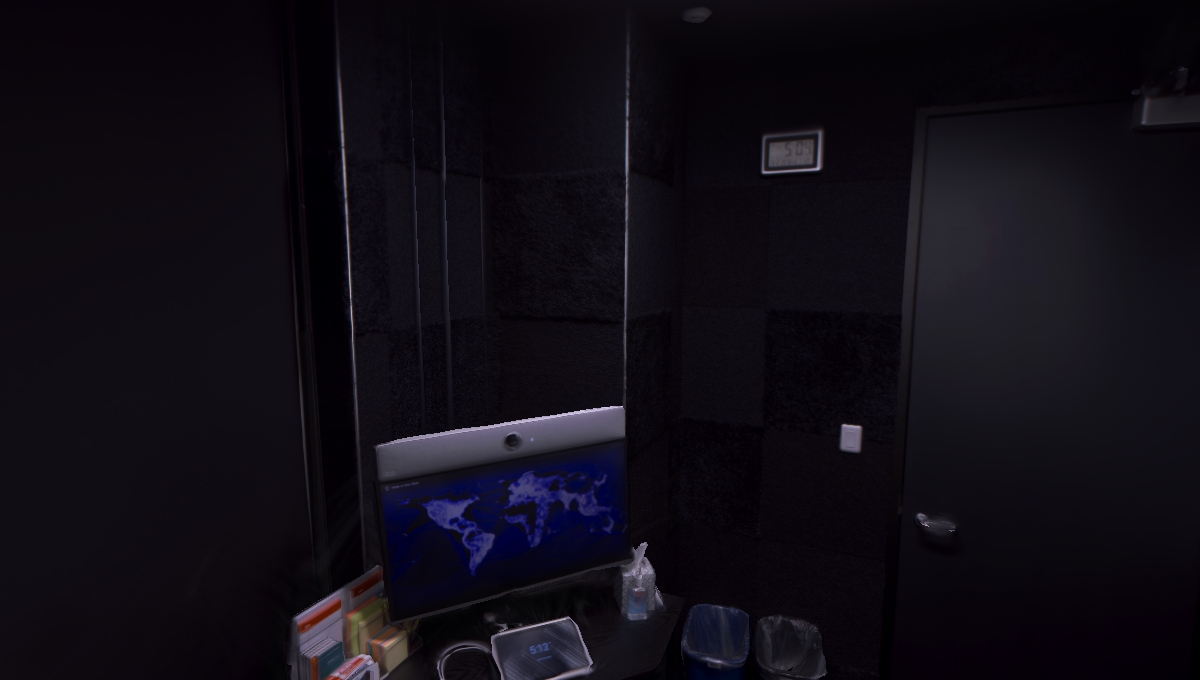}} &
    \makecell{\includegraphics[width=\sz\linewidth]{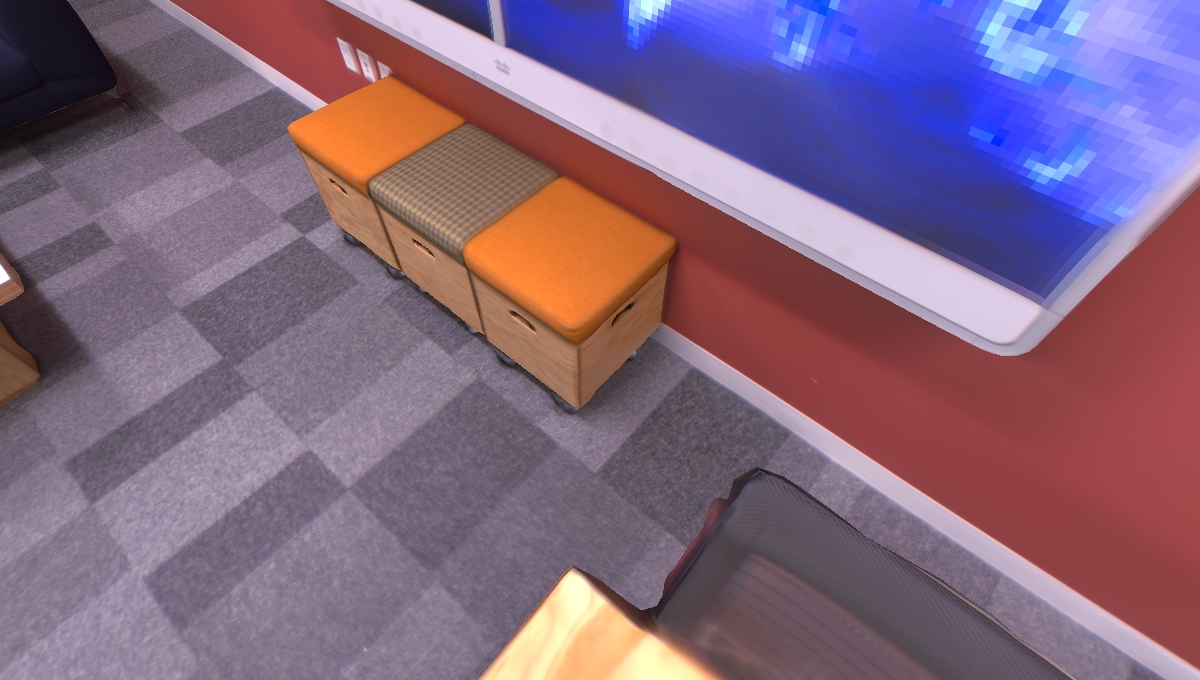}}   &
    \makecell{\includegraphics[width=\sz\linewidth]{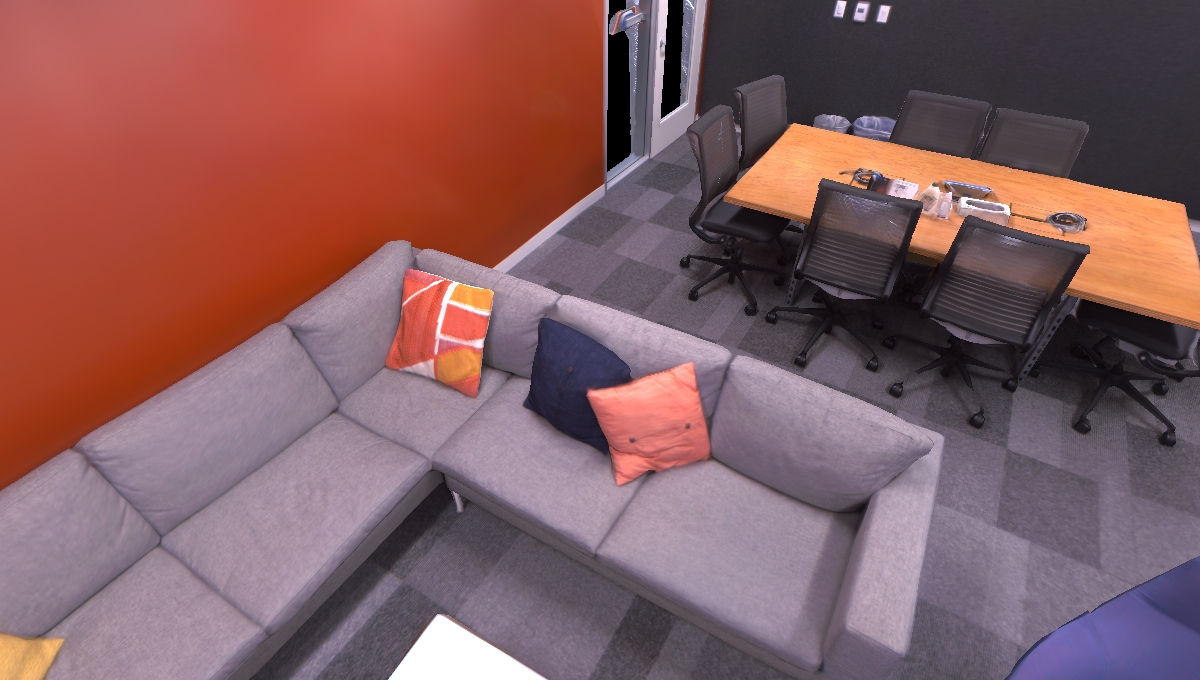}} &
    \makecell{\includegraphics[width=\sz\linewidth]{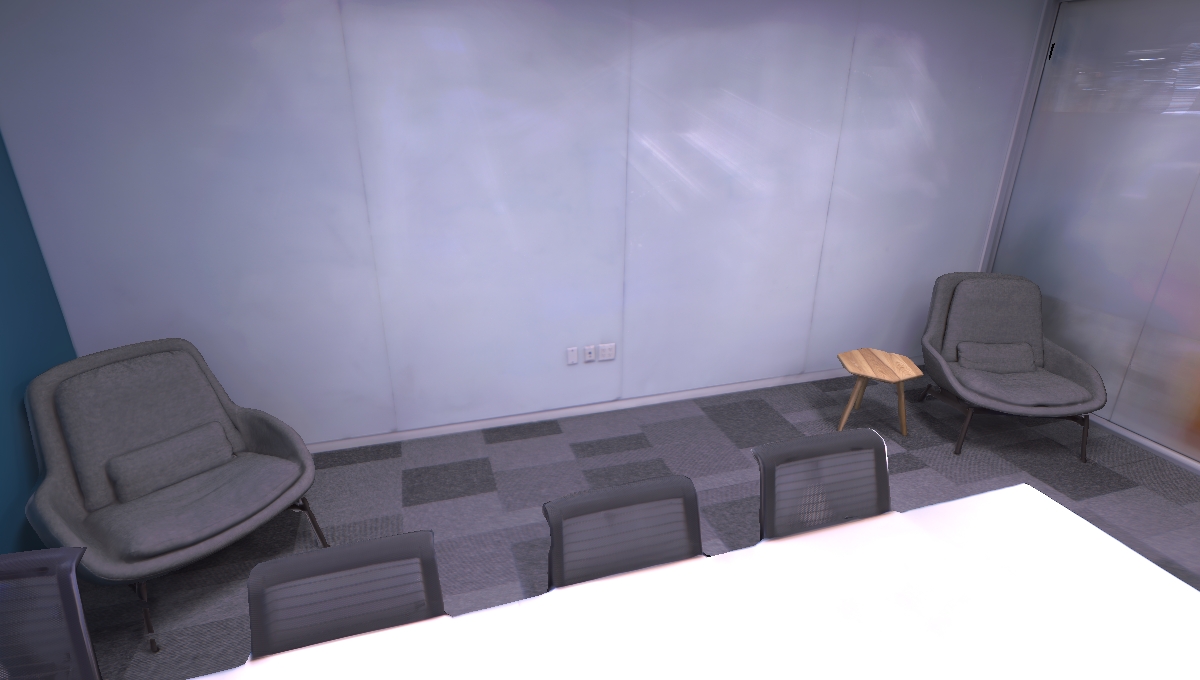}}        \\
\vspace{-0.5mm}
    \makecell{\rotatebox{90}{MonoGS~\cite{matsuki2024gaussian}}}                               &
    \makecell{\includegraphics[width=\sz\linewidth]{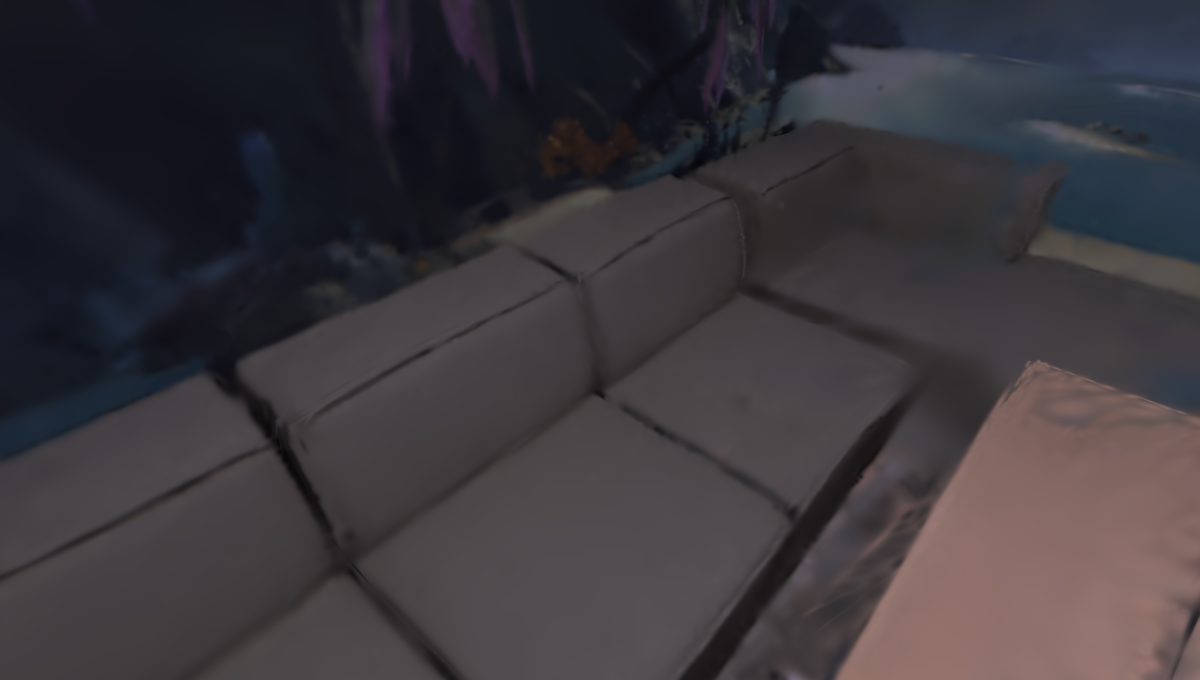}}        &
    \makecell{\includegraphics[width=\sz\linewidth]{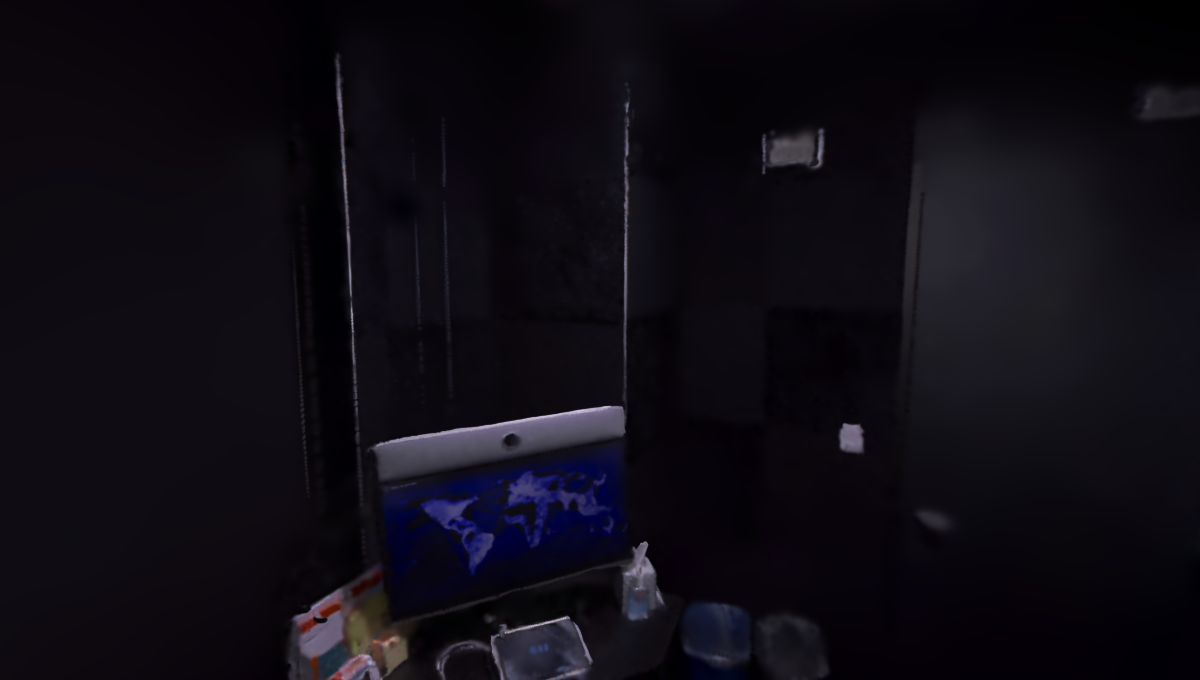}} &
    \makecell{\includegraphics[width=\sz\linewidth]{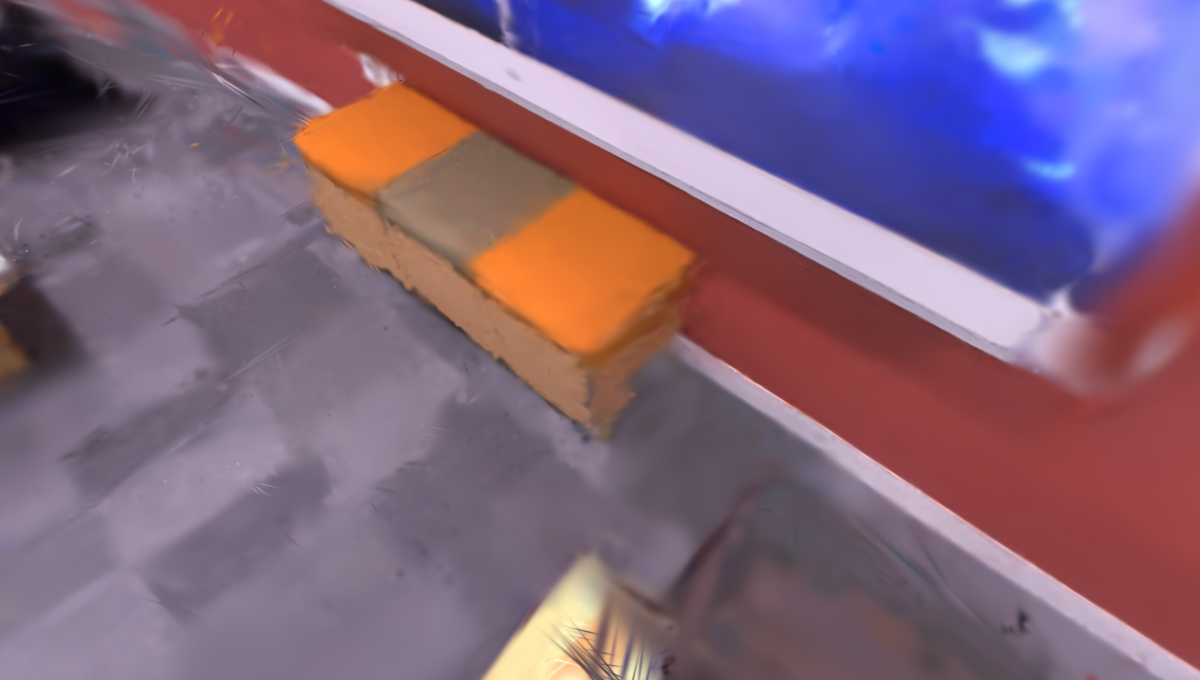}}   &
    \makecell{\includegraphics[width=\sz\linewidth]{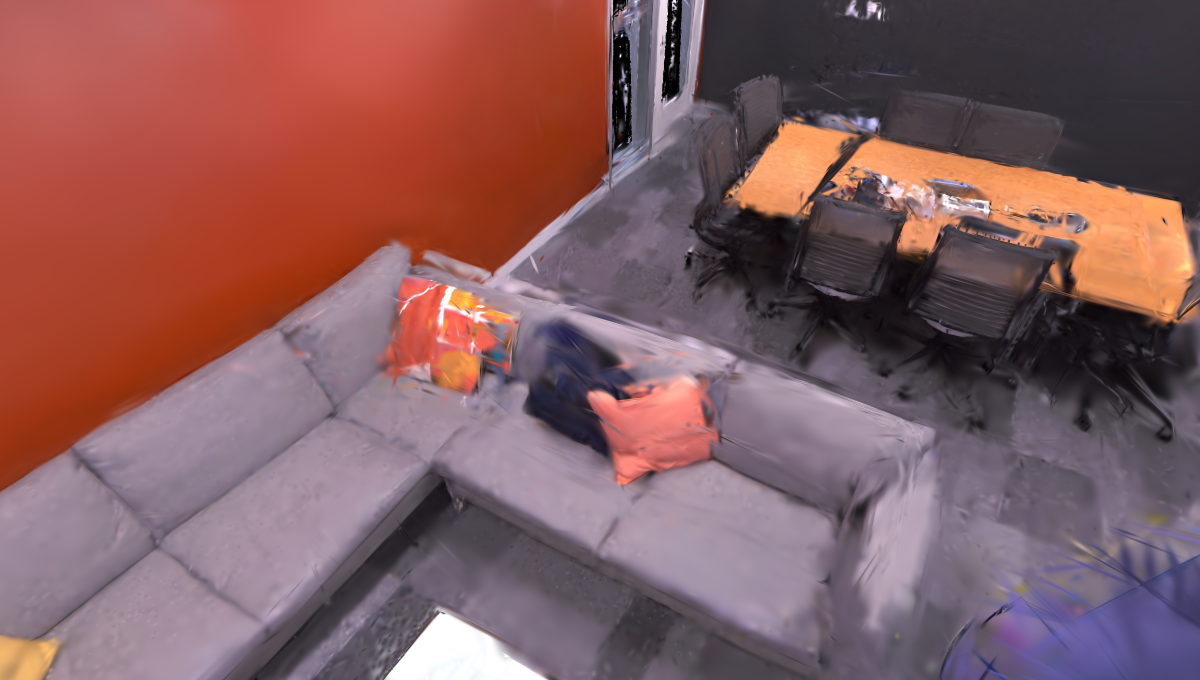}} &
    \makecell{\includegraphics[width=\sz\linewidth]{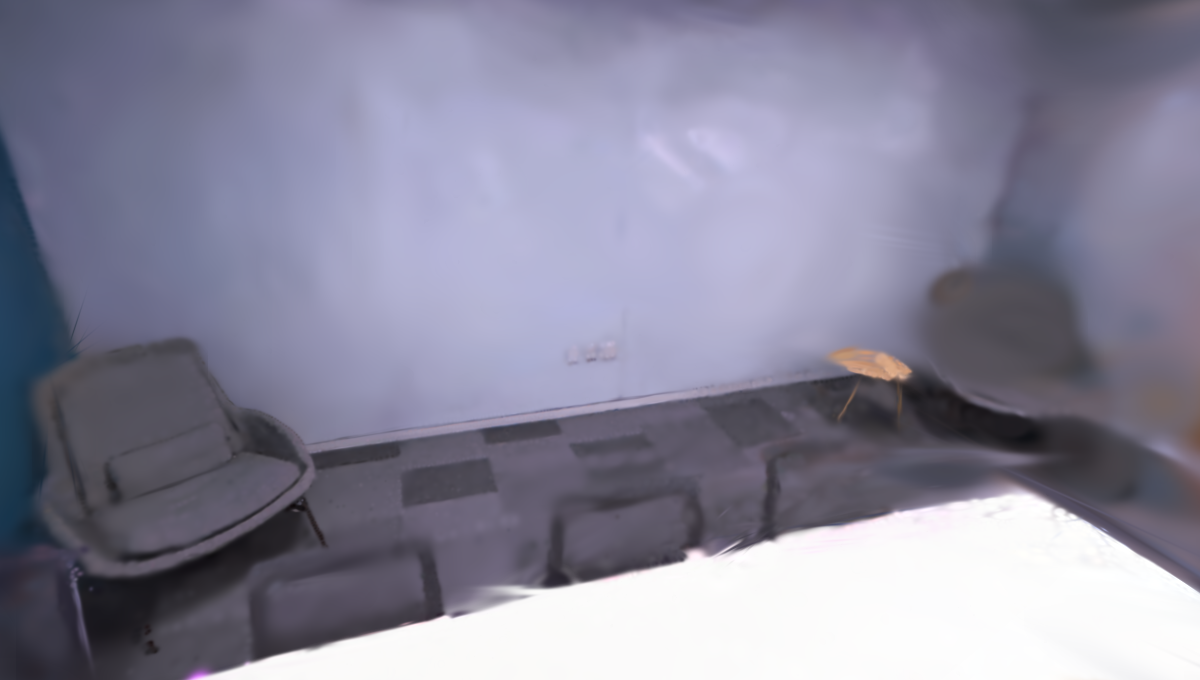}}       \\
\vspace{-0.5mm}
    \makecell{\rotatebox{90}{Photo-SLAM~\cite{huang2024photo}}}                               &
    \makecell{\includegraphics[width=\sz\linewidth]{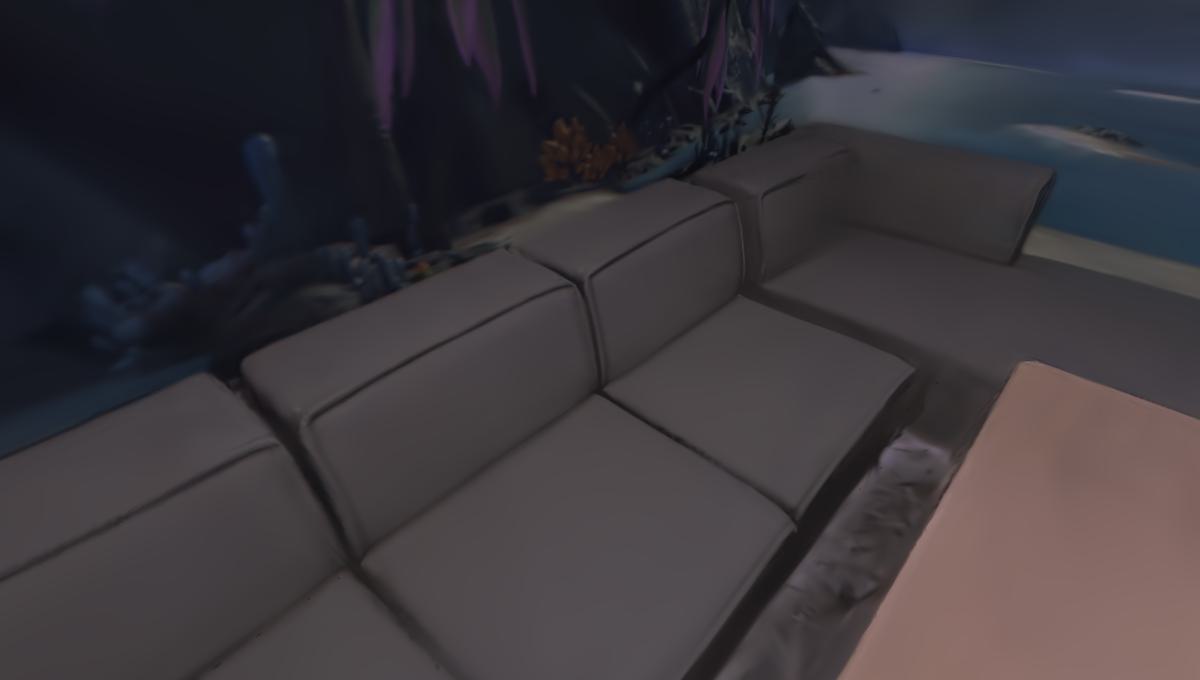}}          &
    \makecell{\includegraphics[width=\sz\linewidth]{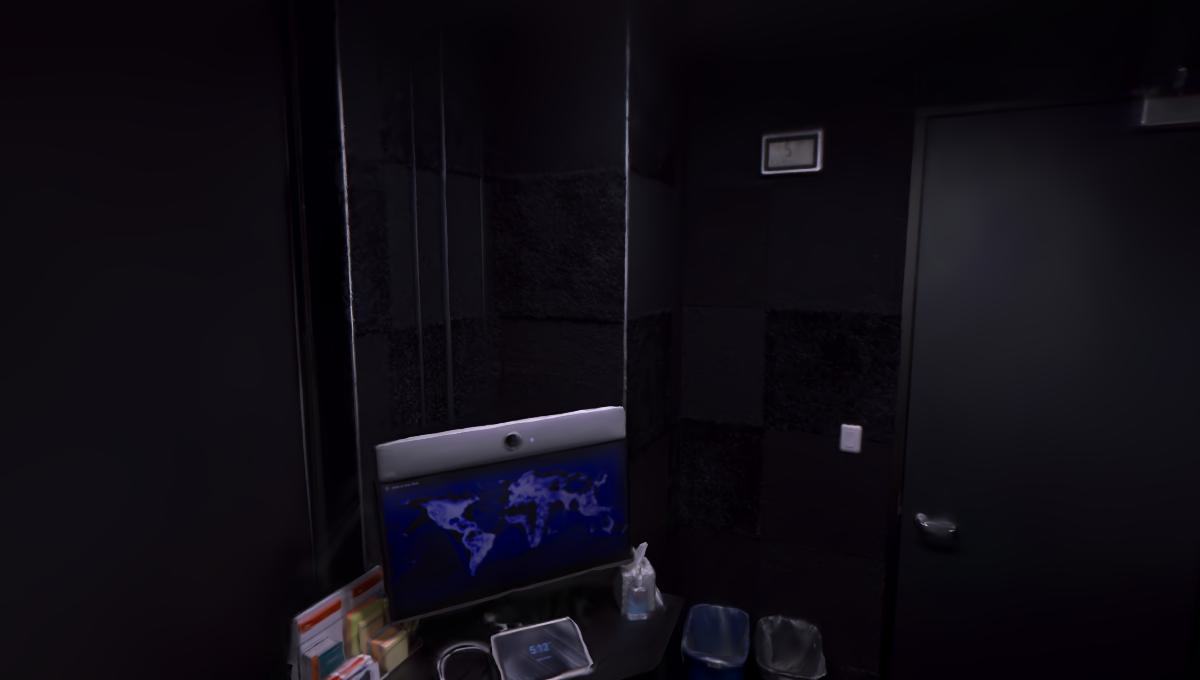}}   &
    \makecell{\includegraphics[width=\sz\linewidth]{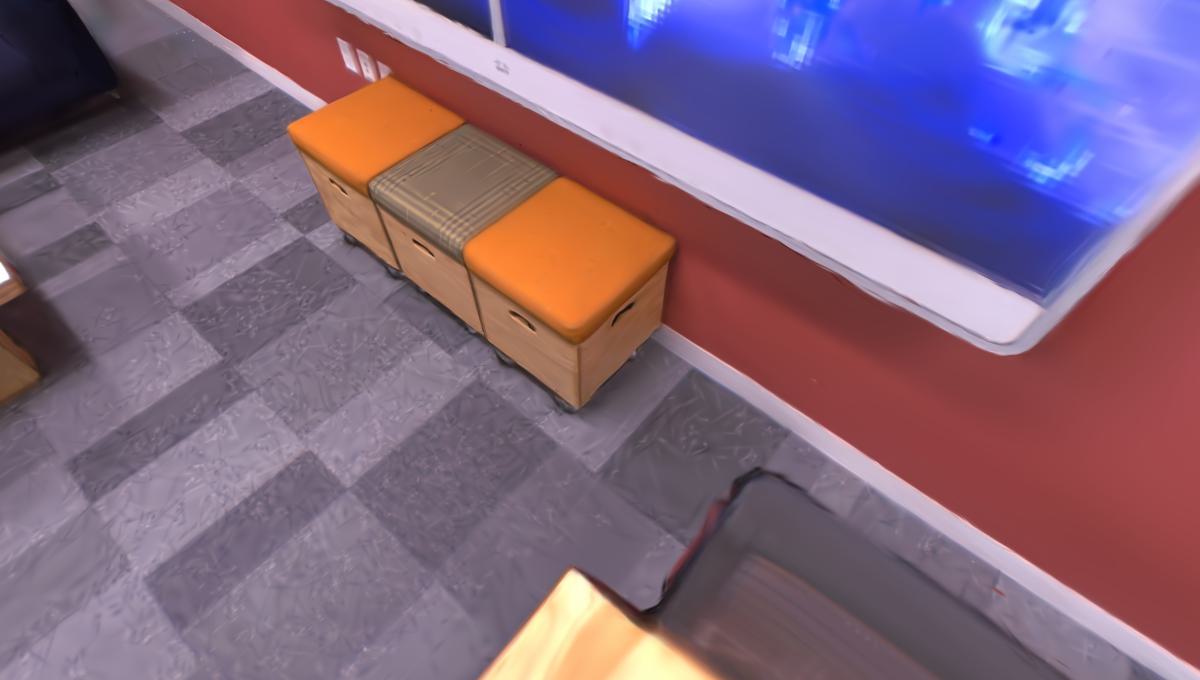}}     &
    \makecell{\includegraphics[width=\sz\linewidth]{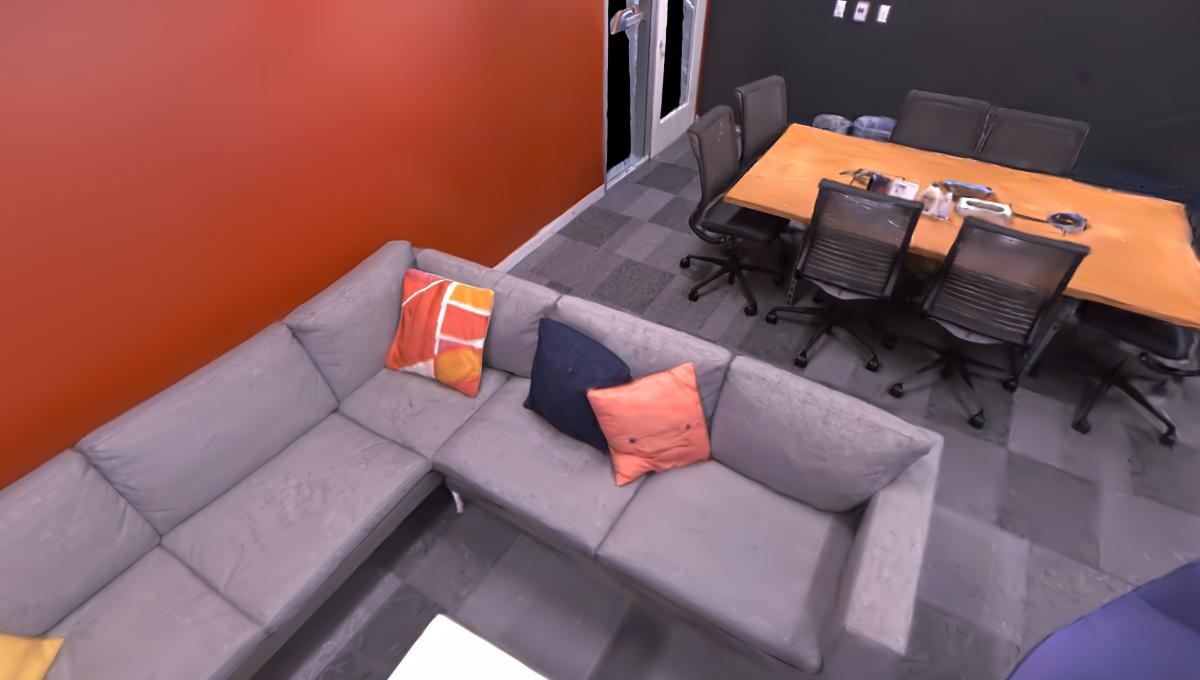}}   &
    \makecell{\includegraphics[width=\sz\linewidth]{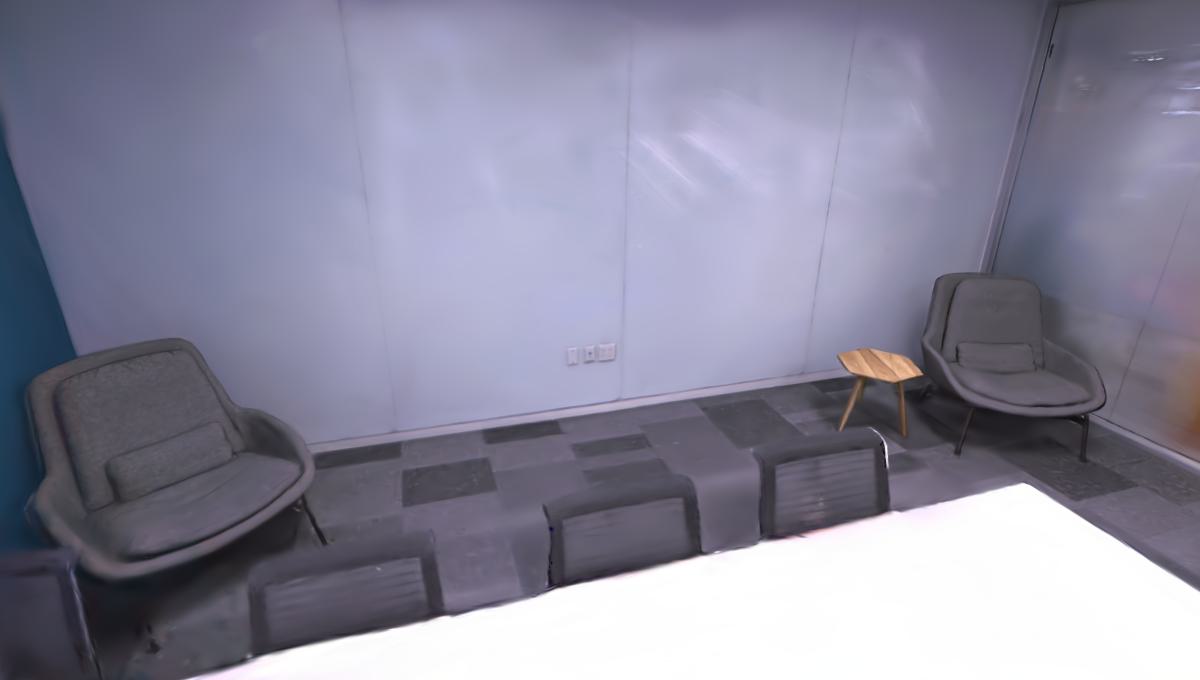}}         \\
\vspace{-0.5mm}
    \makecell{\rotatebox{90}{CaRtGS~\cite{feng2024cartgs}}}                            &
    \makecell{\includegraphics[width=\sz\linewidth]{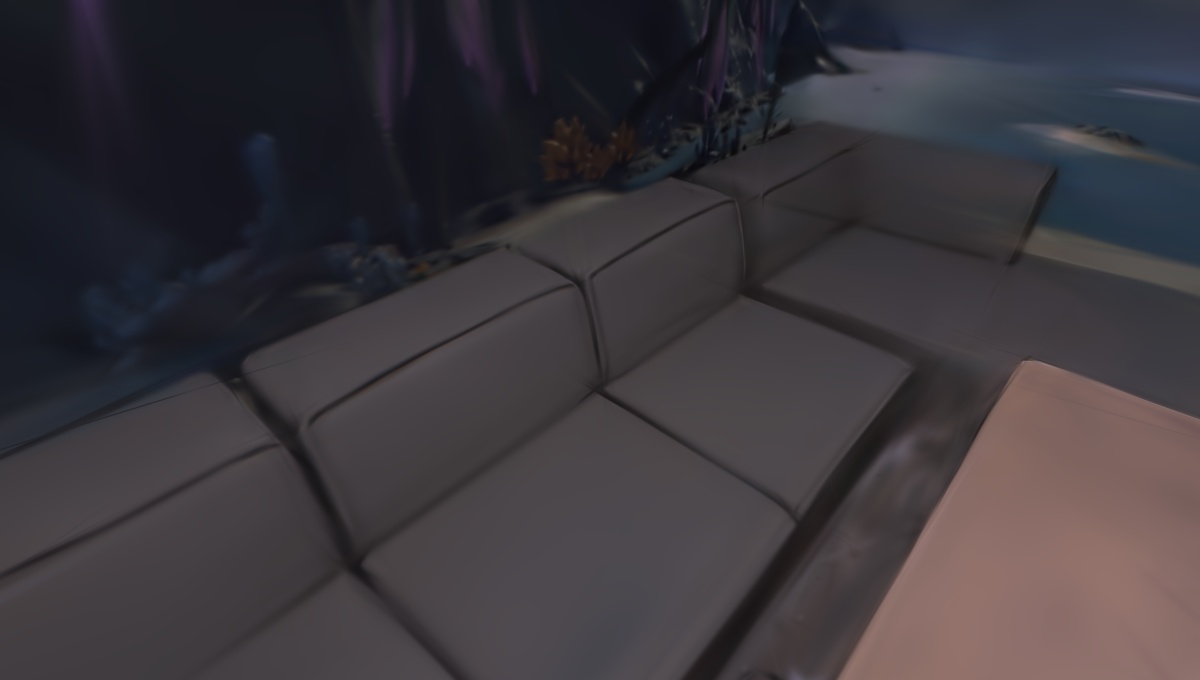}}         &
    \makecell{\includegraphics[width=\sz\linewidth]{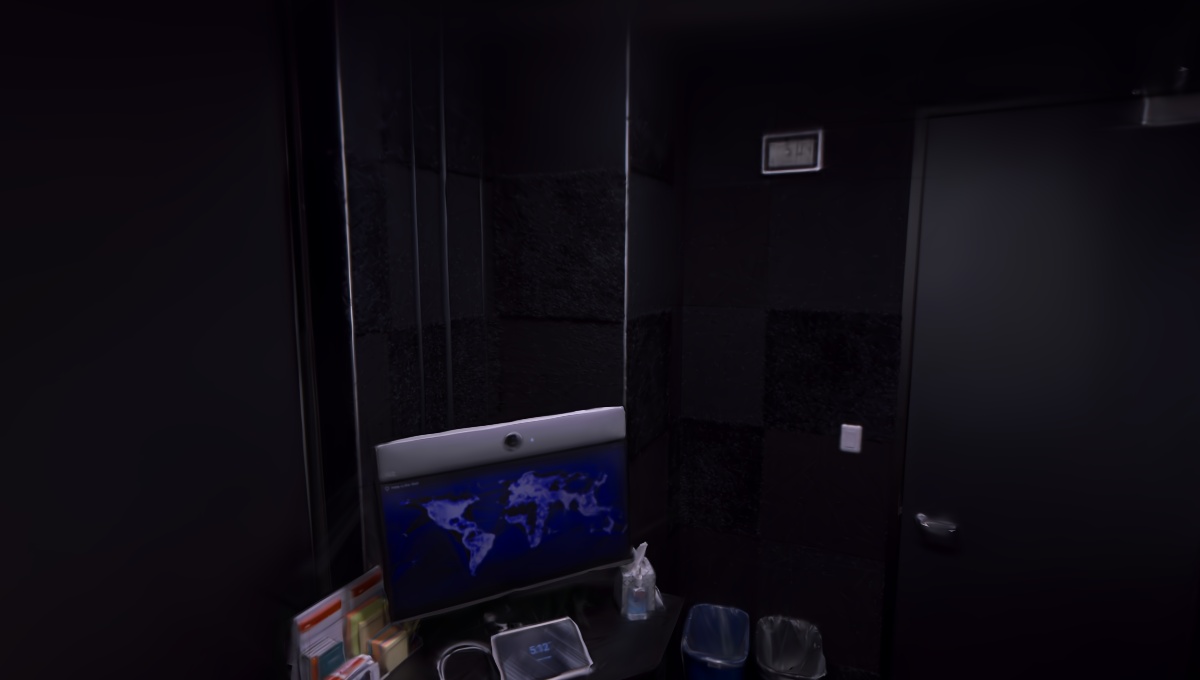}}  &
    \makecell{\includegraphics[width=\sz\linewidth]{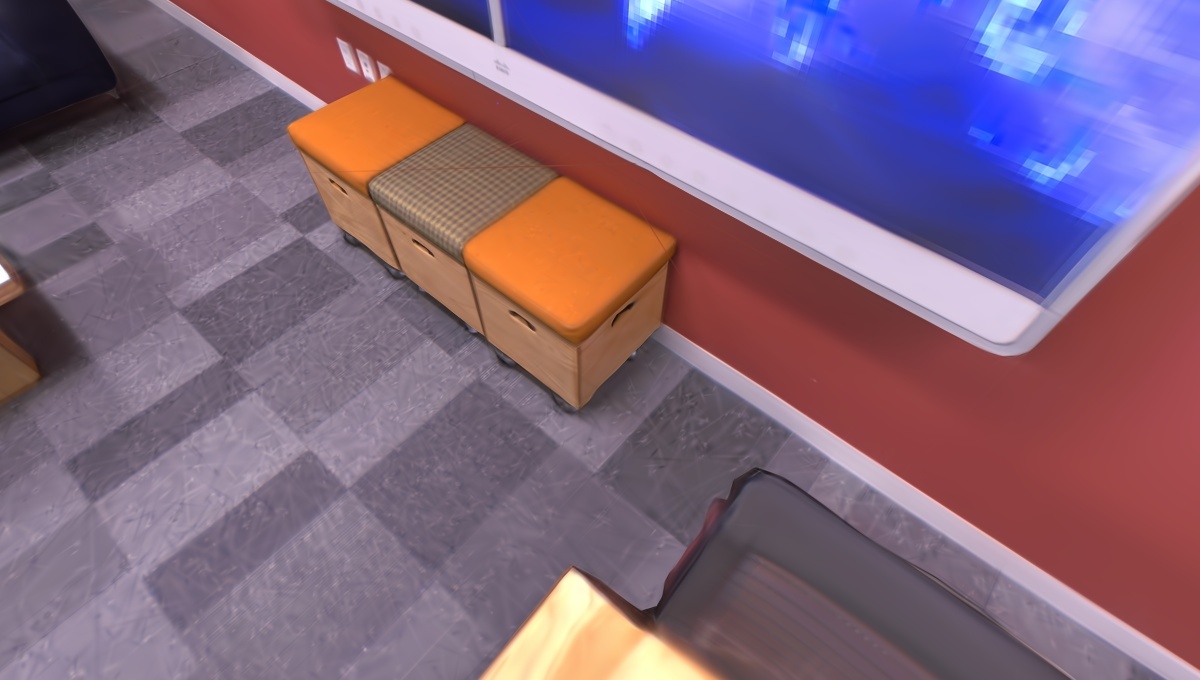}}    &
    \makecell{\includegraphics[width=\sz\linewidth]{figs/rendering/mono/o3_photo.jpg}}  &
    \makecell{\includegraphics[width=\sz\linewidth]{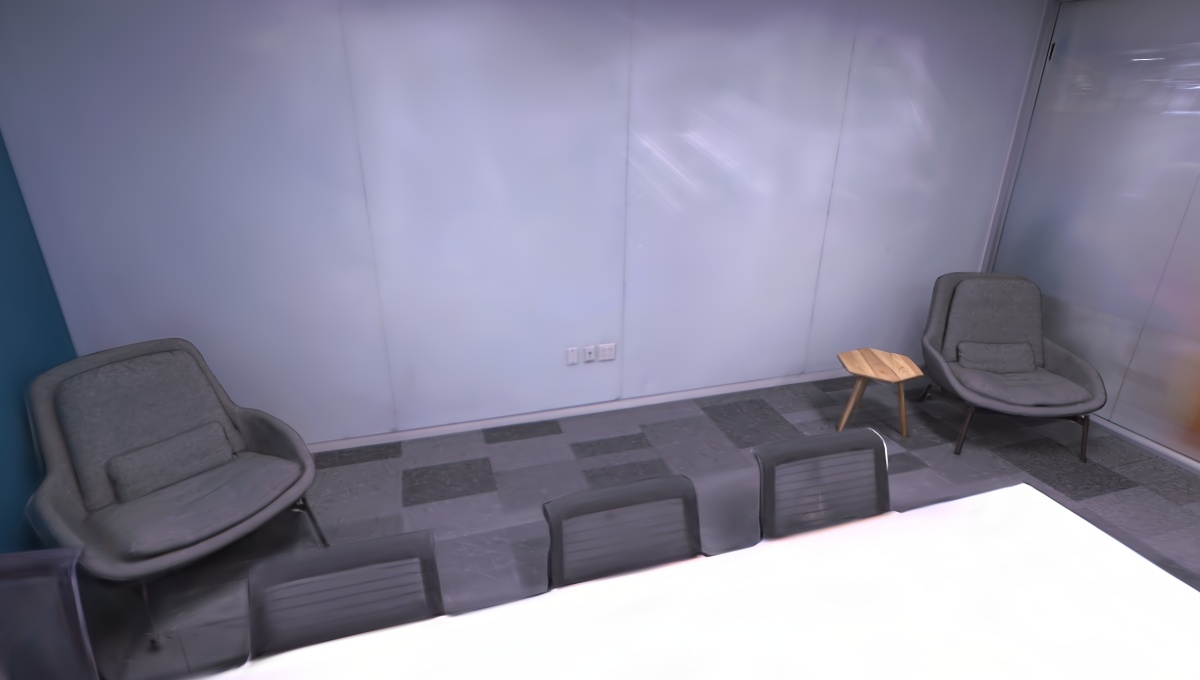}}        \\
\vspace{-0.5mm}
    \makecell{\rotatebox{90}{RP-SLAM (ours)}}                           &
    \makecell{\includegraphics[width=\sz\linewidth]{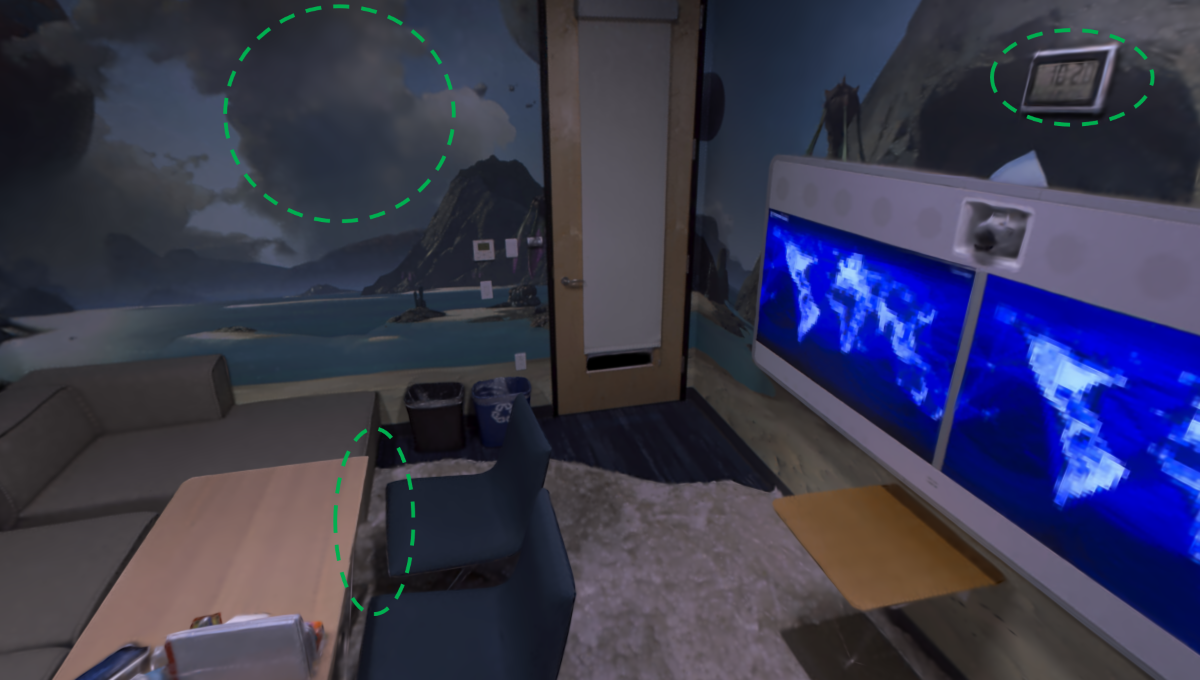}}         &
    \makecell{\includegraphics[width=\sz\linewidth]{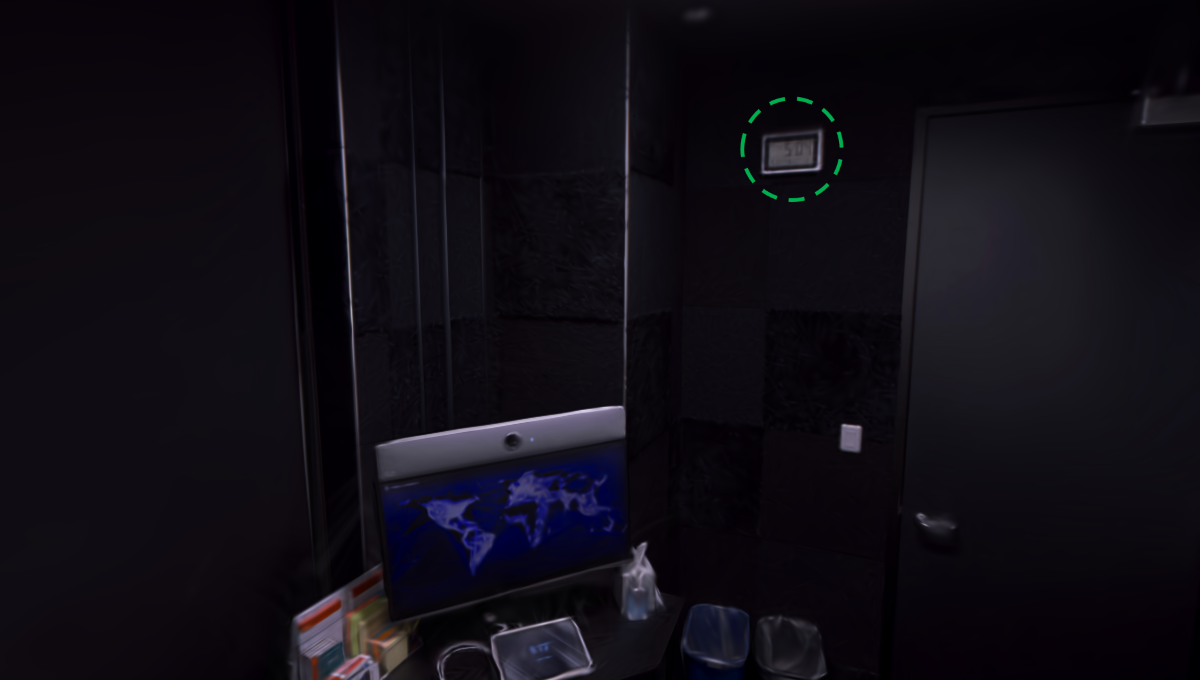}}  &
    \makecell{\includegraphics[width=\sz\linewidth]{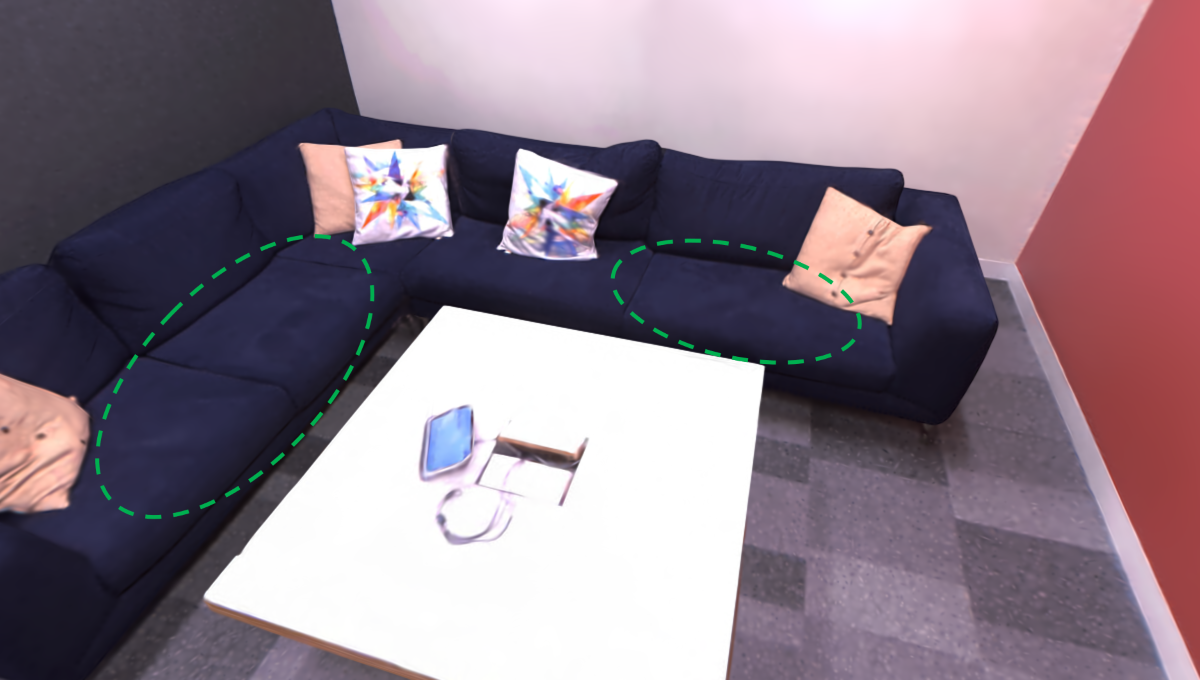}}    &
    \makecell{\includegraphics[width=\sz\linewidth]{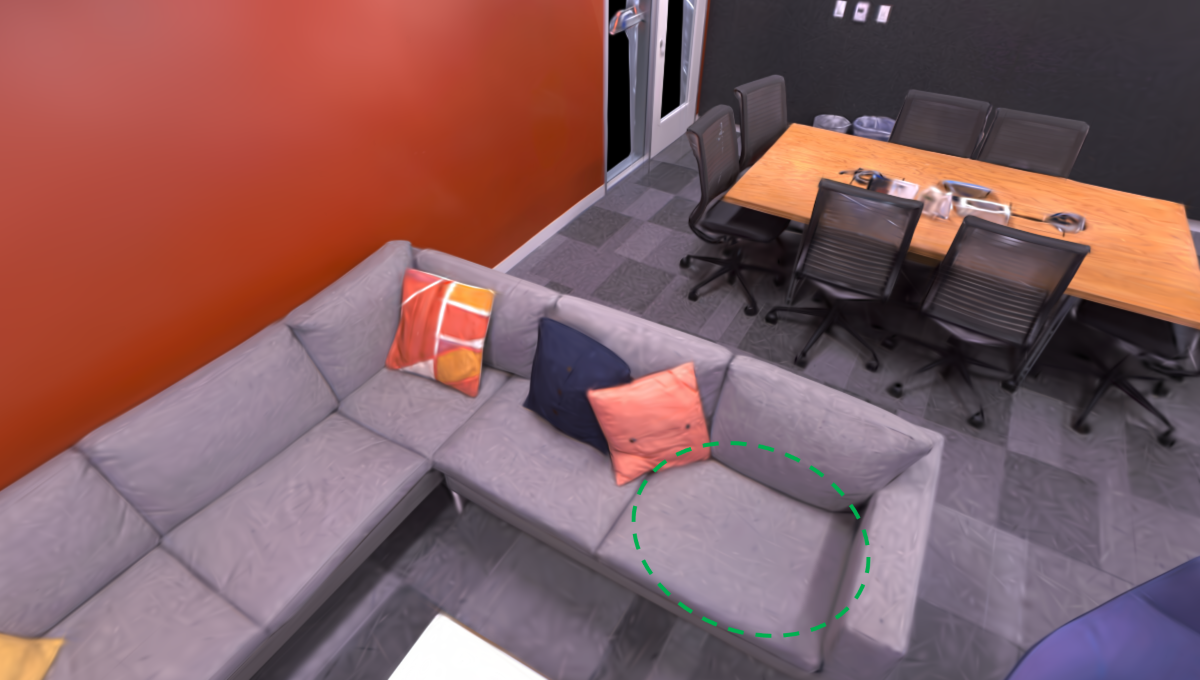}}  &
    \makecell{\includegraphics[width=\sz\linewidth]{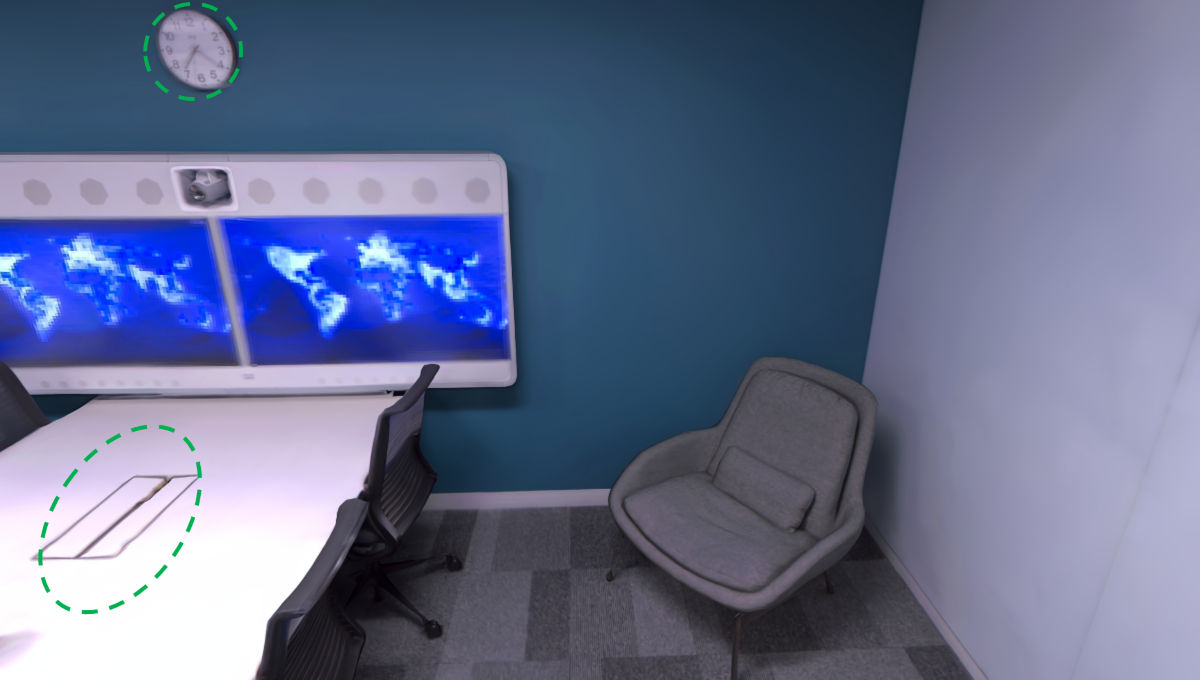}}        \\
                                                                                       &
    \makecell{(a)}                                                                     &
    \makecell{(b)}                                                                     &
    \makecell{(c)}                                                                     &
    \makecell{(d)}                                                                     &
    \makecell{(e)}                                                                      \\
  \end{tabular}
  \caption{Qualitative comparisons on Replica\cite{straub2019replica} dataset in the monocular case. The green dashed boxes in our method mark areas where RP-SLAM outperforms other methods, such as sharper textures and fewer artefacts. Zoom in for a clearer view.}
  \label{render_mono}
\end{figure*}

\subsection{Rendering Quality Results}
The results of the rendering quality of RP-SLAM on the Replica\cite{straub2019replica}, TUM\cite{sturm2012benchmark}, and ScanNet++\cite{yeshwanth2023scannet++} datasets in RGB-D mode demonstrate that RP-SLAM is consistently effective in achieving realistic reconstructions. With regard to the Replica dataset (Tab.~\ref{rgbd_replica}), RP-SLAM achieves the highest average PSNR (37.12) and the lowest LPIPS (0.056) while maintaining a high SSIM (0.971). This is a notable improvement over the results obtained by MonoGS\cite{matsuki2024gaussian}, Photo-SLAM\cite{huang2024photo}, SplaTAM\cite{keetha2024splatam}, and CaRtGS\cite{feng2024cartgs}. While RTG-SLAM\cite{peng2024rtg} achieves the highest SSIM, RP-SLAM produces results that are nearly identical, with only one-sixth of its model size. Moreover, CaRtGS attains the highest LPIPS in several scenarios, yet its model size is approximately twice that of our own. Similarly, on the TUM dataset (Tab.~\ref{tum}), RP-SLAM achieves a PSNR of 23.09, an SSIM of 0.805, and an LPIPS of 0.200. In comparison to MonoGS, RP-SLAM demonstrates superior performance in terms of PSNR, while exhibiting a model size that is only 1 Mb larger. While SplaTAM and Photo-SLAM achieve higher SSIM and LPIPS, respectively, their model sizes exceed ours by a factor of 35 and 4, respectively. Finally, on the ScanNet++ dataset (Tab.~\ref{scannet++}), RP-SLAM consistently outperforms all baseline methods in both the training views and the novel views, and maintains the lowest model size.

The superior rendering quality observed in these datasets can be attributed to the combined effect of Efficient Incremental Mapping (EIM) and Dynamic Keyframe Window (DKW) optimization. Adaptive sampling in EIM, guided by the image gradient, ensures that regions with high-frequency textures receive sufficient detail while reducing redundancy in regions of lesser importance and ensures the efficient allocation of computational resources. Furthermore, the DKW employs the use of co-visible keyframes to maintain consistency across keyframes, thus ensuring the robustness of the reconstructed scene and, to a certain extent, addressing the issue of forgetting that can occur during successive optimization processes. Collectively, these two modules enable RP-SLAM to generate realistic and consistent scene representations.

Furthermore, for the monocular case (Tab.~\ref{mono_replica} and \ref{tum}), RP-SLAM attains a PSNR of 35.31, an SSIM of 0.940, and an LPIPS of 0.067 on the Replica dataset, and on the TUM dataset it achieves a PSNR of 21.89, an SSIM of 0.755 and an LPIPS of 0.209, consistently outperforming state-of-the-art monocular methods. These results demonstrate that accurate initialization of Gaussian primitives using the sparse point cloud generated by ORB-SLAM3 provides a reliable foundation for subsequent optimization. Concurrently, EIM and DKW also facilitate efficient scene representation and consistent scene optimization for the monocular case. This combination allows RP-SLAM to overcome the inherent limitations of monocular SLAM, such as geometric ambiguity and lack of depth cues, to produce detailed and realistic reconstructions.

The rendering results for all datasets demonstrate the efficacy of the core modules of RP-SLAM. The efficiency of image sampling and map representation is enhanced through the implementation of EIM, while ensuring the retention of essential details within complex regions. Concurrently, the DKW addresses the issue of forgetting and ensures the consistency of the reconstructed scene, thus enabling RP-SLAM to maintain robust performance over longer sequences. In monocular case, the MKI ensures accurate initialization of Gaussian primitives, thereby providing the requisite geometric basis for high-fidelity mapping. Collectively, these modules enable RP-SLAM to achieve state-of-the-art rendering quality in both RGB-D and monocular cases.

\subsection{Computational and Storage Efficiency}
The computational and storage efficiency of RP-SLAM is demonstrated through a comprehensive evaluation of the Replica\cite{straub2019replica}, TUM\cite{sturm2012benchmark}, and ScanNet++\cite{yeshwanth2023scannet++} datasets, as detailed in Tab.~\ref{mono_replica} to \ref{scannet++}. The results demonstrate that RP-SLAM ensures photorealistic reconstruction while maintaining a compact model size and achieving a high frame rate in comparison to state-of-the-art methods.

The high frame rates achieved by RP-SLAM can be attributed to the decoupling of camera tracking from Gaussian primitives optimization. By employing ORB-SLAM3\cite{campos2021orb} for direct camera pose estimation, RP-SLAM circumvents the arduous iterative procedure necessitated for the optimization of the camera pose with respect to the scene representation. This separation permits the system to allocate computational resources in a more efficient manner, thereby achieving real-time performance. As illustrated in Tab.~\ref{mono_replica}, RP-SLAM attains an average frame rate of 17.3 FPS in monocular mode on the Replica dataset, which is considerably superior to that of MonoGS, which exhibits a lower frame rate due to the overhead of coupled optimization. Similarly, in RGB-D mode (Tab.~\ref{rgbd_replica}), RP-SLAM achieves an average frame rate of 18.3 FPS, which outperforms MonoGS\cite{matsuki2024gaussian}, RTG-SLAM\cite{peng2024rtg}, and SplaTAM\cite{keetha2024splatam}. 

\begin{figure*}
  \centering
  \scriptsize
  \setlength{\tabcolsep}{0.5pt}
  \newcommand{\sz}{0.19}  %
  \begin{tabular}{lcccccc}
\vspace{-0.5mm}
    \makecell{\rotatebox{90}{GT}}                               &
    \makecell{\includegraphics[width=\sz\linewidth]{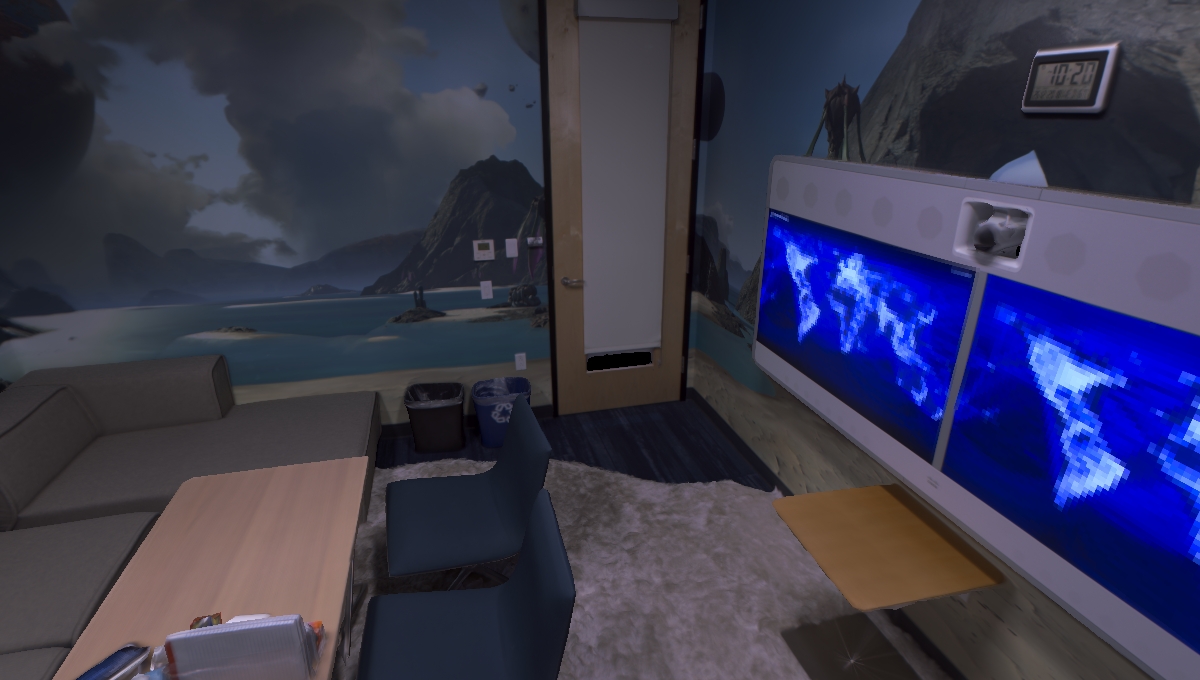}}        &
    \makecell{\includegraphics[width=\sz\linewidth]{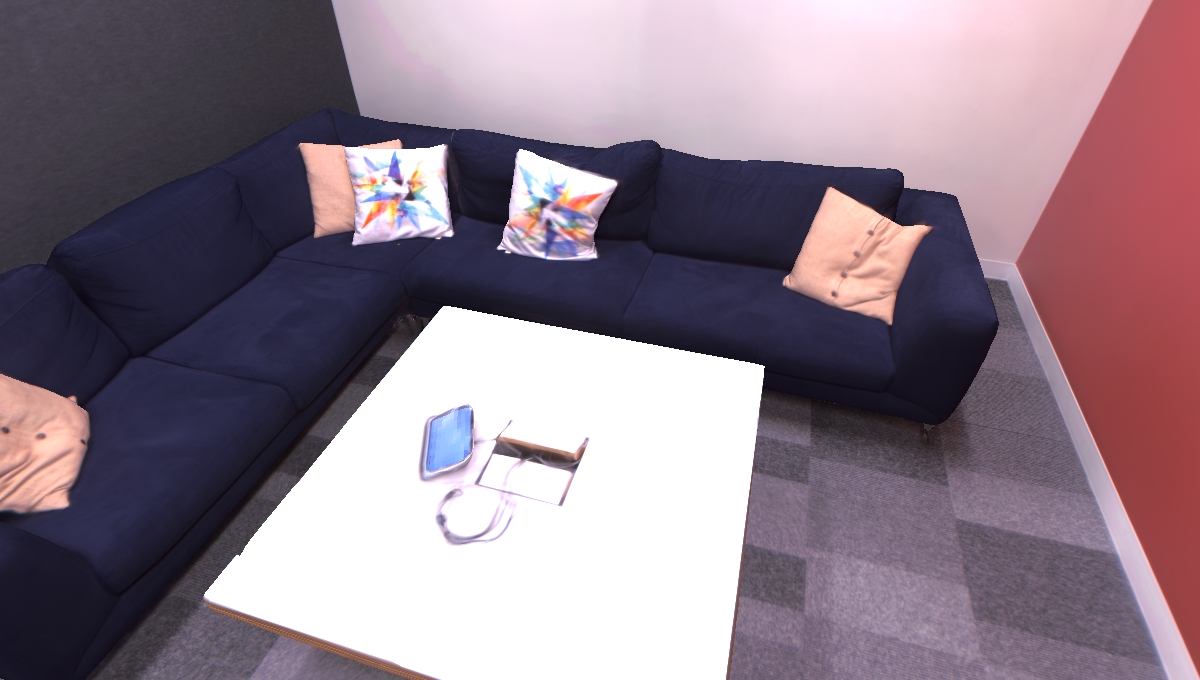}} &
    \makecell{\includegraphics[width=\sz\linewidth]{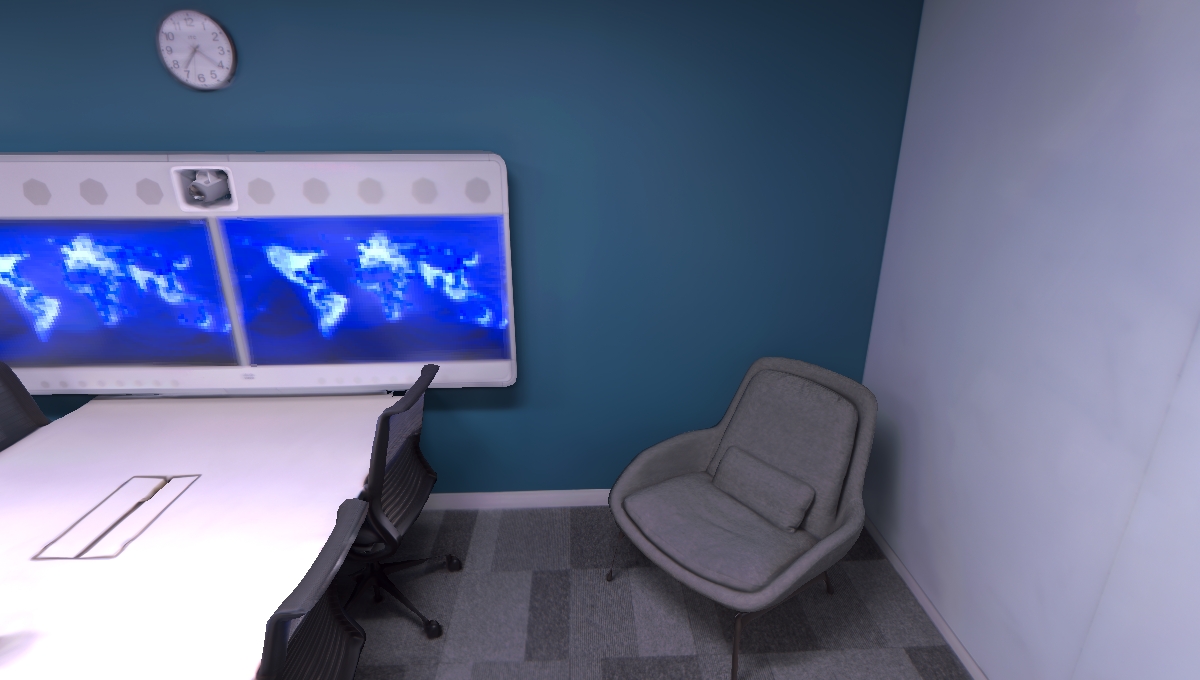}}   &
    \makecell{\includegraphics[width=\sz\linewidth]{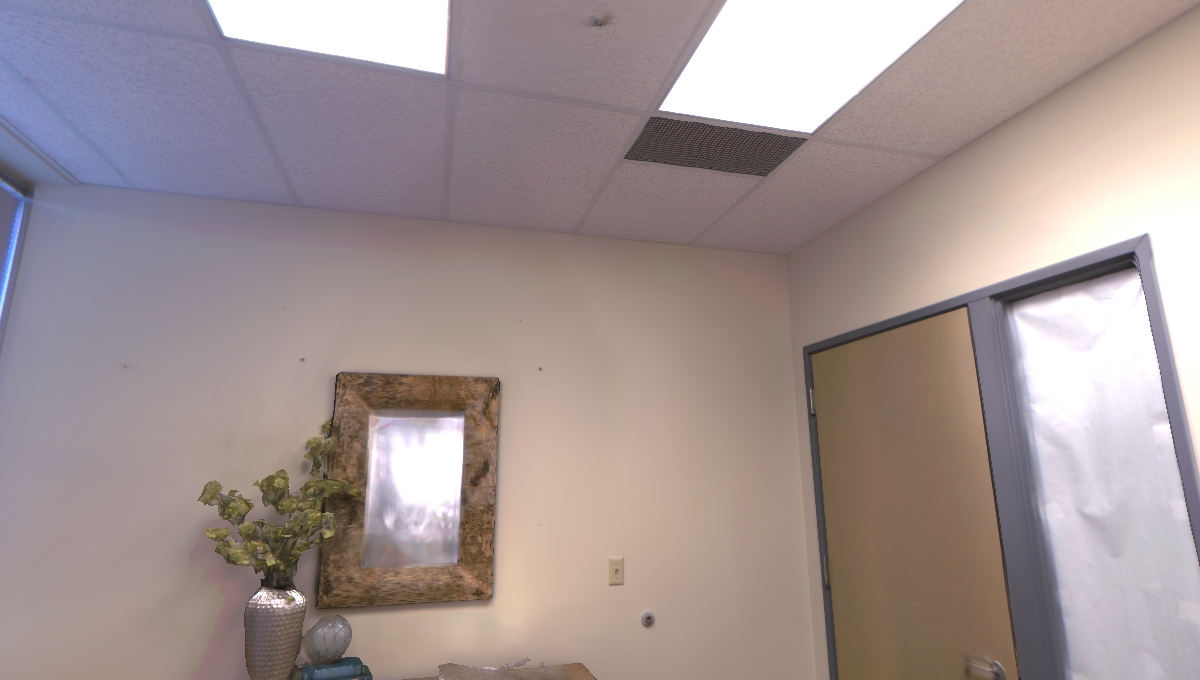}} &
    \makecell{\includegraphics[width=\sz\linewidth]{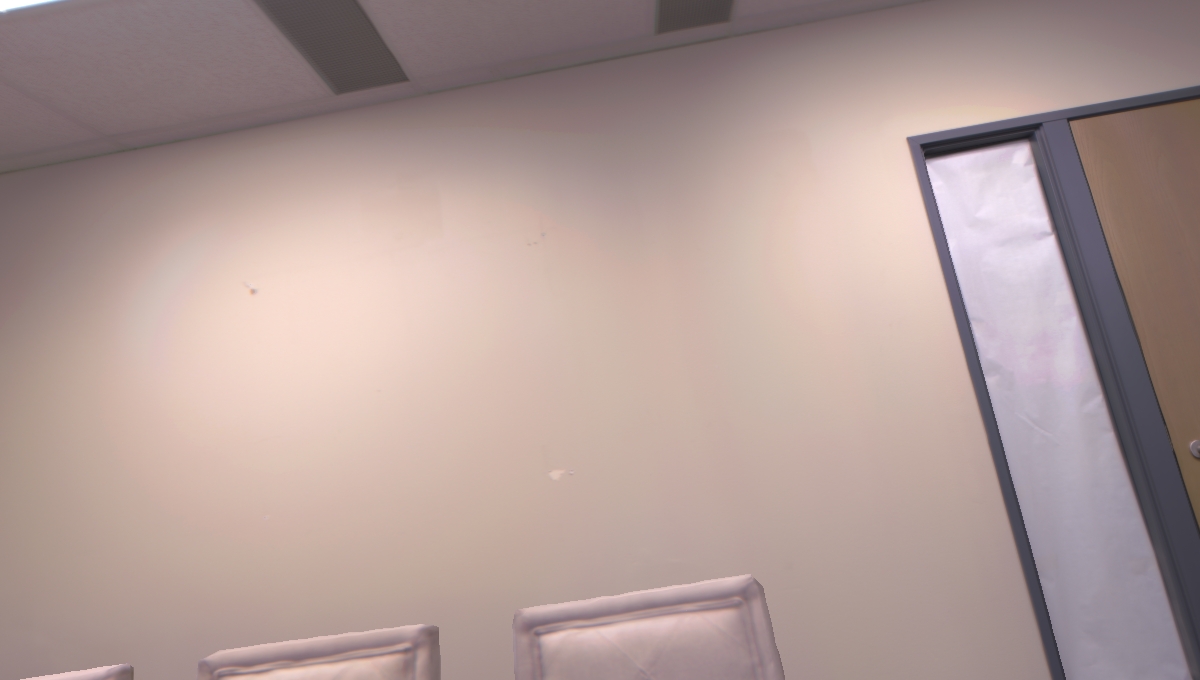}}       \\
\vspace{-0.5mm}
    \makecell{\rotatebox{90}{MonoGS~\cite{matsuki2024gaussian}}}                               &
    \makecell{\includegraphics[width=\sz\linewidth]{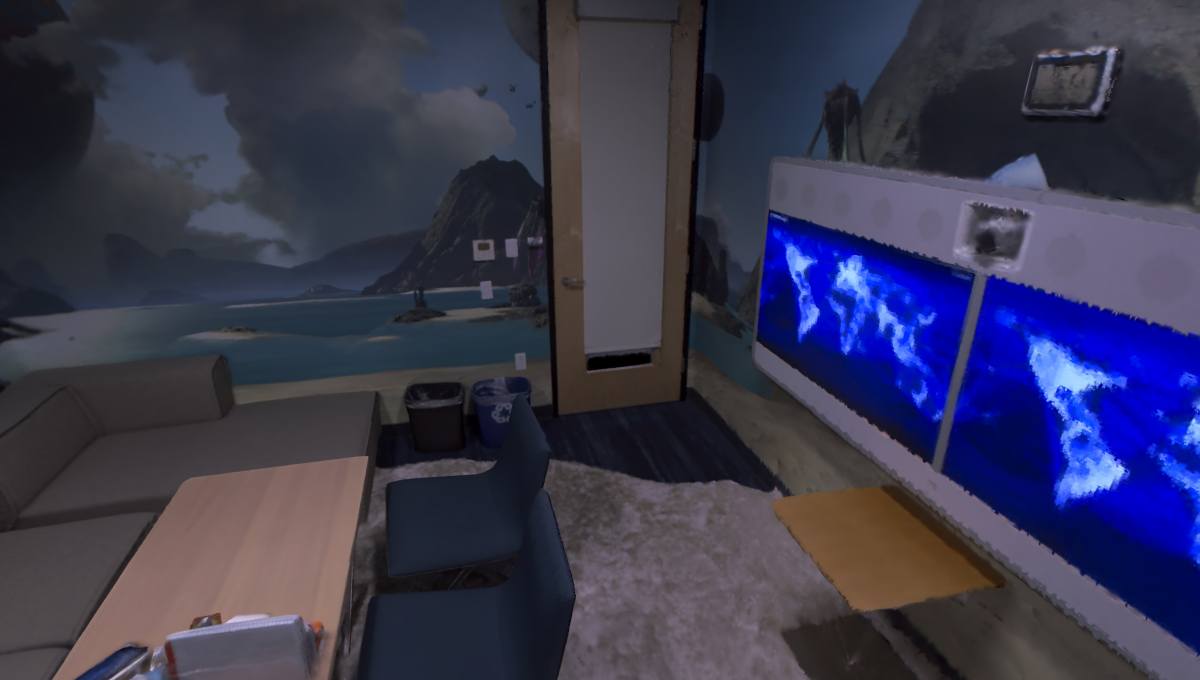}}        &
    \makecell{\includegraphics[width=\sz\linewidth]{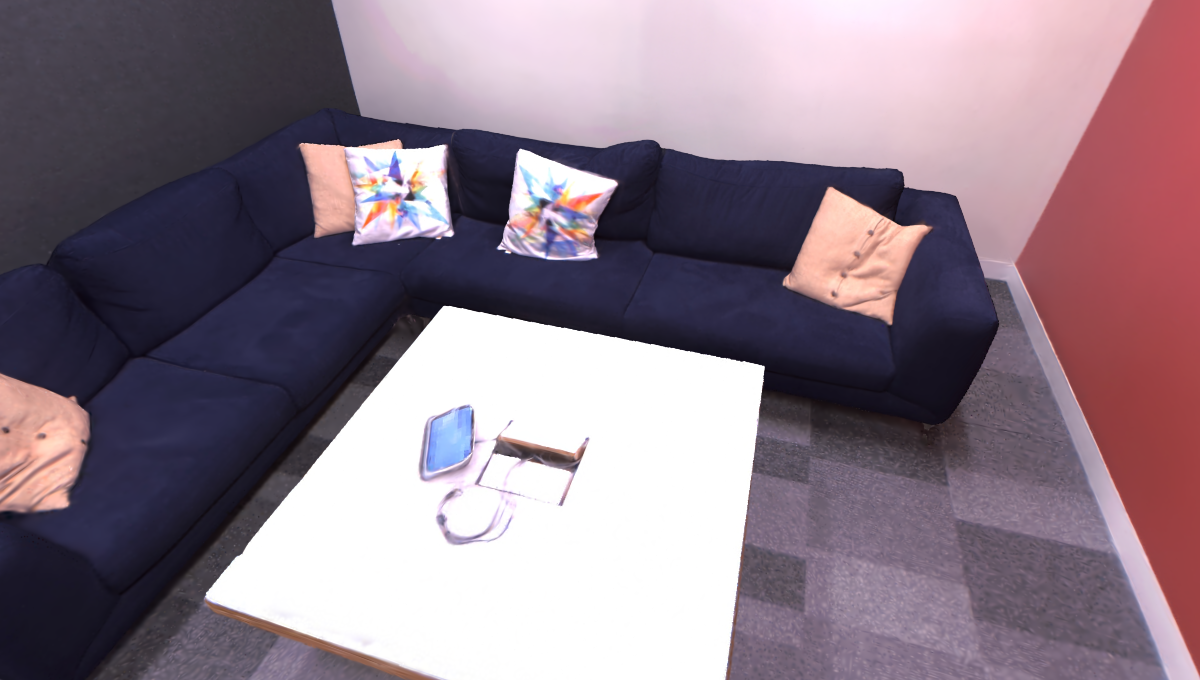}} &
    \makecell{\includegraphics[width=\sz\linewidth]{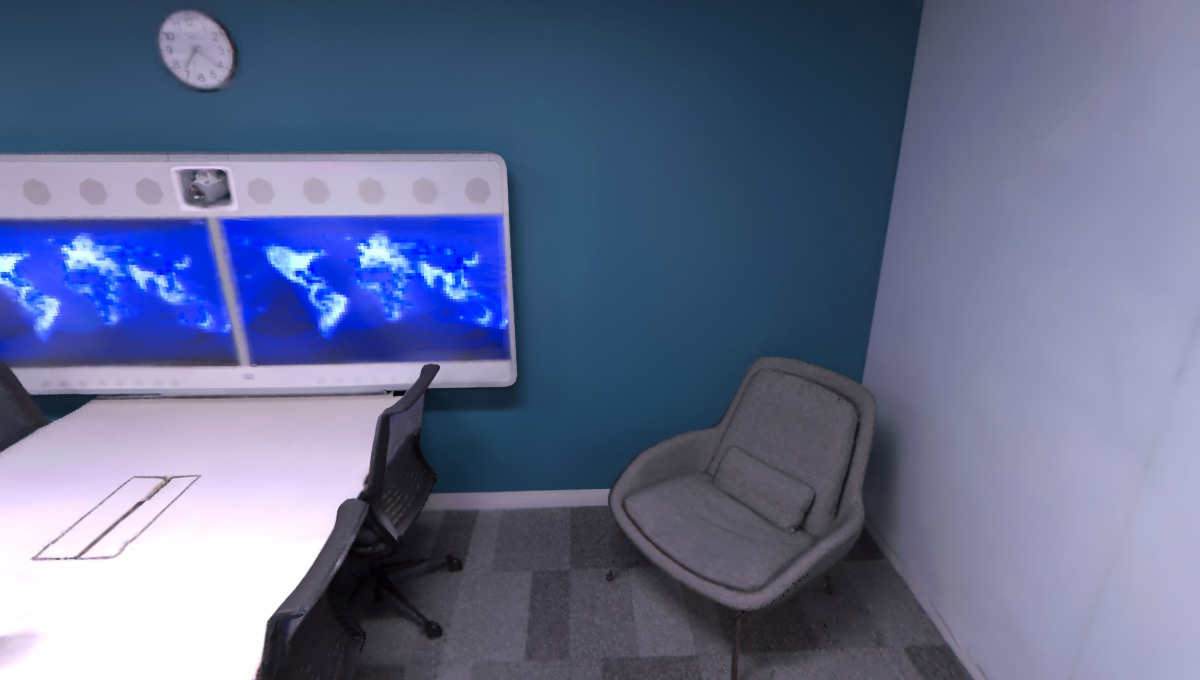}}   &
    \makecell{\includegraphics[width=\sz\linewidth]{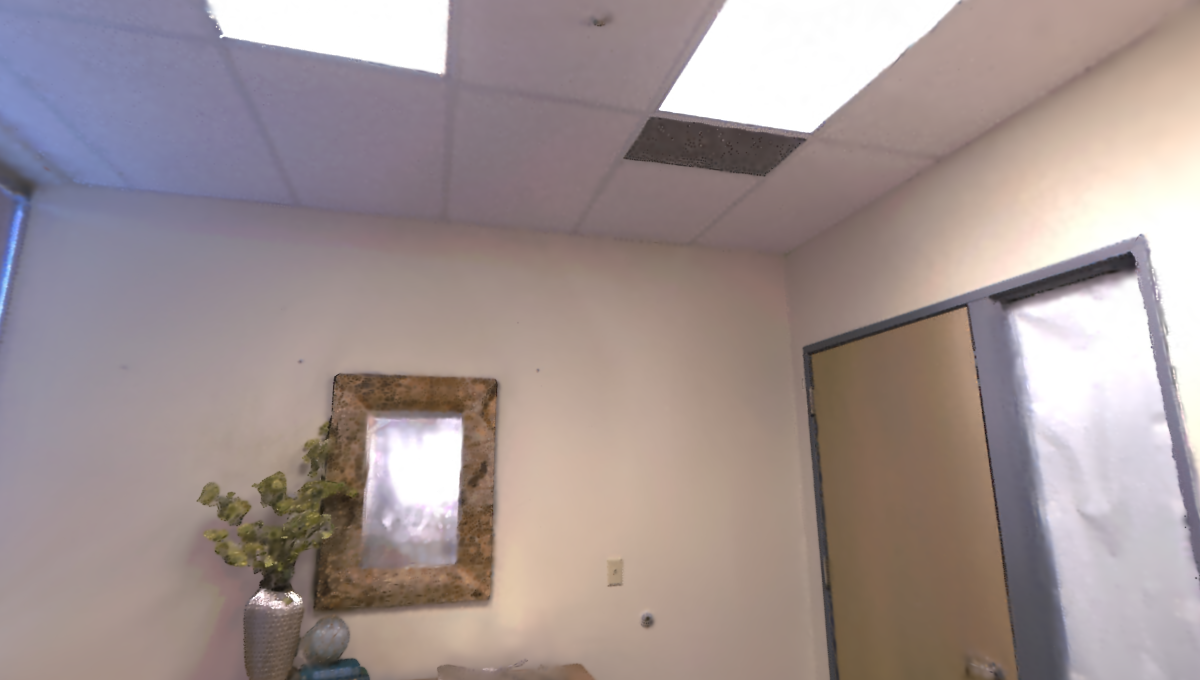}} &
    \makecell{\includegraphics[width=\sz\linewidth]{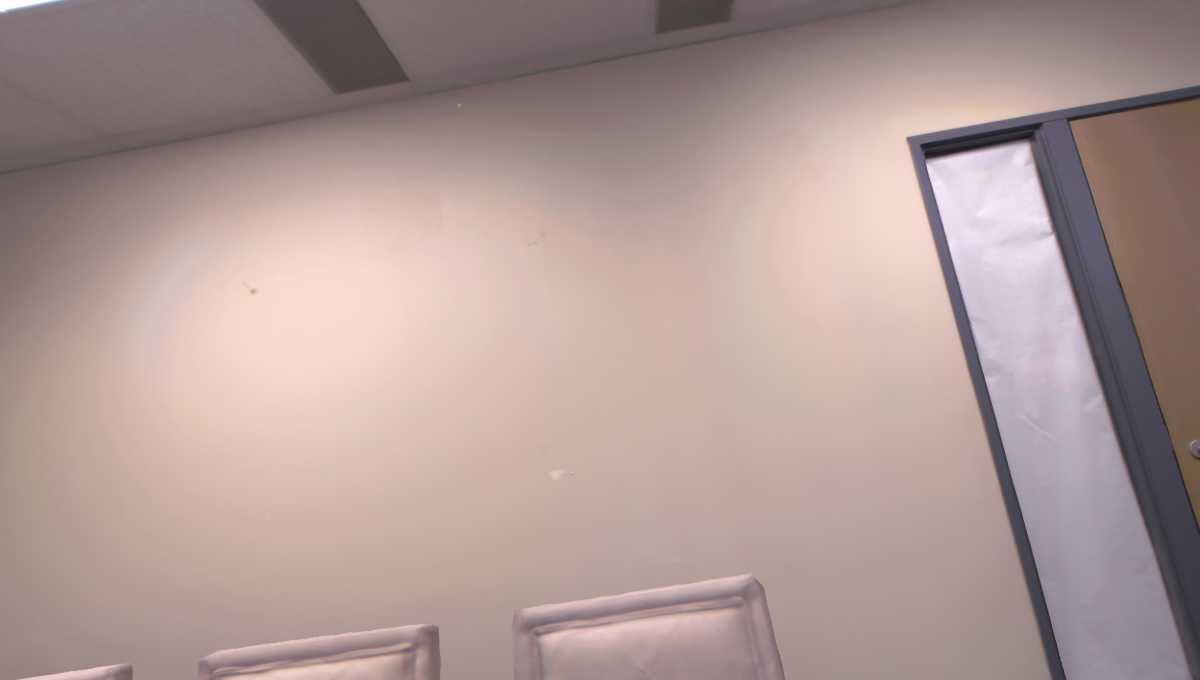}}        \\
\vspace{-0.5mm}
    \makecell{\rotatebox{90}{Photo-SLAM~\cite{huang2024photo}}}                               &
    \makecell{\includegraphics[width=\sz\linewidth]{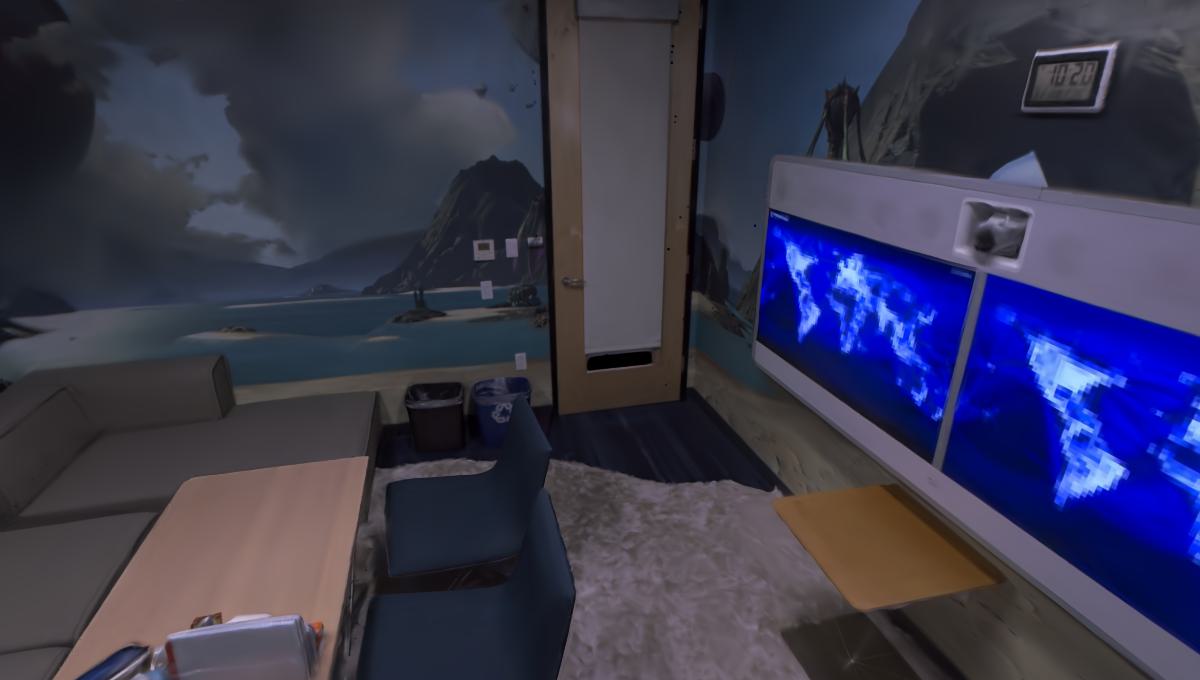}}          &
    \makecell{\includegraphics[width=\sz\linewidth]{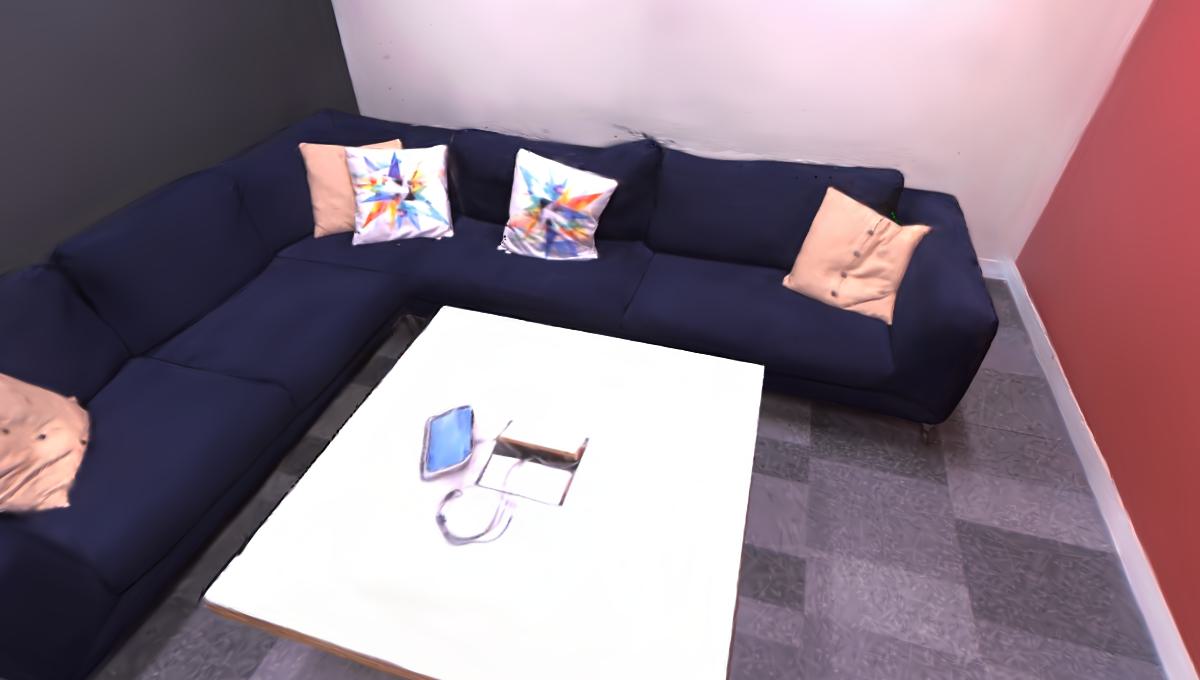}}   &
    \makecell{\includegraphics[width=\sz\linewidth]{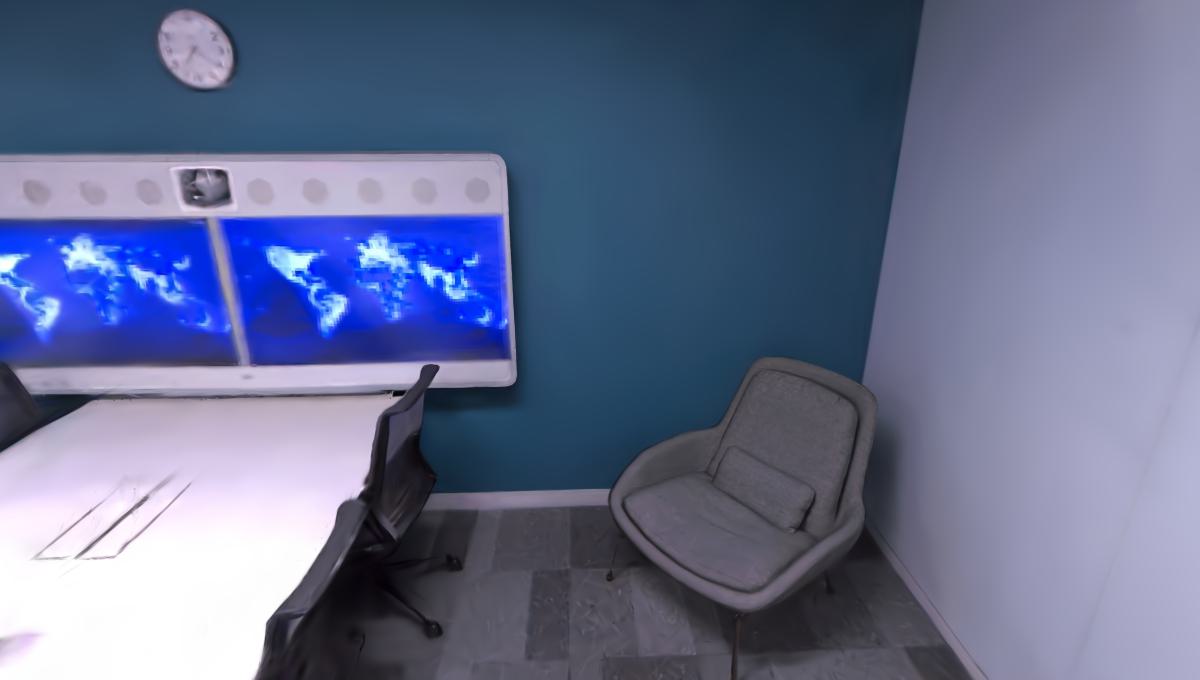}}     &
    \makecell{\includegraphics[width=\sz\linewidth]{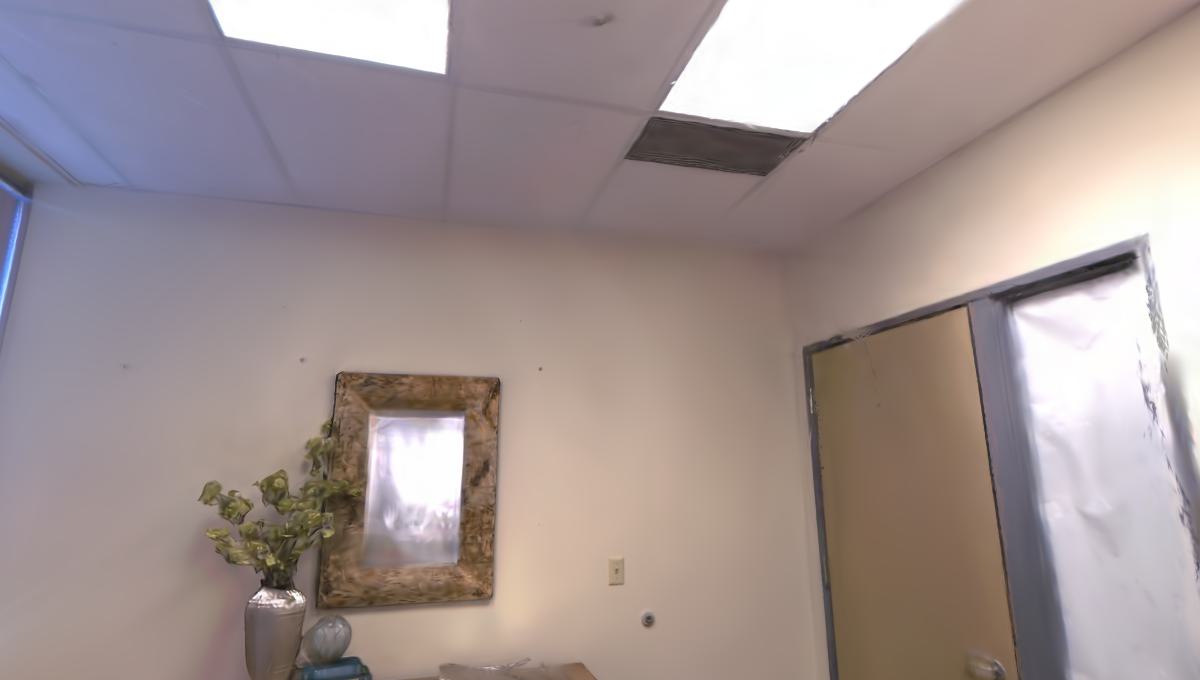}}   &
    \makecell{\includegraphics[width=\sz\linewidth]{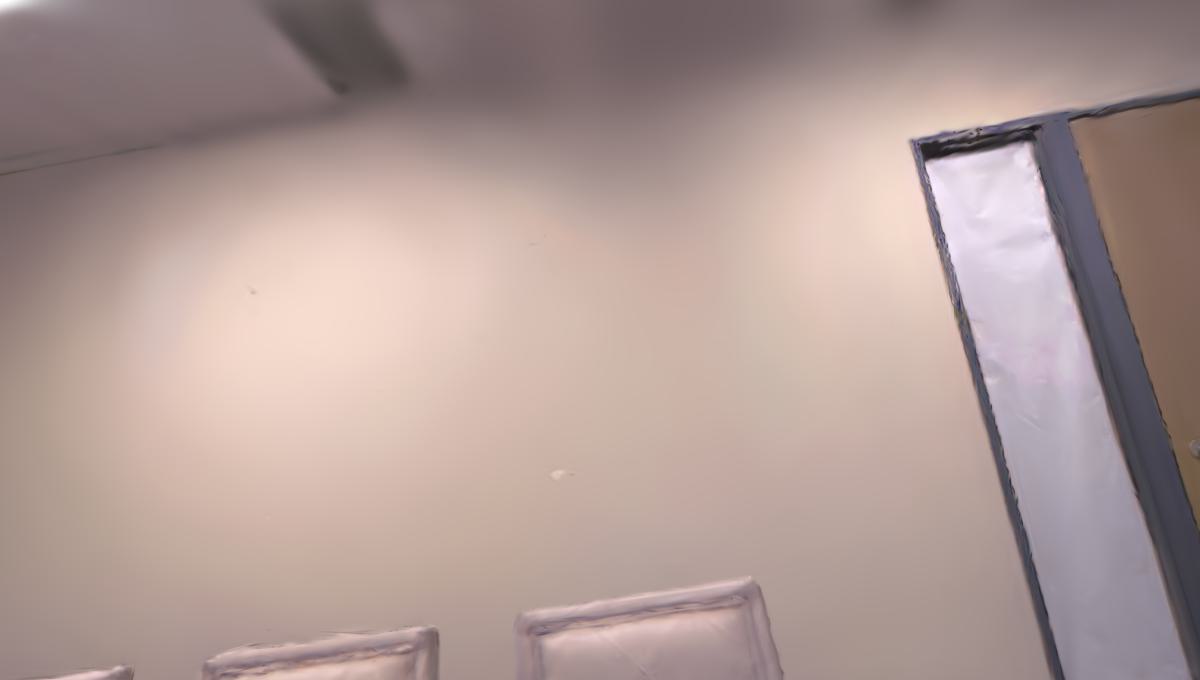}}          \\
\vspace{-0.5mm}
    \makecell{\rotatebox{90}{RTG-SLAM~\cite{peng2024rtg}}}                               &
    \makecell{\includegraphics[width=\sz\linewidth]{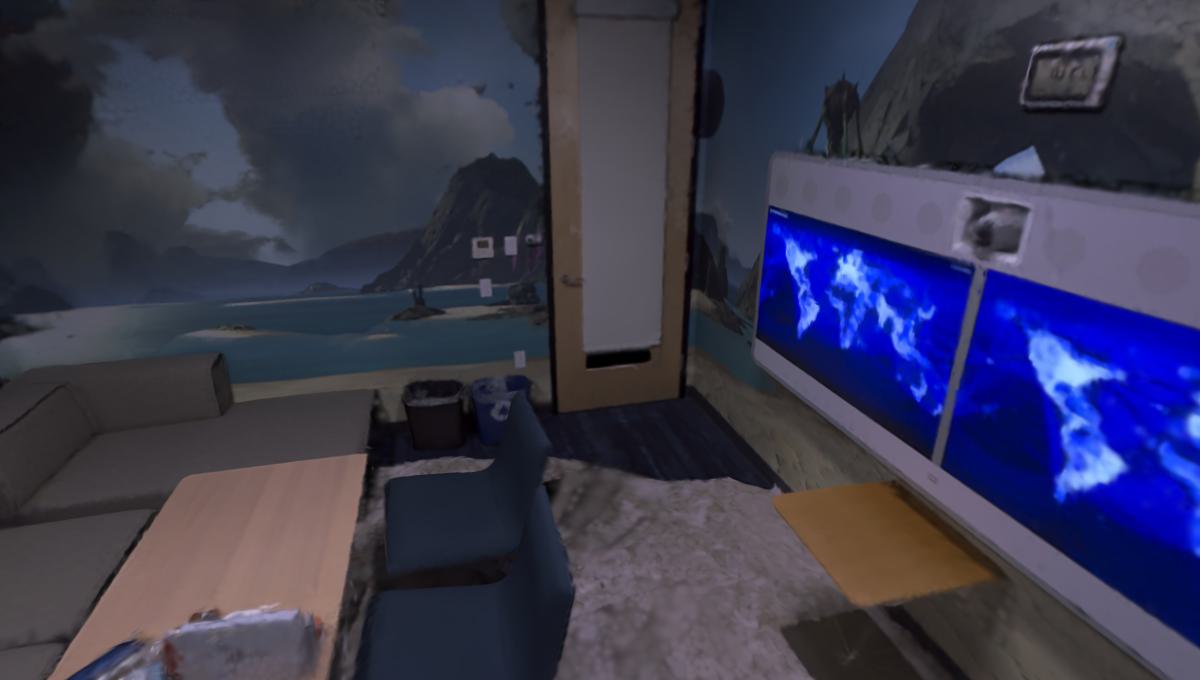}}          &
    \makecell{\includegraphics[width=\sz\linewidth]{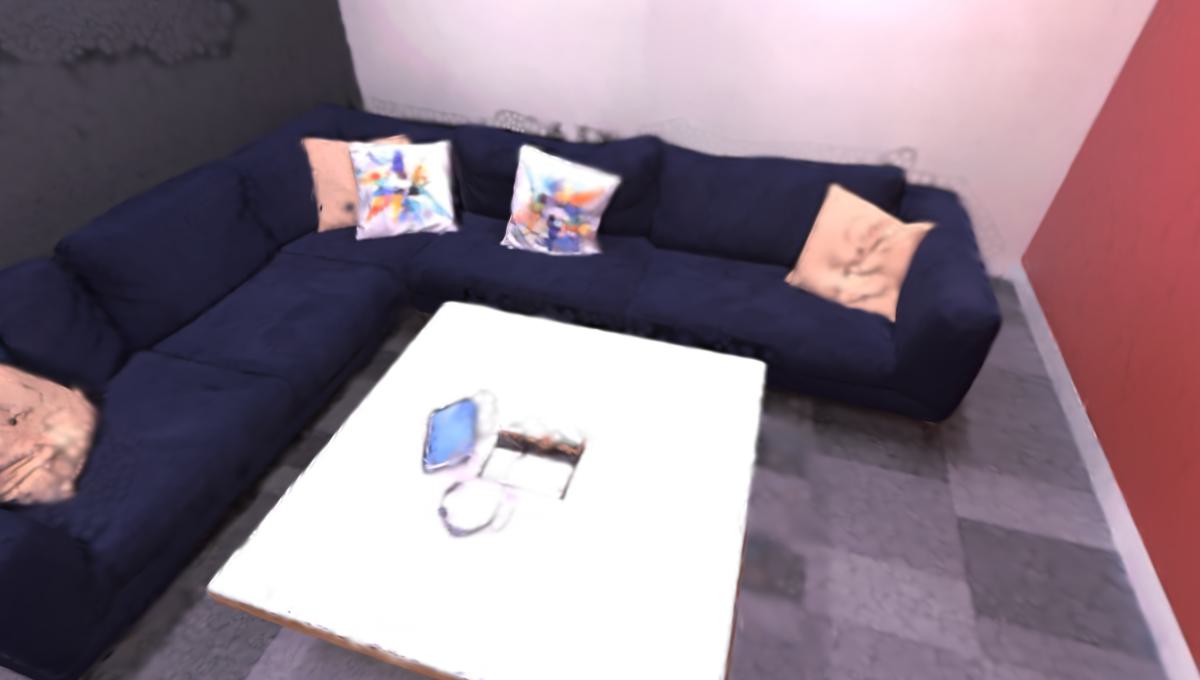}}   &
    \makecell{\includegraphics[width=\sz\linewidth]{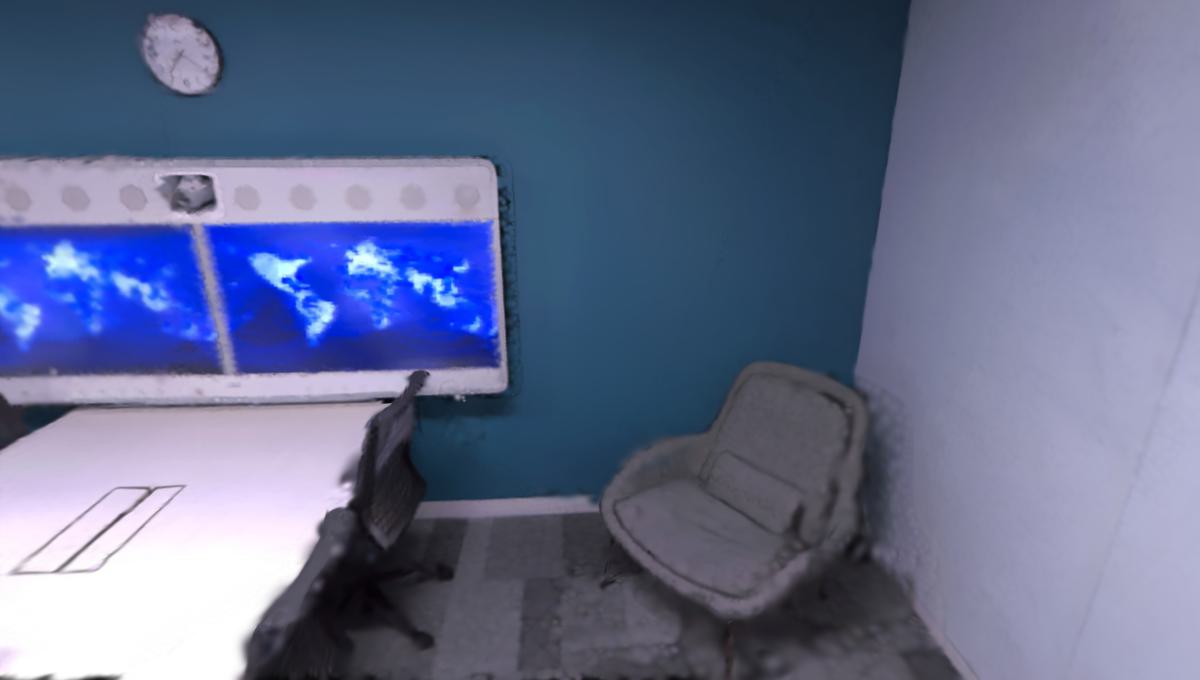}}     &
    \makecell{\includegraphics[width=\sz\linewidth]{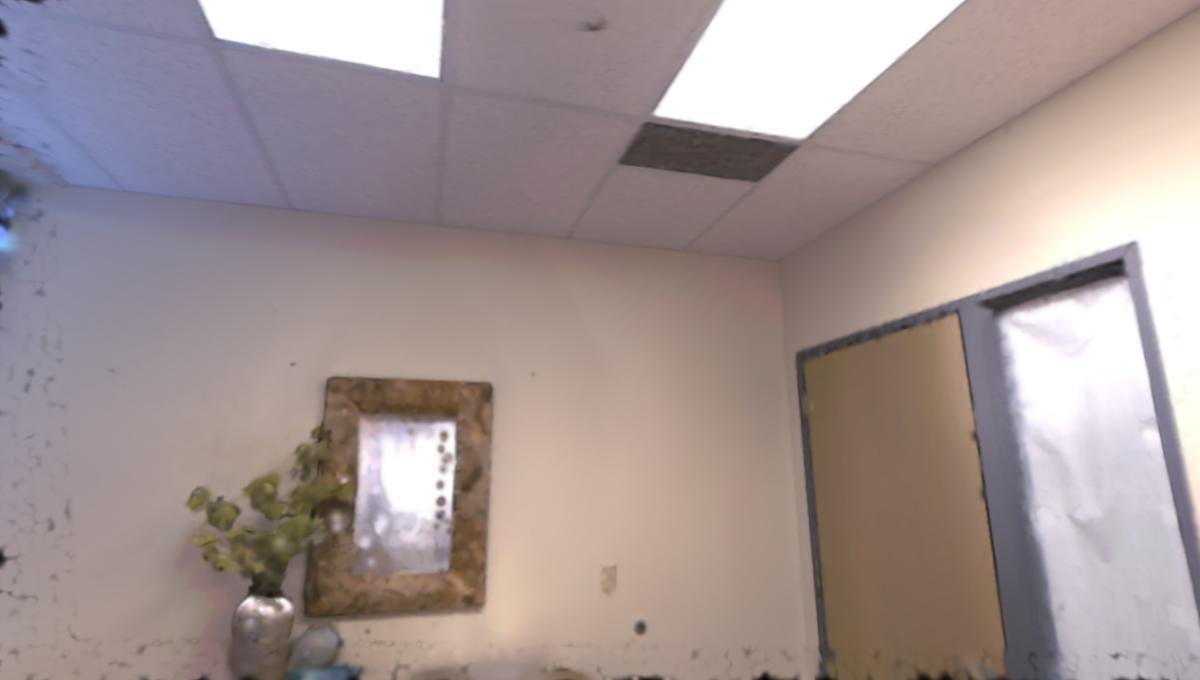}}   &
    \makecell{\includegraphics[width=\sz\linewidth]{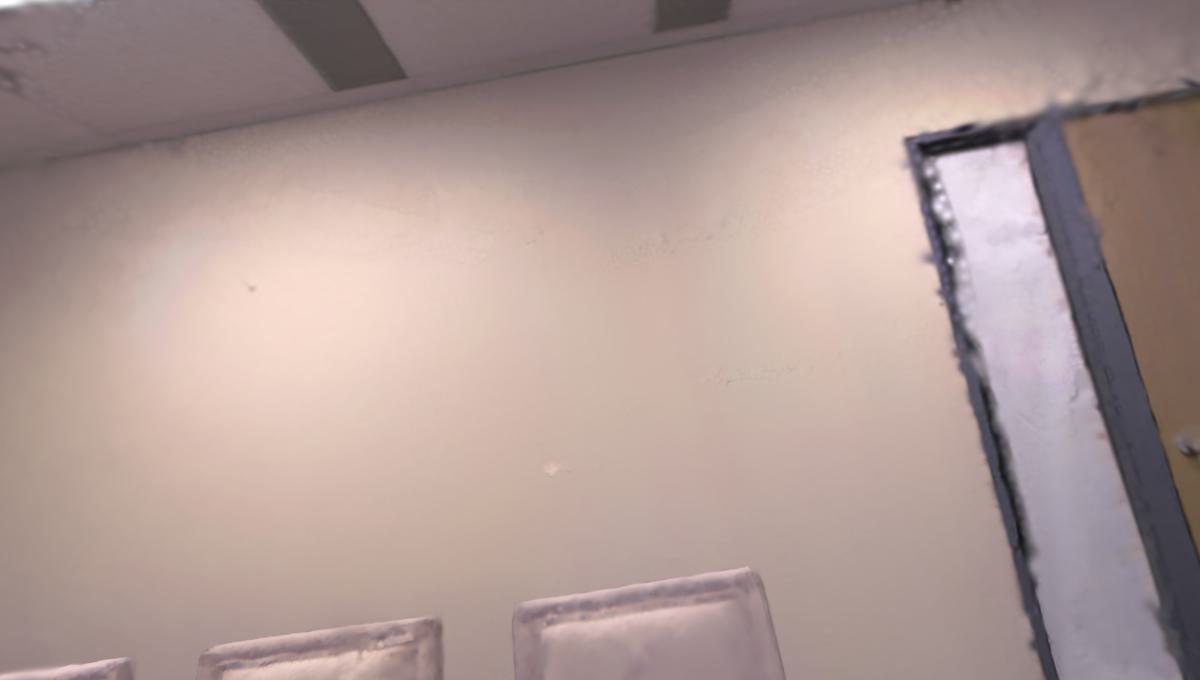}}         \\
\vspace{-0.5mm}
    \makecell{\rotatebox{90}{CaRtGS~\cite{feng2024cartgs}}}                            &
    \makecell{\includegraphics[width=\sz\linewidth]{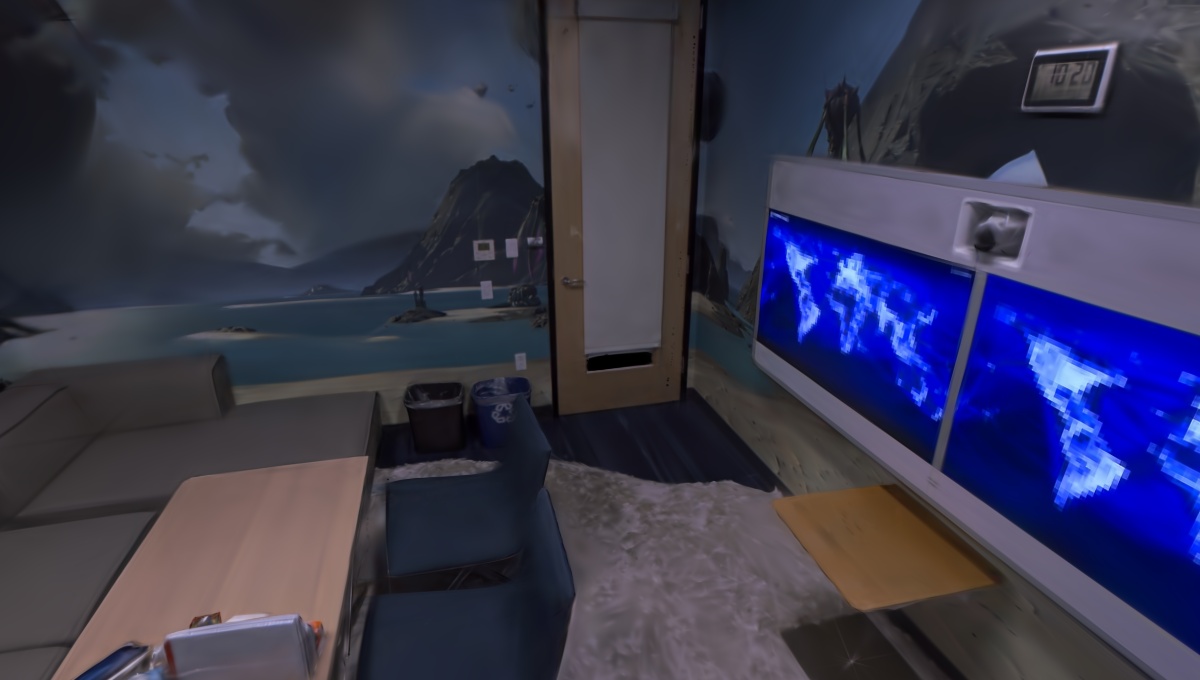}}         &
    \makecell{\includegraphics[width=\sz\linewidth]{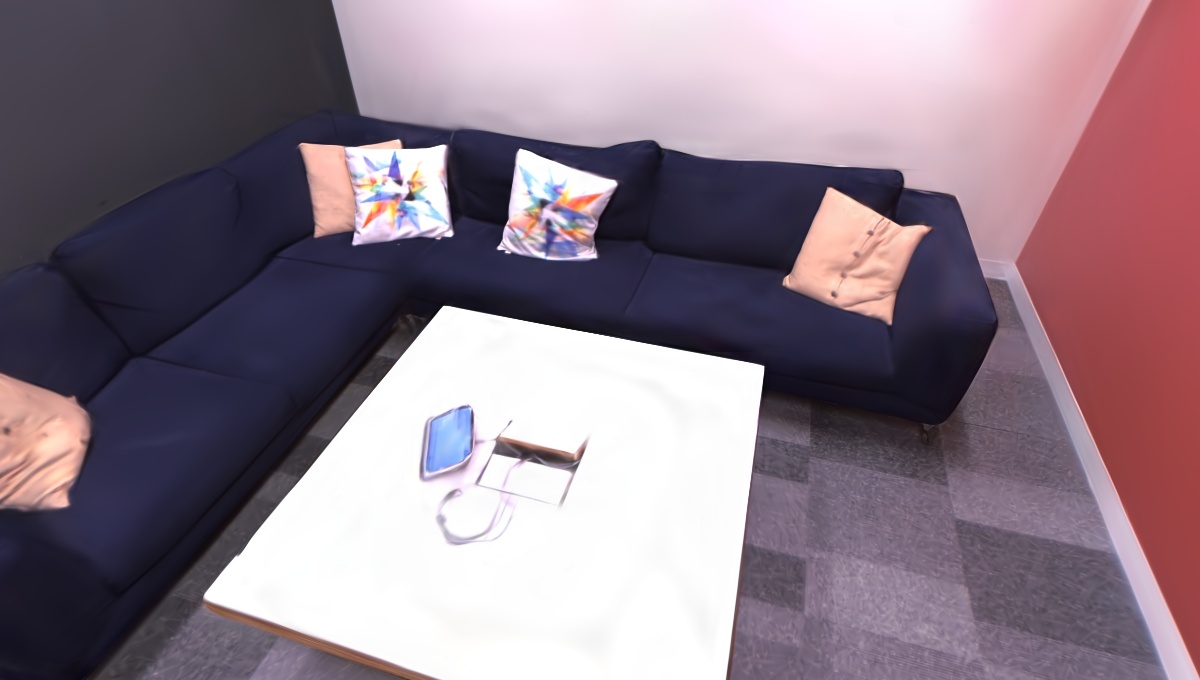}}  &
    \makecell{\includegraphics[width=\sz\linewidth]{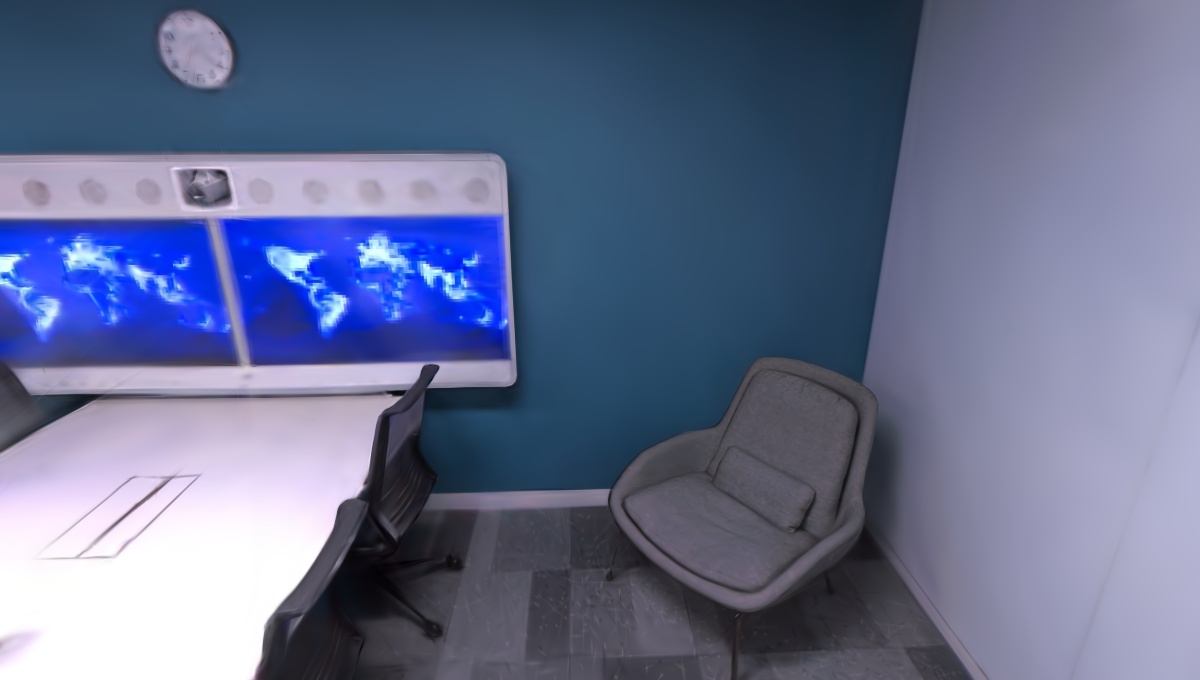}}    &
    \makecell{\includegraphics[width=\sz\linewidth]{figs/rendering/rgbd/r1_photo.jpg}}  &
    \makecell{\includegraphics[width=\sz\linewidth]{figs/rendering/rgbd/r2_photo.jpg}}        \\
\vspace{-0.5mm}
    \makecell{\rotatebox{90}{RP-SLAM (ours)}}                           &
    \makecell{\includegraphics[width=\sz\linewidth]{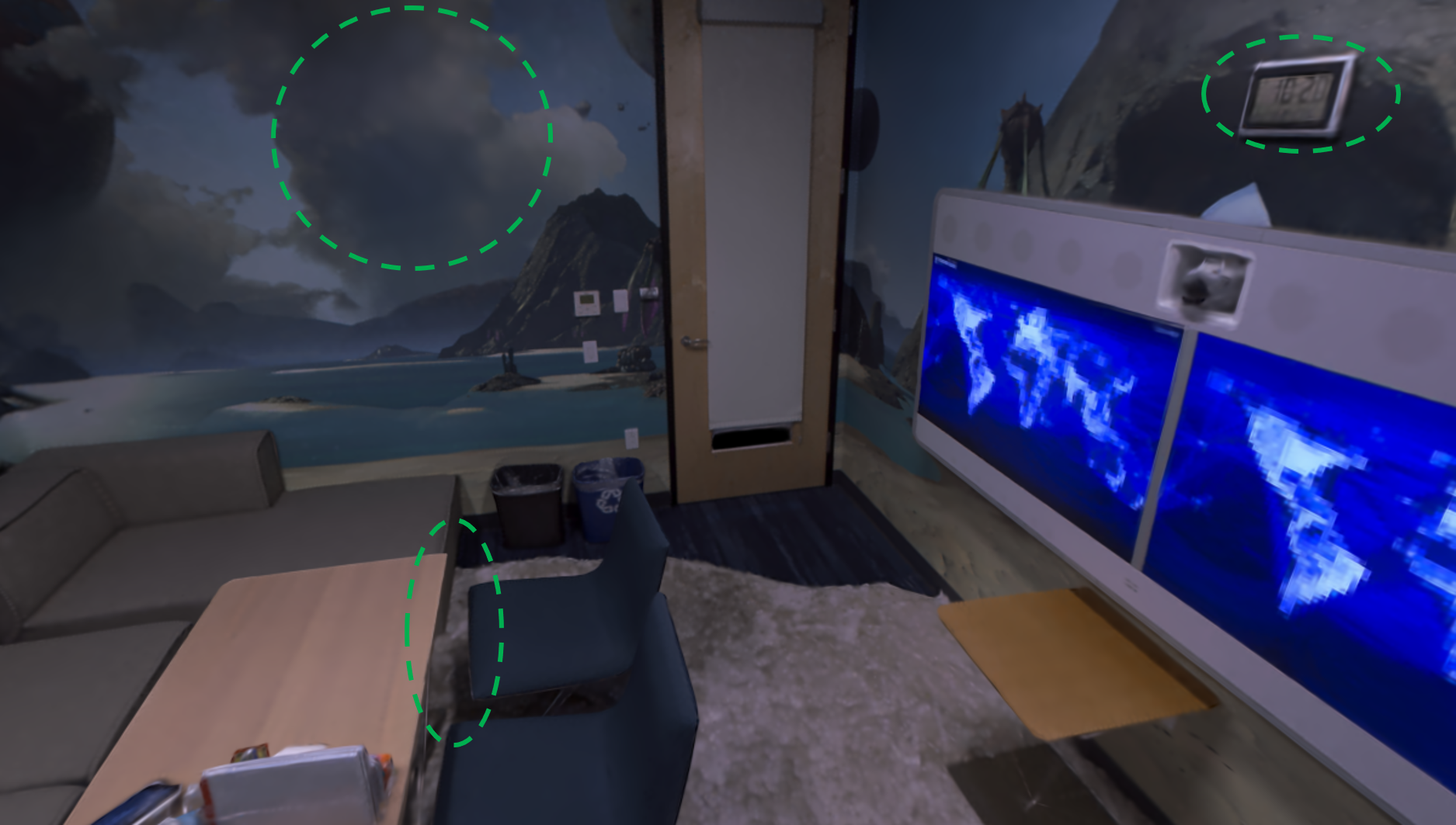}}         &
    \makecell{\includegraphics[width=\sz\linewidth]{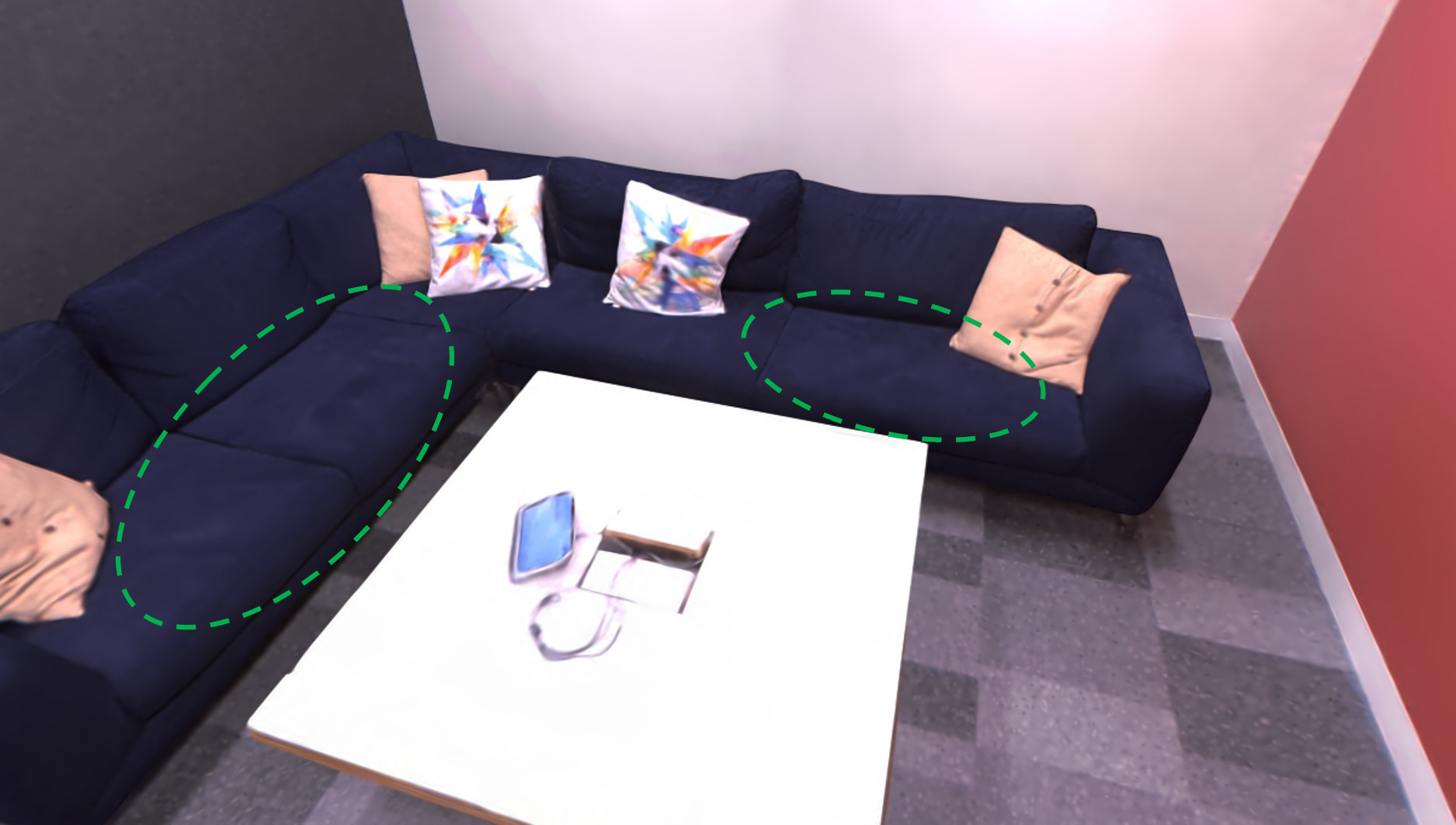}}  &
    \makecell{\includegraphics[width=\sz\linewidth]{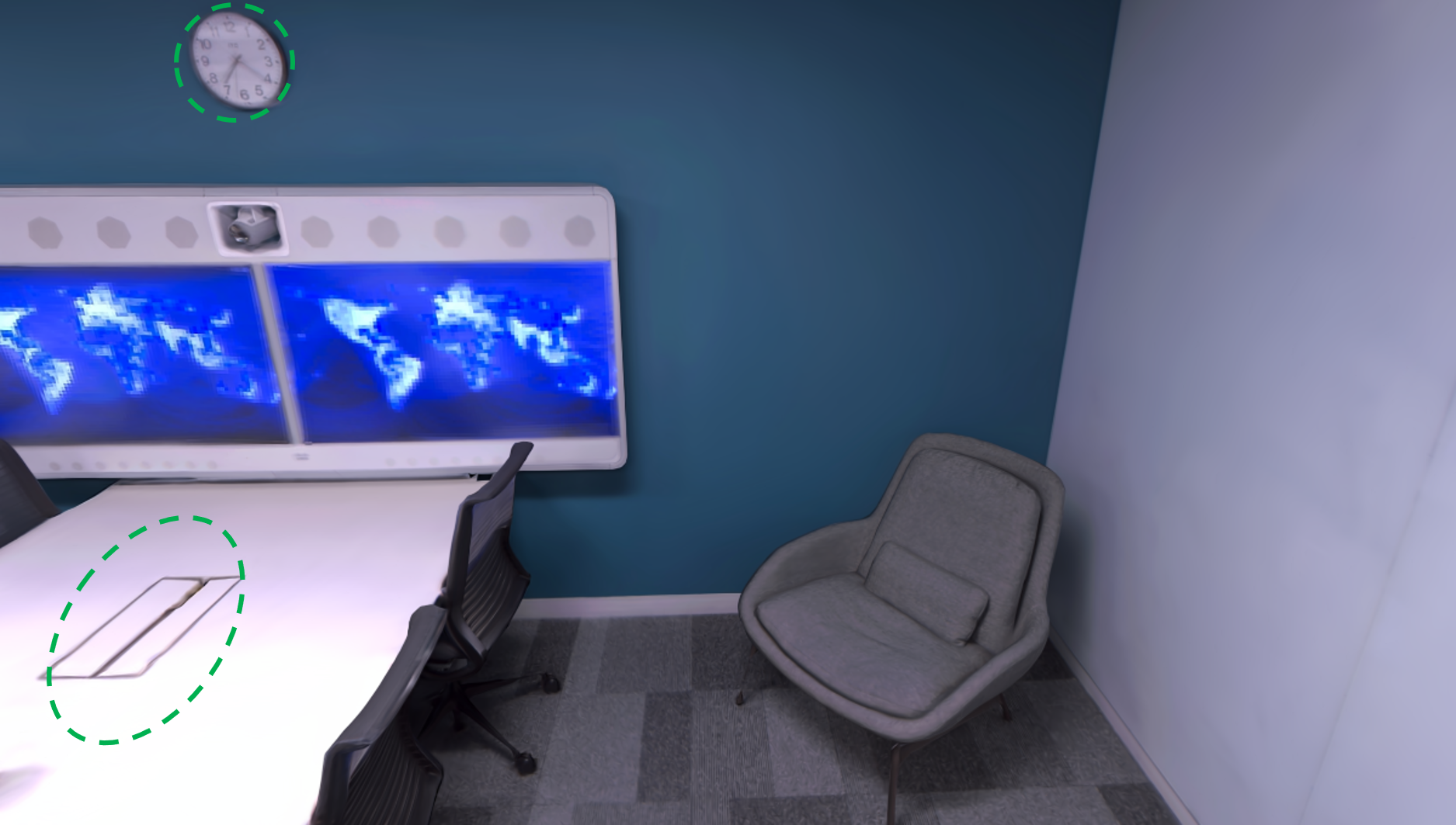}}    &
    \makecell{\includegraphics[width=\sz\linewidth]{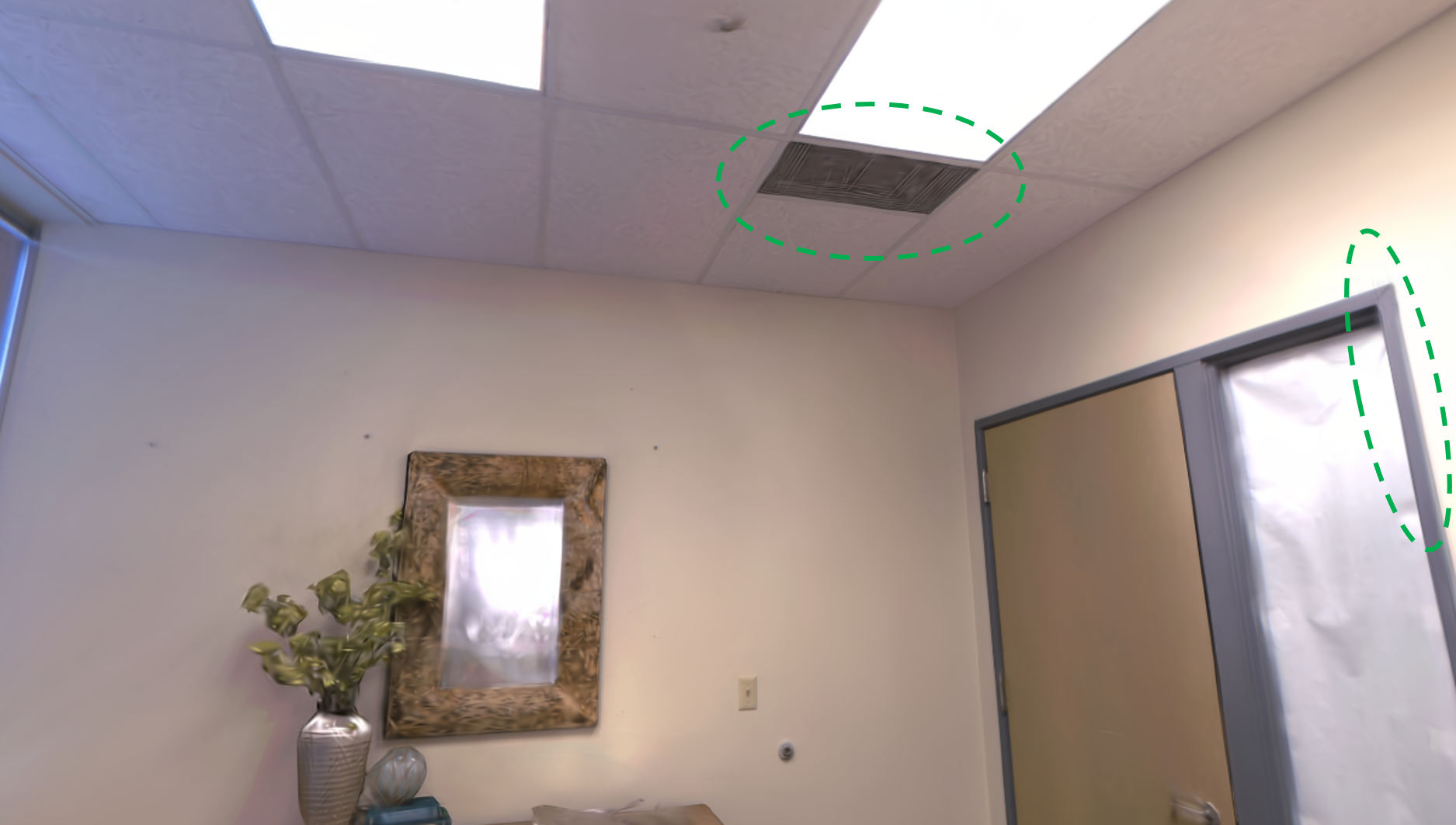}}  &
    \makecell{\includegraphics[width=\sz\linewidth]{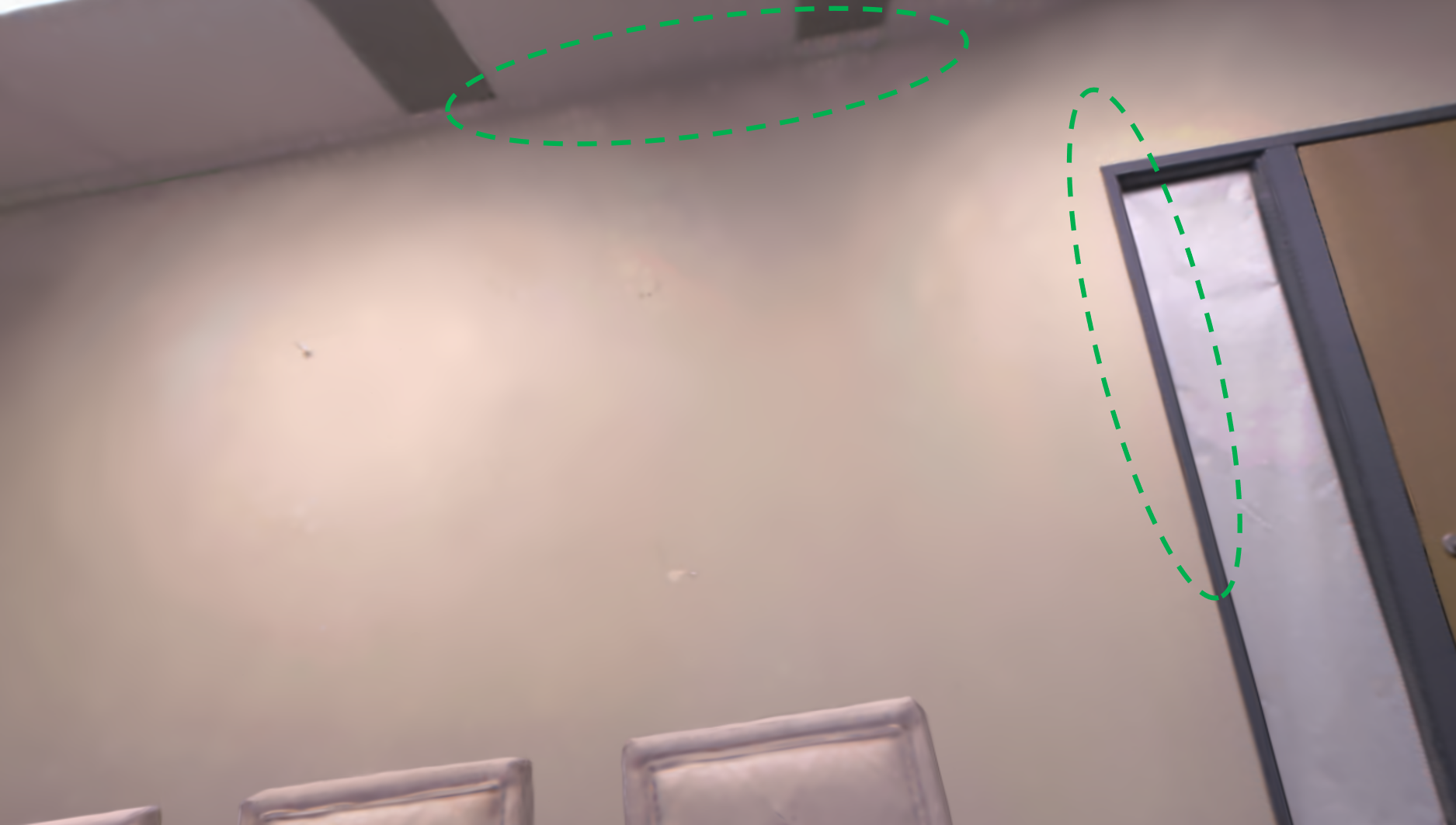}}         \\
                                                                                       &
    \makecell{(a)}                                                                     &
    \makecell{(b)}                                                                     &
    \makecell{(c)}                                                                     &
    \makecell{(d)}                                                                     &
    \makecell{(e)}                                                                       \\
  \end{tabular}
  \caption{Qualitative comparisons on Replica\cite{straub2019replica} dataset in the RGBD case. The green dashed boxes in our method mark areas where RP-SLAM outperforms other methods, such as sharper textures and fewer artefacts. Zoom in for a clearer view.}
  \label{render_rgbd}
\end{figure*}

The compact model size of RP-SLAM is attributed to EIM, which represents the scene in an efficient manner by reducing redundancy without compromising the rendering quality. EIM employs quadtree-based adaptive sampling oriented to the image gradient, which ensures that regions with a high level of detail are adequately sampled while avoiding unnecessary repetitive sampling in regions with a more even texture. Furthermore, the implementation of KNN-besed filtering facilitates the refinement of the Gaussian primitives set through the elimination of redundancy, thereby enhancing the efficiency of the map representation. As illustrated in Tab.~\ref{mono_replica}, the average model size of RP-SLAM in monocular mode on the Replica dataset is 11.3 Mb, which is notably smaller than that of Photo-SLAM\cite{huang2024photo} (23.14 Mb) and that of CaRtGS\cite{feng2024cartgs} (15.3 Mb). In RGB-D mode (Tab.~\ref{rgbd_replica}), the average model size of RP-SLAM is 10.0 Mb, which is significantly smaller than that of all baseline methods. On the TUM dataset (Tab.~\ref{tum}), RP-SLAM maintains a compact model size of 6.3 Mb in monocular mode and 3.8 Mb in RGB-D mode, which outperforms all methods except MonoGS\cite{matsuki2024gaussian}. Similarly, the smallest model size was obtained on the ScanNet++ dataset (Tab.~\ref{scannet++}).

The combined impact of the decoupled sampling method and EIM is further emphasized when considering the balance between computational efficiency and rendering quality. Despite achieving higher frame rates and smaller model sizes, RP-SLAM does not compromise the quality of the reconstructed scene. This balance is critical for applications that require both real-time and high-fidelity reconstruction. In conclusion, the computational and storage efficiency of RP-SLAM is markedly enhanced by the decoupled method for camera pose estimation and scene representation optimization, in addition to EIM. Collectively, these approaches empower RP-SLAM to attain real-time operation at high frame rates while maintaining a compact model size, and furthermore, provide realistic and consistent scene reconstruction.

\begin{figure*}
  \centering
  \scriptsize
  \setlength{\tabcolsep}{0.5pt}
  \newcommand{\sz}{0.24}  %
  \begin{tabular}{lcccccc}
\vspace{-0.5mm}
    \makecell{\rotatebox{90}{GT}}                               &
    \makecell{\includegraphics[width=\sz\linewidth]{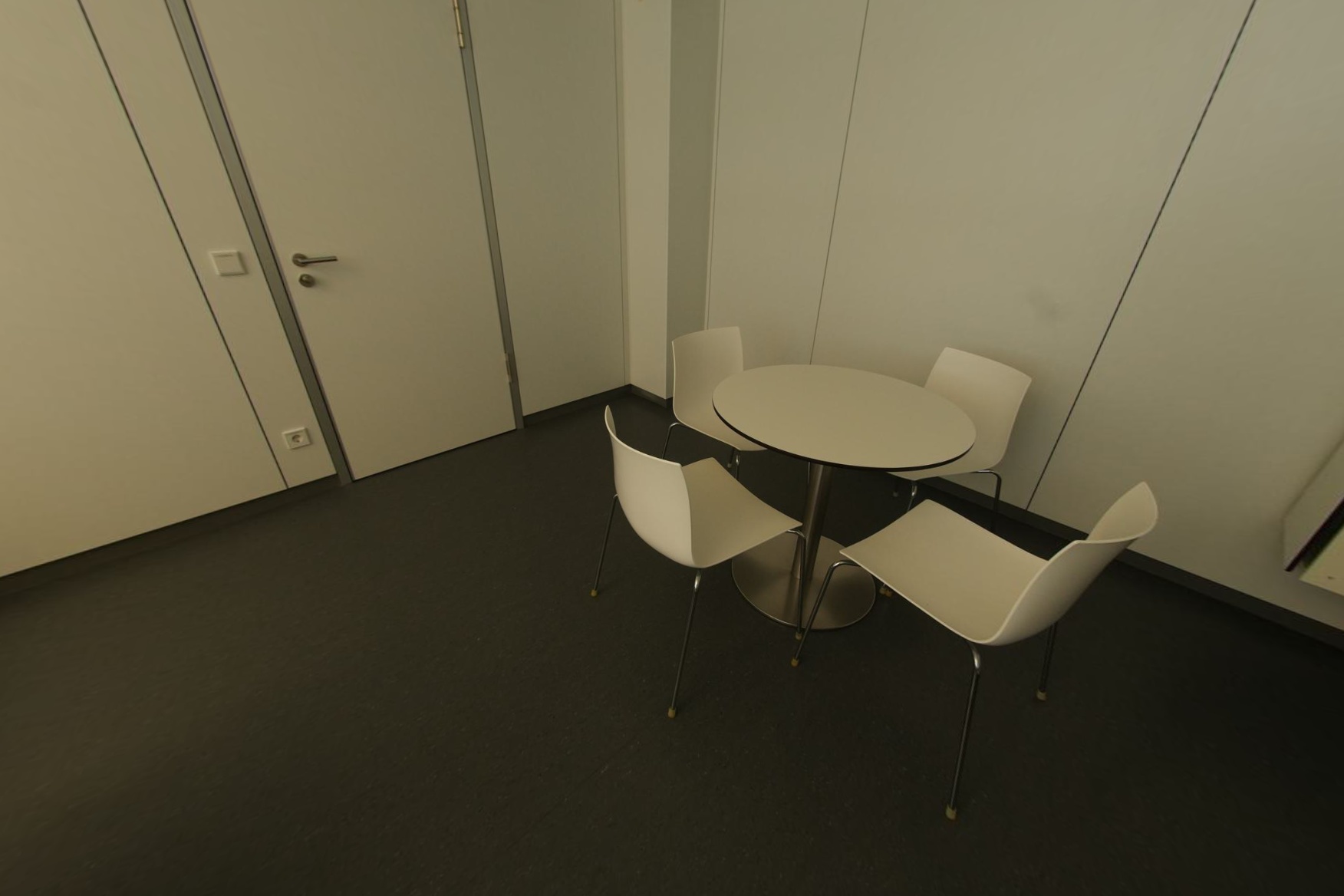}}        &
    \makecell{\includegraphics[width=\sz\linewidth]{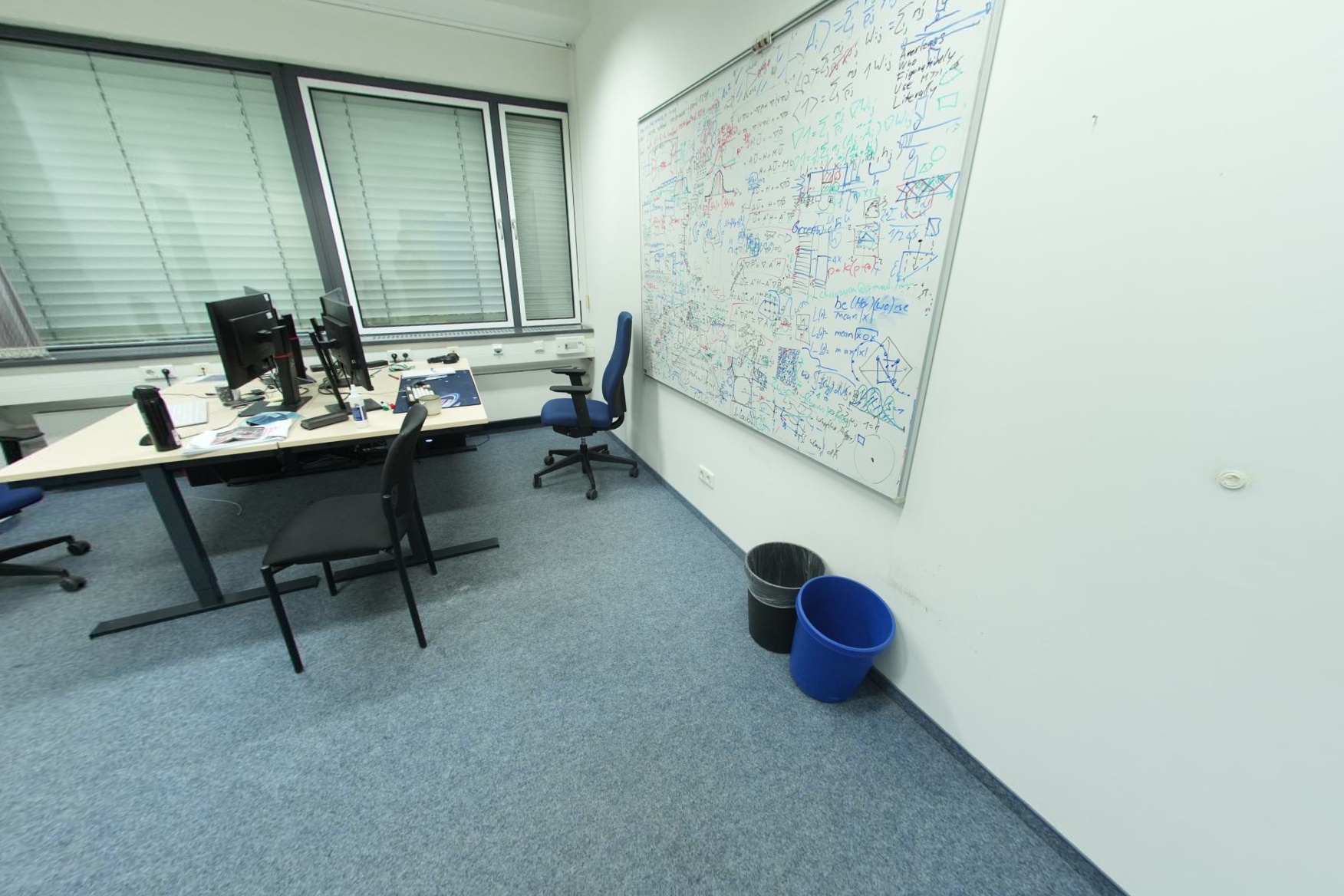}} &
    \makecell{\includegraphics[width=\sz\linewidth]{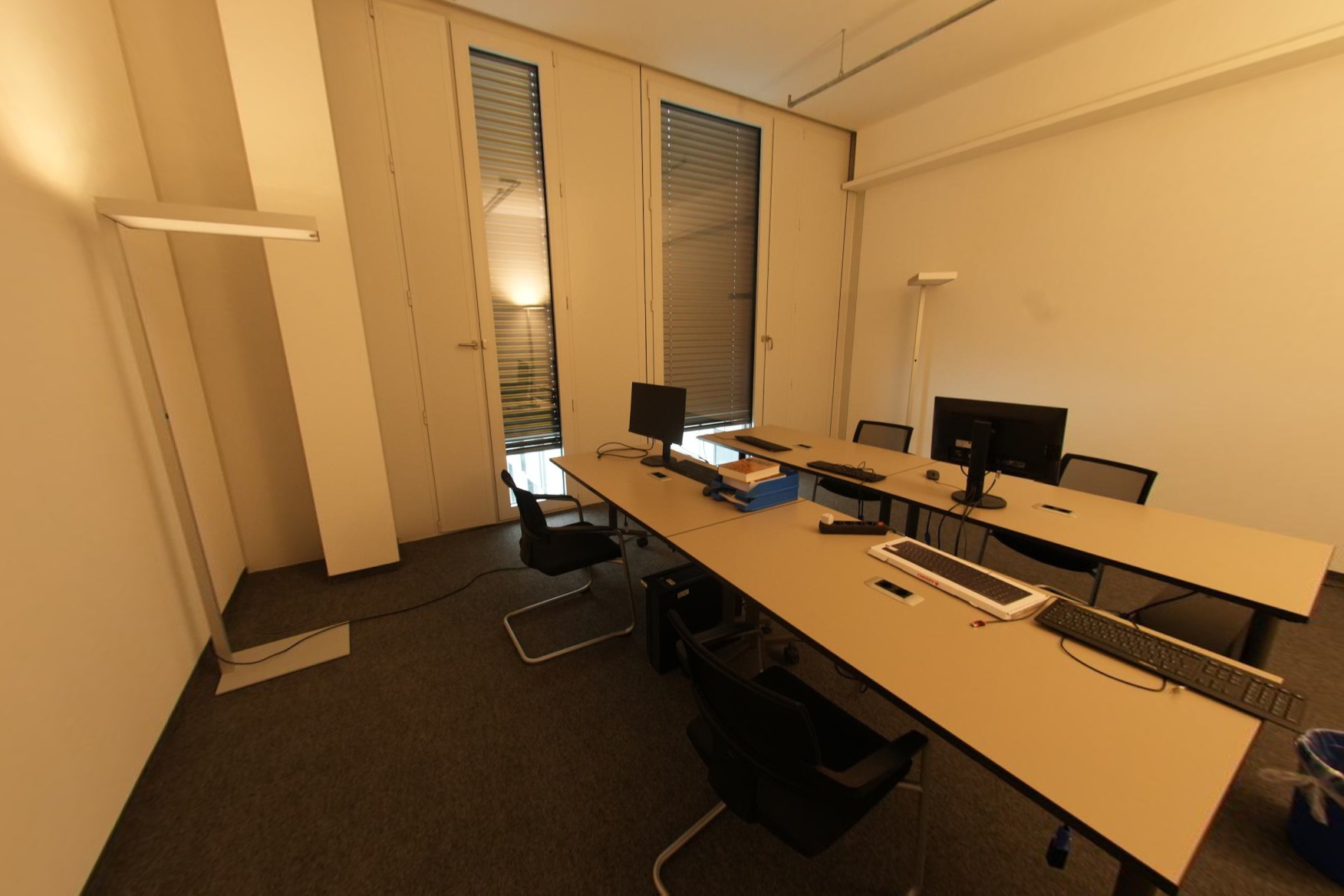}}   &
    \makecell{\includegraphics[width=\sz\linewidth]{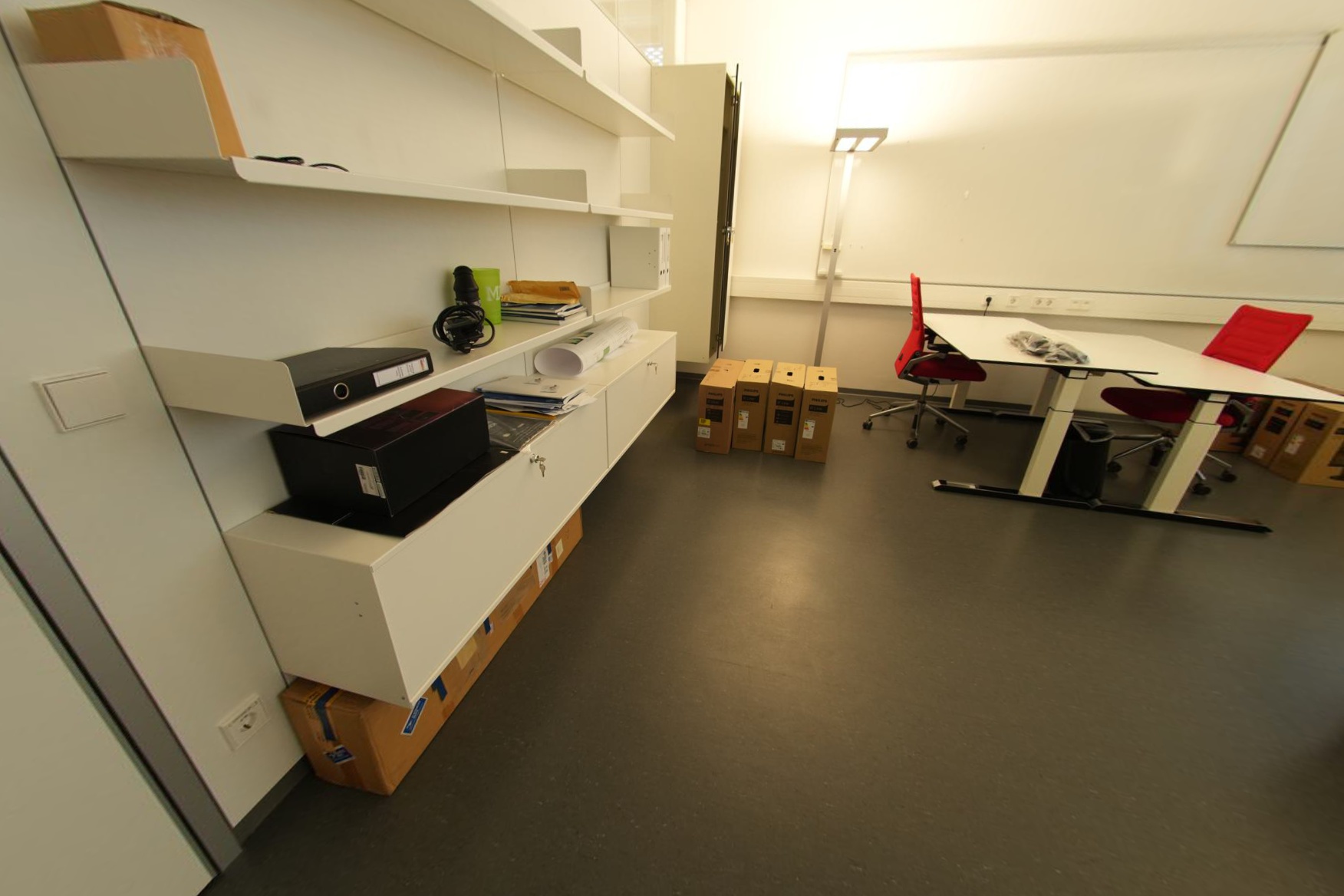}}       \\
\vspace{-0.5mm}
    \makecell{\rotatebox{90}{SplaTAM~\cite{keetha2024splatam}}}                               &
    \makecell{\includegraphics[width=\sz\linewidth]{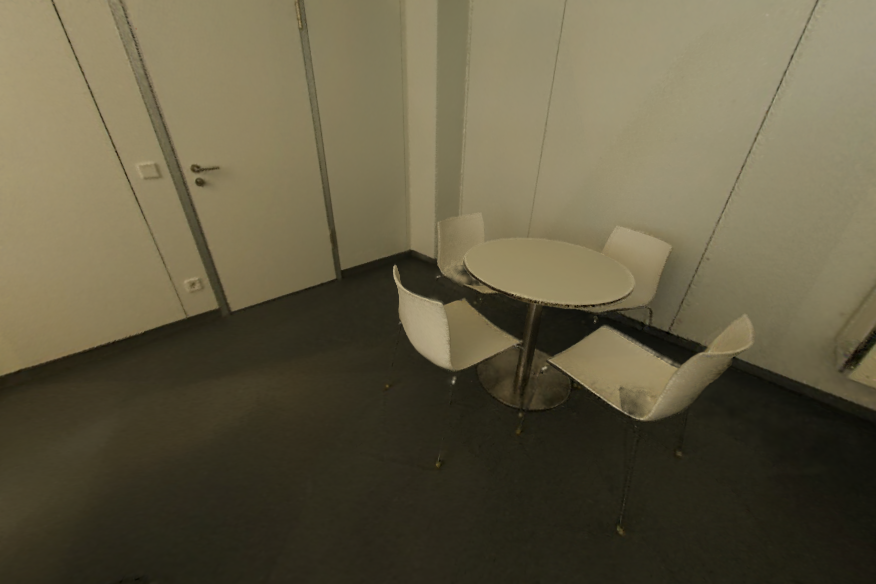}}        &
    \makecell{\includegraphics[width=\sz\linewidth]{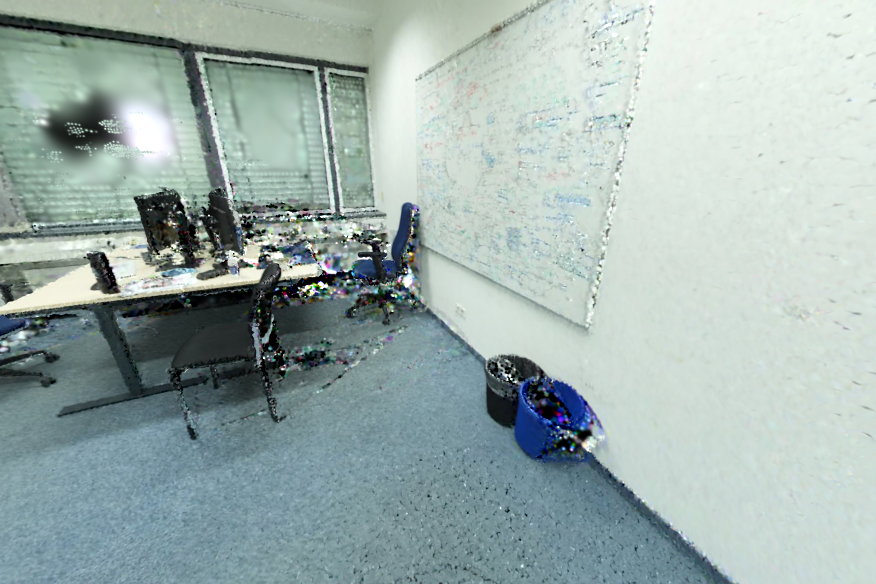}} &
    \makecell{\includegraphics[width=\sz\linewidth]{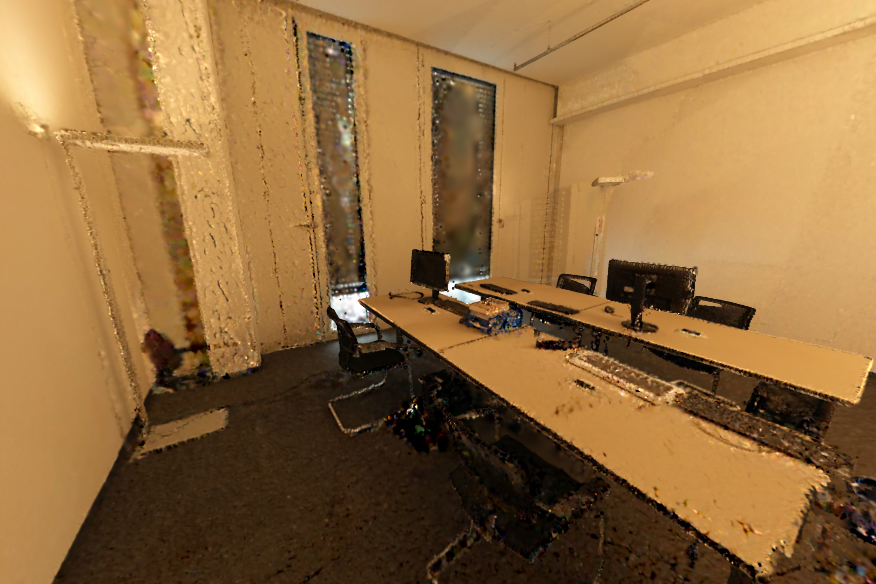}}   &
    \makecell{\includegraphics[width=\sz\linewidth]{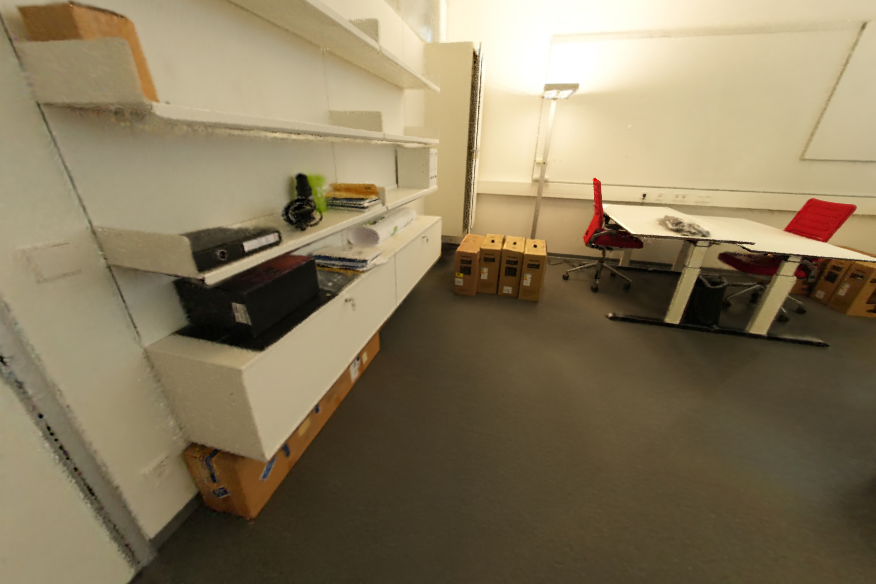}}       \\
\vspace{-0.5mm}
    \makecell{\rotatebox{90}{RTG-SLAM~\cite{peng2024rtg}}}                               &
    \makecell{\includegraphics[width=\sz\linewidth]{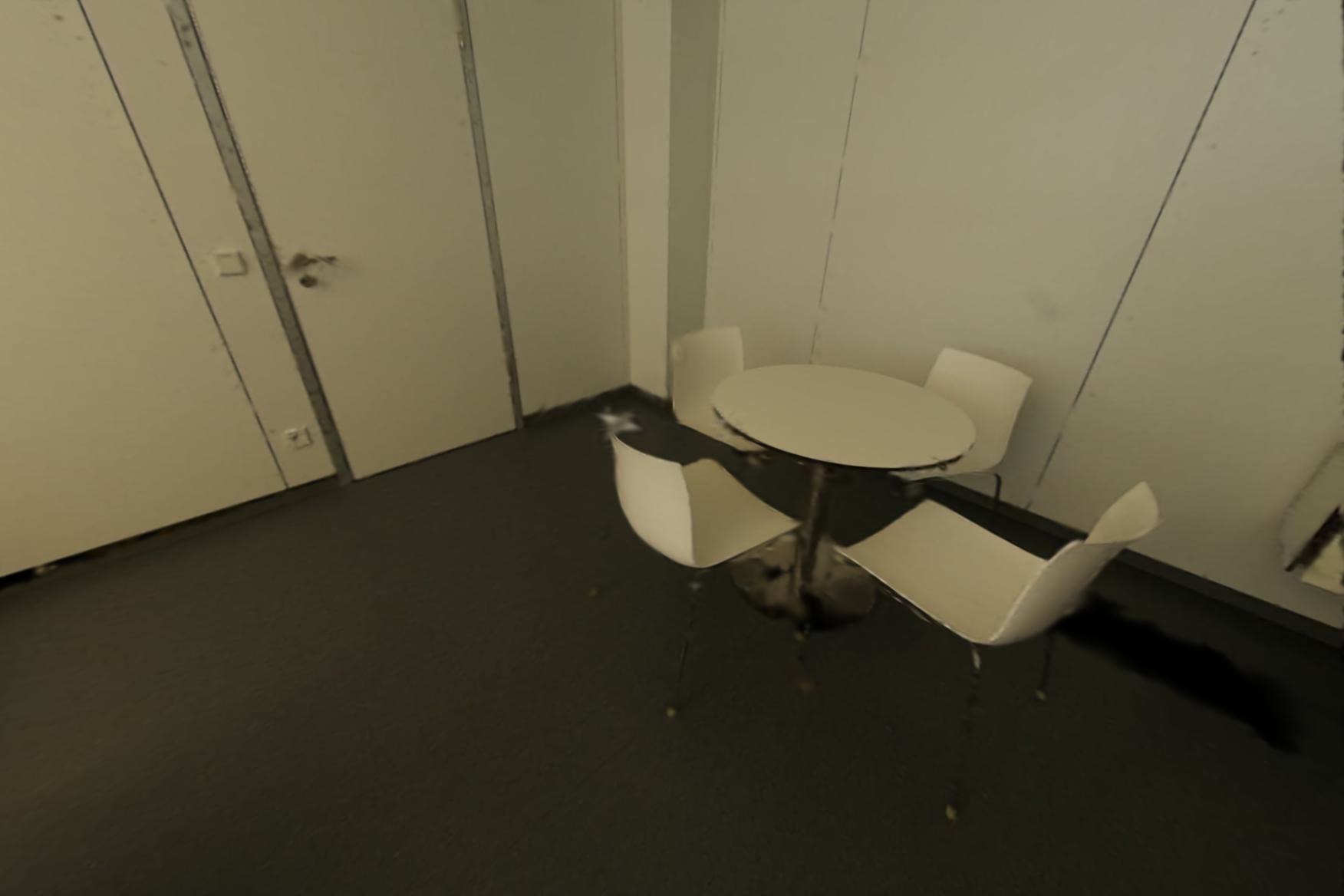}}          &
    \makecell{\includegraphics[width=\sz\linewidth]{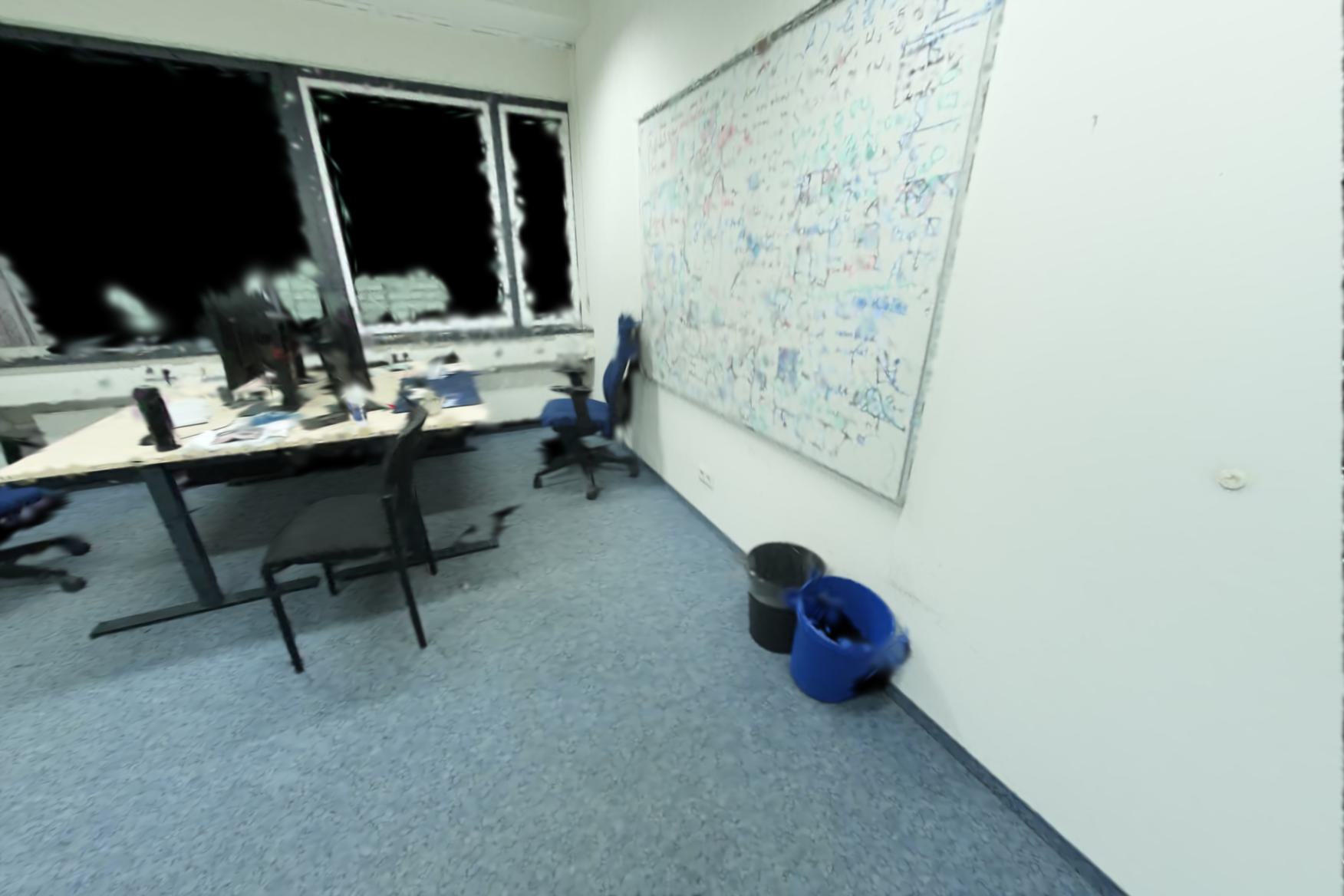}}   &
    \makecell{\includegraphics[width=\sz\linewidth]{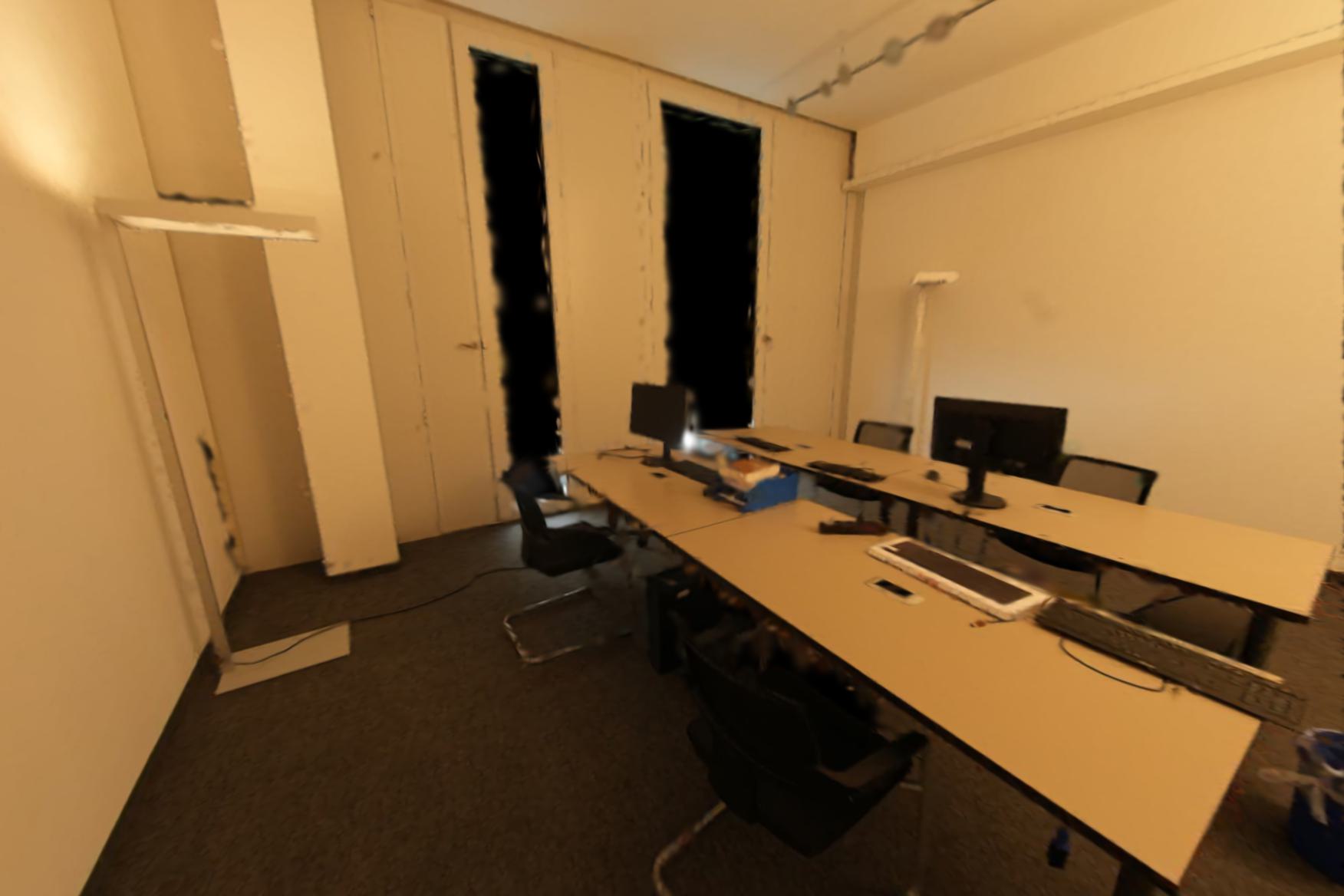}}     &
    \makecell{\includegraphics[width=\sz\linewidth]{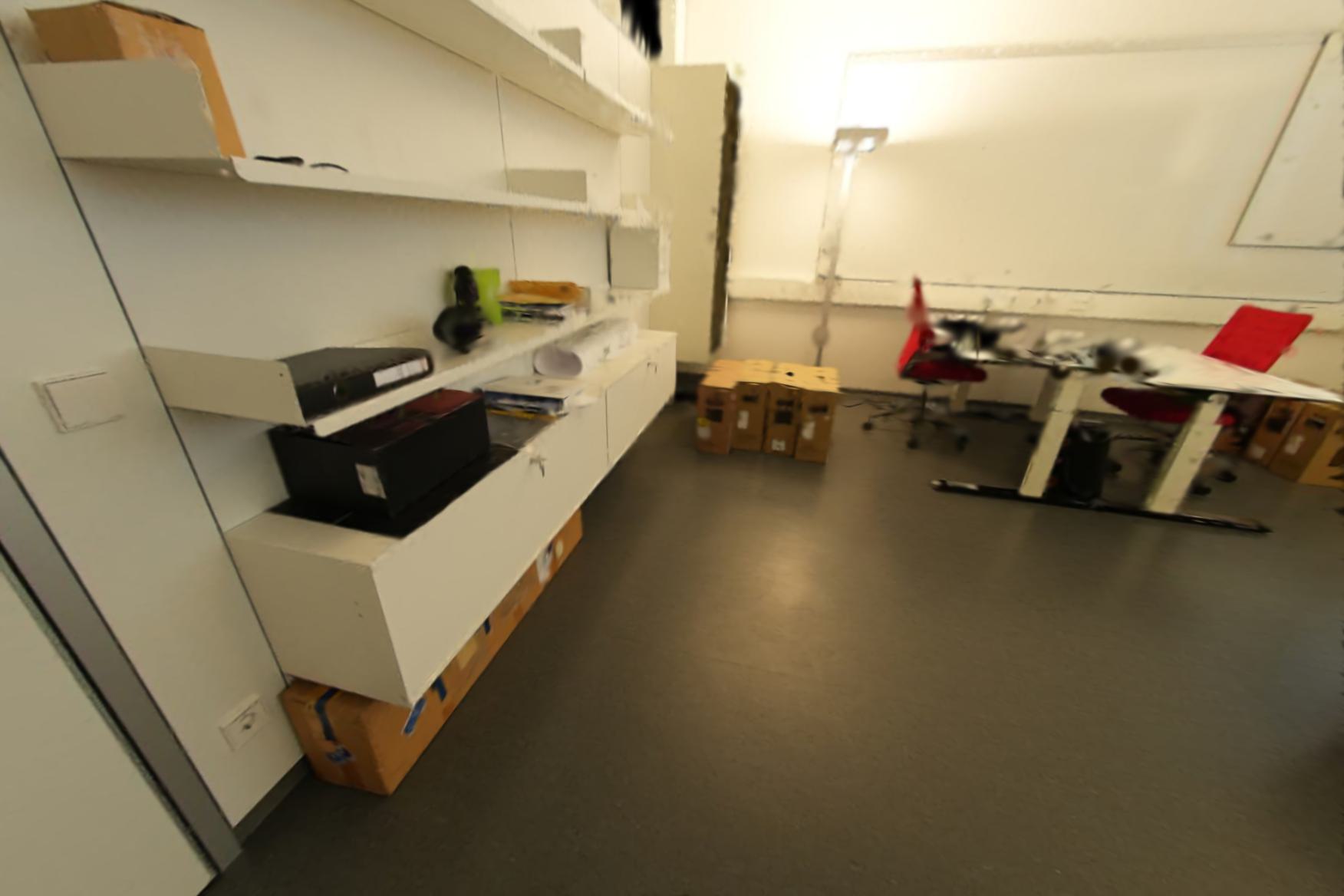}}          \\
\vspace{-0.5mm}
    \makecell{\rotatebox{90}{RP-SLAM (ours)}}                           &
    \makecell{\includegraphics[width=\sz\linewidth]{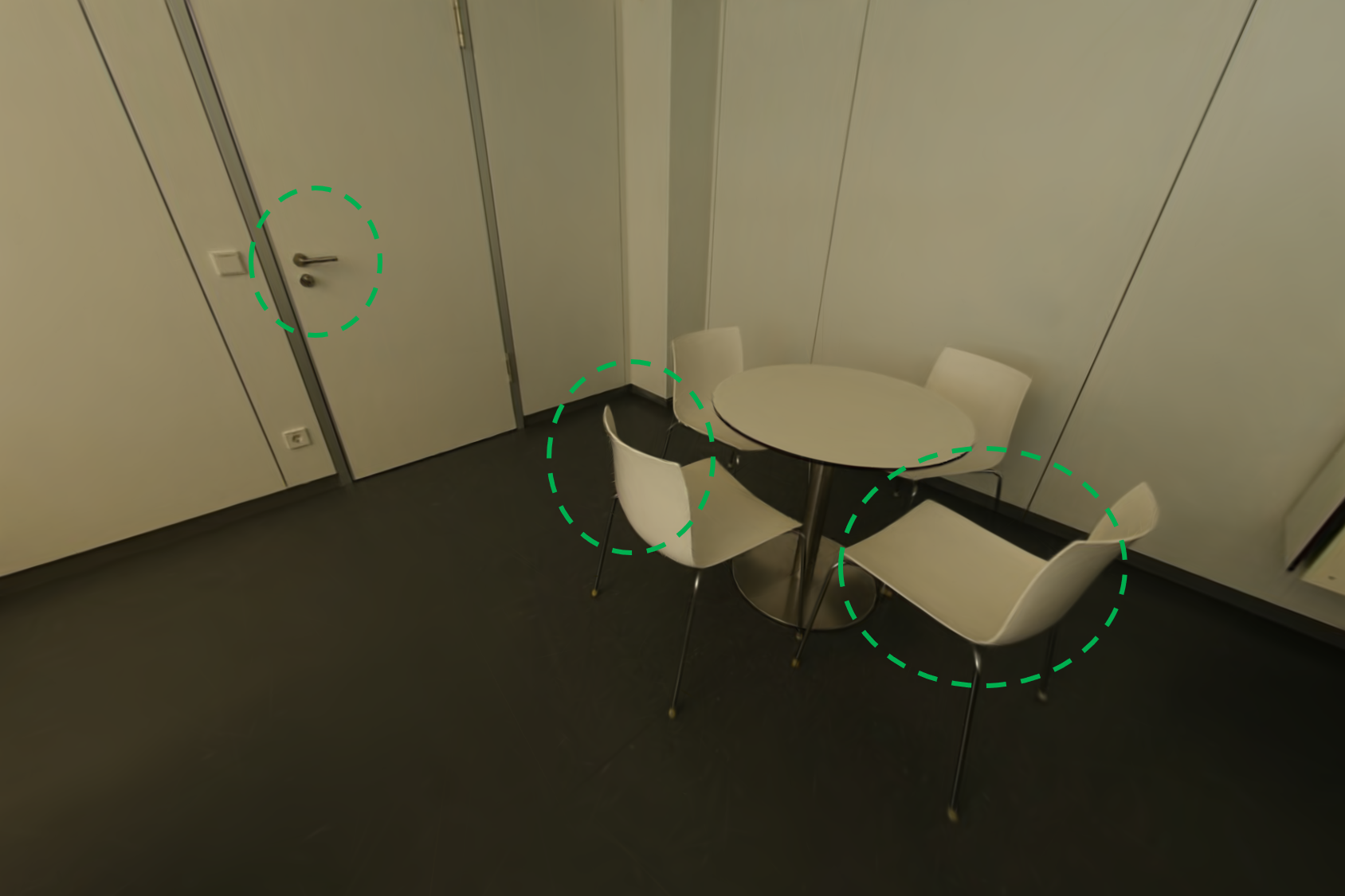}}         &
    \makecell{\includegraphics[width=\sz\linewidth]{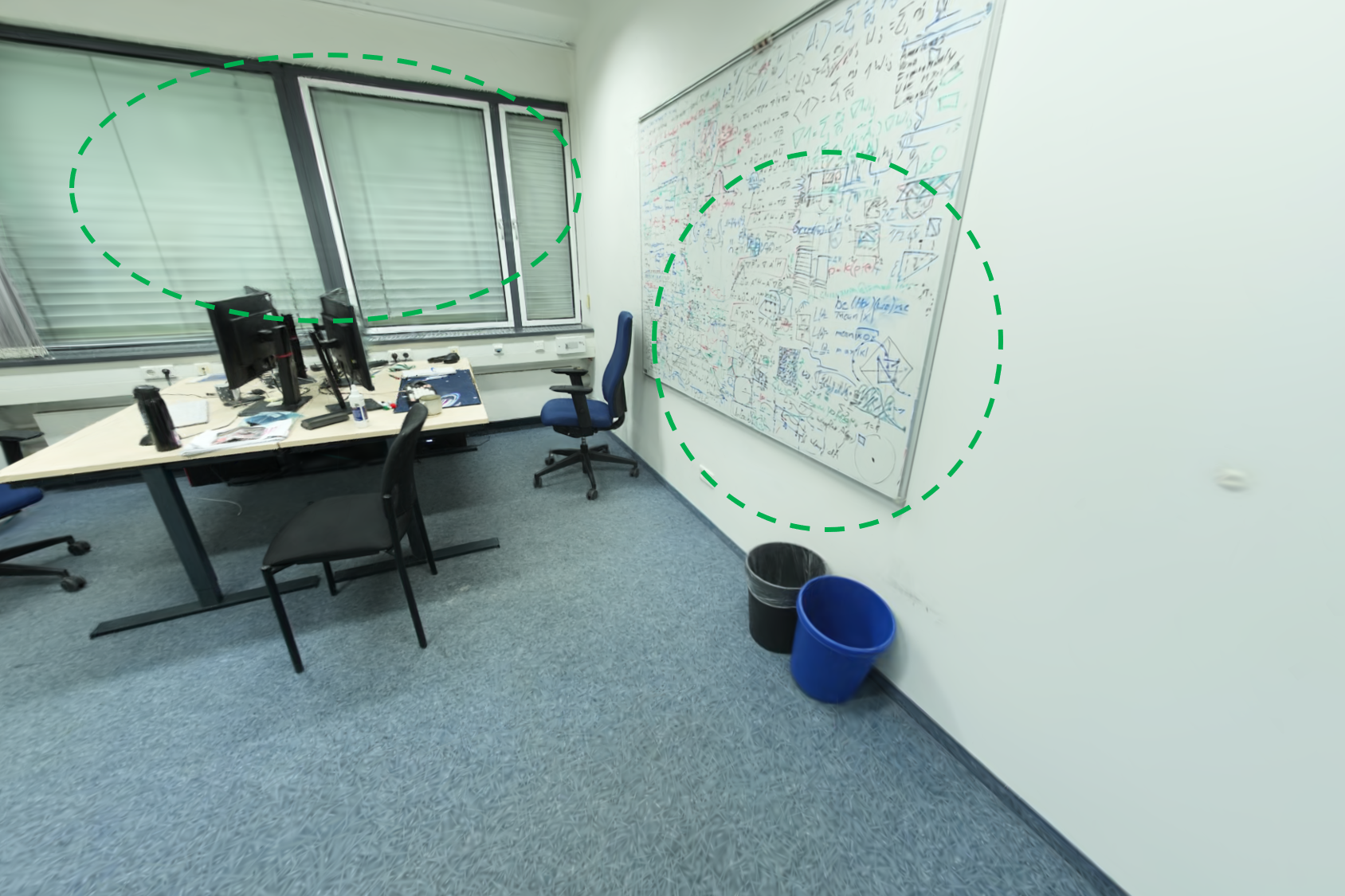}}  &
    \makecell{\includegraphics[width=\sz\linewidth]{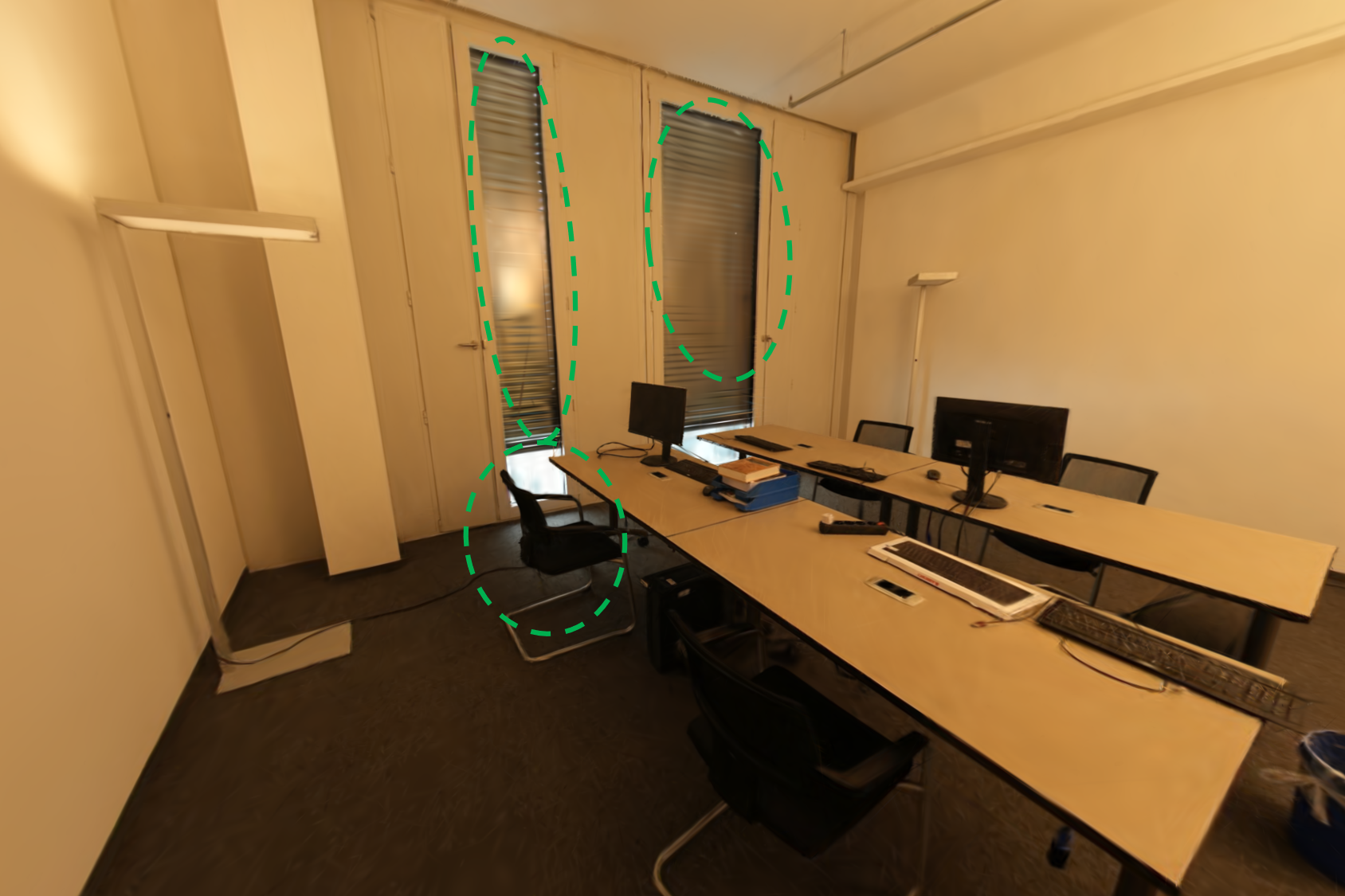}}    &
    \makecell{\includegraphics[width=\sz\linewidth]{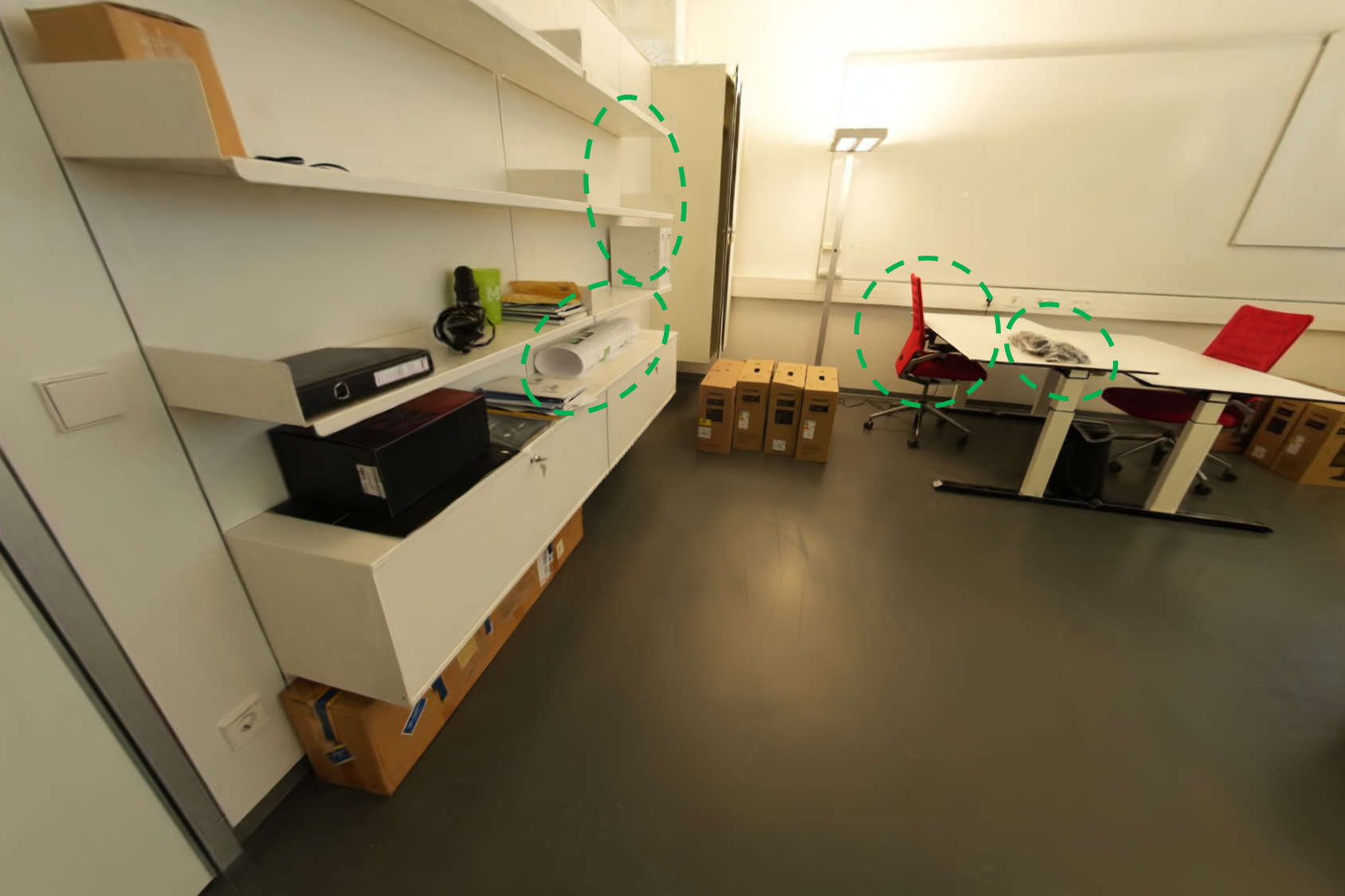}}         \\
                                                                                       &
    \makecell{(a)}                                                                     &
    \makecell{(b)}                                                                     &
    \makecell{(c)}                                                                     &
    \makecell{(d)}                                                                       \\
  \end{tabular}
  \caption{Qualitative comparisons on ScanNet++\cite{yeshwanth2023scannet++} dataset in the RGBD case. The green dashed boxes in our method mark areas where RP-SLAM outperforms other methods, such as sharper textures and fewer artefacts. Zoom in for a clearer view.}
  \label{render_scannetpp}
\end{figure*}

\subsection{Ablation Study}
\subsubsection{Different Modules Ablation}
The ablation study presented in Tab.~\ref{ab_modules} evaluates the contributions of the Monocular Keyframe Initialization (MKI), Efficient Incremental Mapping (EIM), and Dynamic Keyframe Window (DKW) in the monocular case of RP-SLAM on Replica\cite{straub2019replica} \texttt{Office0}. The analysis focuses on PSNR and model size (MS) metrics, demonstrating the roles of these modules in improving rendering quality and model representation efficiency.

In order to establish a baseline for comparison, each of the three modules of RP-SLAM is replaced with the corresponding module of MonoGS\cite{matsuki2024gaussian}, including random depth initialization, random sampling and fixed keyframe window. When all three modules are disabled, the baseline achieves a PSNR of 34.87 and a model size of 19.3 Mb. The lack of structured initialization, adaptive sampling, and consistency optimization results in suboptimal performance, with limited reconstruction quality and inefficient map representation. 

With only MKI enabled, the PSNR is increased to 36.67 and the model size is moderately reduced to 16.8 Mb. MKI ensures accurate placement of Gaussian primitives by leveraging the sparse point cloud generated by ORB-SLAM3\cite{campos2021orb}, which reduces the generation of redundant primitives to a certain extent. This module provides a solid geometric foundation for subsequent optimization, thereby improving reconstruction quality while maintaining reasonable storage efficiency.

Enabling only EIM drastically reduces the model size to 15.6 Mb and achieves a comparable result of 34.23 in PSNR to the baseline method. EIM employs quadtree-based adaptive sampling to focus computational resources on regions with high texture complexity, while minimizing redundant sampling of smooth regions and further eliminating redundant primitives through KNN filtering. Although the PSNR is not improved, the representation efficiency improvement can be clearly seen from the reduction of model size.

Activating only the DKW improves PSNR to 35.28 while slightly reducing the model size to 18.9 Mb. The DKW employs the use of co-visible keyframes to construct the aforementioned window during the iteration process, thereby mitigating the issue of forgetting and ensuring temporal consistency during the mapping process, which in turn improves the quality of the scene rendering.

RP-SLAM exhibits optimal performance with a PSNR of 37.74 and a model size of 11.8 Mb when all three modules (MKI, EIM, and DKW) are enabled. This configuration demonstrates the efficacy of the intermodule synergy. MKI ensures accurate initialization of Gaussian primitives, EIM minimizes redundancy while concentrating resources on critical regions, and DKW maintains temporal consistency and robustness against forgetting. Collectively, these modules empower RP-SLAM to attain the state-of-the-art rendering quality while markedly enhancing storage and model representation efficiency.

\subsubsection{Different Mininum Cell Sizes Ablation}
The ablation study presented in Tab.~\ref{ab_cell_size} examines the impact of minimum cell size in the quadtree-based adaptive sampling strategy on the performance of RP-SLAM in the RGB-D case. The minimum cell size determines the granularity of sampling during the mapping process, which directly affects the reconstruction quality (as measured by PSNR) and model efficiency (as measured by model size). The results demonstrate a clear trade-off between these two factors as the minimum cell size is varied.

When the minimum cell size is set to 4, the system exhibits the highest PSNR of 42.08, indicating that the reconstruction is of superior quality, with texture-rich regions being represented in detail. This configuration is particularly well-suited to the rendering of complex textures with high resolution in ScanNet++\cite{yeshwanth2023scannet++}, thereby ensuring a high level of detail in the reconstructed scene, as illustrated in Fig.~\ref{cell_size_render}. However, for common scenes, particularly those of low resolution, this results in an increase in the model size to 29.4 Mb, which reflects overly dense sampling in highly detailed sections, as illustrated in Fig.\ref{sampling} (a). Increasing the minimum cell size to 16 leads to a drastic reduction in the model size to just 5.8 Mb, showcasing the efficiency of the adaptive sampling mechanism in reducing redundancy. However, this configuration sacrifices rendering quality, with the PSNR dropping to 37.12. The coarser sampling reduces the system’s ability to capture fine-grained details in complex regions, as shown in Fig.\ref{sampling} (c), resulting in lower fidelity reconstructions. The configuration with a minimum cell size of 8 exhibits a balanced performance for common scenes, as evidenced by a PSNR of 41.26 and a model size of 10.1 MB. This setup provides high-quality reconstruction while maintaining the scene's efficiency and compactness. 

By modifying the minimum cell size, the system can be adapted to suit different scenarios, with the option of prioritizing either reconstruction quality or model efficiency in accordance with the specific requirements of the application.

\subsection{Qualitative Results}
The qualitative results presented in Fig.~\ref{render_mono}, Fig.~\ref{render_rgbd} and Fig.~\ref{render_scannetpp} compare RP-SLAM with several state-of-the-art methods, including MonoGS\cite{matsuki2024gaussian}, Photo-SLAM\cite{huang2024photo}, RTG-SLAM\cite{peng2024rtg}, SplaTAM\cite{keetha2024splatam} and CaRtGS\cite{feng2024cartgs}, in both monocular and RGB-D scenarios on  Replica\cite{straub2019replica} and ScanNet++\cite{yeshwanth2023scannet++} datasets.

In the monocular case (Fig.~\ref{render_mono}), RP-SLAM demonstrates superior performance in terms of texture clarity when compared to the baseline methods. To illustrate, in column (a), the remaining methods are largely unable to capture the intricate textures of the sofa, floor and table. In contrast, RP-SLAM accurately reconstructs the texture and edge shapes of these objects, closely aligning with the ground truth. A similar trend is observed in other scenes, where RP-SLAM produces clearer and more complete reconstructions. In the RGB-D case (Fig.~\ref{render_rgbd} and Fig.~\ref{render_scannetpp}), RP-SLAM once again demonstrates a notable enhancement in rendering quality relative to other comparable methods. For instance, in Fig.~\ref{render_rgbd} (c), RP-SLAM is capable of rendering with greater clarity, as evidenced by the sharpness of the clocks. Similarly, as illustrated in Fig.~\ref{render_scannetpp} (b), RP-SLAM is capable of acquiring more detailed information, such as the handwriting on the whiteboard, while simultaneously reducing the occurrence of artefacts, such as the windows.

In both monocular and RGB-D cases, RP-SLAM achieves superior quality performance by effectively balancing reconstruction fidelity and efficiency. The integration of MKI, EIM and DKW allows RP-SLAM to address the major limitations of the other methods. MKI provides robust initialization of Gaussian primitives, enabling accurate reconstruction even in monocular mode, while EIM dynamically adapts the sampling to focus on regions of interest, ensuring high quality reconstruction with minimal redundancy. DKW ensures temporal consistency and reduces artefacts caused by forgetting problems in sequential optimization. In conclusion, the qualitative results substantiate the efficacy of our RP-SLAM in attaining optimal rendering quality.

\section{Conclusion}
This paper introduces RP-SLAM, a 3D Gaussian splatting-based visual SLAM system designed to achieve real-time photorealistic scene reconstruction in monocular and RGB-D settings. Through the decoupling of camera poses estimation and scene optimization, RP-SLAM leverages feature-based SLAM for efficient and robust tracking while utilizing adaptive sampling and Gaussian primitives filtering to maintain high-quality scene representation with minimal redundancy. The dynamic keyframe window optimization ensures temporal consistency, addressing the forgetting problem during sequential mapping. For monocular configurations, the proposed Gaussian primitives initialization method provides a strong geometric foundation, enabling accurate reconstructions. Extensive evaluations validate the effectiveness of RP-SLAM in achieving state-of-the-art rendering quality and compact model size. 

\textbf{Feature Work}: Although this method is effective in balancing photorealistic reconstruction and model size, it is currently unable to accommodate dynamic scenes. Our future research will focus on extending the method to address this limitation.

\bibliographystyle{IEEEtran}
\bibliography{references}
\end{document}